%% file: main.tex
\documentclass{article}

\input{math_commands.tex}

\usepackage{microtype}
\usepackage{graphicx}
\usepackage{booktabs} 

\usepackage{hyperref}

\usepackage[accepted]{icml2023}

\usepackage{amsmath}
\usepackage{amssymb}
\usepackage{mathtools}
\usepackage{amsthm}

\usepackage[capitalize,noabbrev]{cleveref}

\theoremstyle{plain}

\theoremstyle{definition}

\theoremstyle{remark}

\usepackage[textsize=tiny]{todonotes}

\usepackage{url}
\usepackage{amsfonts}       
\usepackage{nicefrac}       
\usepackage{xcolor}         
\usepackage{makecell}
\usepackage{caption}
\usepackage{stackrel}
\usepackage{enumitem}
\usepackage{multirow}
\usepackage{wrapfig}
\usepackage{xcolor}
\usepackage{colortbl}

\usepackage{algorithm}
\usepackage{algorithmic}

\usepackage{caption}
\usepackage{subcaption}
\definecolor{navyblue}{rgb}{0.0, 0.0, 0.5}
\definecolor{darkblue}{rgb}{0.0, 0.0, 0.55}
\hypersetup{
    colorlinks = true,
    citecolor=darkblue,
    linkcolor=red,
    urlcolor  = blue,
    anchorcolor = blue
}

\definecolor{realblue}{RGB}{0,0,255}

\definecolor{gg}{gray}{0.92}
\newcolumntype{a}{>{\columncolor{gg}}c}

\icmltitlerunning{Personalized Subgraph Federated Learning}

\begin{document}

\twocolumn[
\icmltitle{Personalized Subgraph Federated Learning}



\icmlsetsymbol{equal}{*}

\begin{icmlauthorlist}
\icmlauthor{Jinheon Baek}{equal,kaist}
\icmlauthor{Wonyong Jeong}{equal,kaist}
\icmlauthor{Jiongdao Jin}{kaist}
\icmlauthor{Jaehong Yoon}{kaist}
\icmlauthor{Sung Ju Hwang}{kaist}
\end{icmlauthorlist}

\icmlaffiliation{kaist}{KAIST}

\icmlcorrespondingauthor{Jinheon Baek, and Sung Ju Hwang}{\{jinheon.baek, sjhwang82\}@kaist.ac.kr}

\icmlkeywords{Machine Learning, ICML}

\vskip 0.3in
]



\printAffiliationsAndNotice{\icmlEqualContribution} 

\input{sections/1_abstract}
\input{sections/2_introduction}

\input{sections/3_related_work}
\input{sections/4_definition}
\input{sections/5_methodology}

\input{sections/6_experiment}

\input{sections/7_conclusion}

\bibliography{custom}
\bibliographystyle{icml2023}

\newpage
\appendix
\onecolumn
\input{sections/8_appendix}

\end{document}

%% file: math_commands.tex
\usepackage{amsmath,amsfonts,bm}

\def\eqref#1{equation~\ref{#1}}

\def\1{\bm{1}}

\DeclareMathAlphabet{\mathsfit}{\encodingdefault}{\sfdefault}{m}{sl}
\SetMathAlphabet{\mathsfit}{bold}{\encodingdefault}{\sfdefault}{bx}{n}



%% file: sections/1_abstract.tex
\begin{abstract}

Subgraphs of a larger global graph may be distributed across multiple devices, and only locally accessible due to privacy restrictions, although there may be links between subgraphs. Recently proposed subgraph Federated Learning (FL) methods deal with those missing links across local subgraphs while distributively training Graph Neural Networks (GNNs) on them. However, they have overlooked the inevitable heterogeneity between subgraphs comprising different \textit{communities} of a global graph, consequently collapsing the incompatible knowledge from local GNN models. To this end, we introduce a new subgraph FL problem, \textit{personalized subgraph FL}, which focuses on the joint improvement of the interrelated local GNNs rather than learning a single global model, and propose a novel framework, \textit{FEDerated Personalized sUBgraph learning} (FED-PUB), to tackle it. Since the server cannot access the subgraph in each client, FED-PUB utilizes functional embeddings of the local GNNs using random graphs as inputs to compute similarities between them, and use the similarities to perform weighted averaging for server-side aggregation. Further, it learns a personalized sparse mask at each client to select and update only the subgraph-relevant subset of the aggregated parameters. We validate our FED-PUB for its subgraph FL performance on six datasets, considering both non-overlapping and overlapping subgraphs, on which it significantly outperforms relevant baselines. Our code is available at https://github.com/JinheonBaek/FED-PUB.

\end{abstract}

%% file: sections/2_introduction.tex
\vspace{-0.125in}
\section{Introduction}
\label{introduction}
\vspace{-0.025in}

Most existing Graph Neural Networks (GNNs)~\citep{GNN} focus on a single graph, whose nodes and edges collected from multiple sources are stored in a central server. For instance, in a social network platform, every user, with his/her social networks, contributes to creating a giant network consisting of all users and their connections. However, in some practical scenarios, each user/institution collects its own private graph, which is only locally accessible due to privacy restrictions. For instance, as described in~\citet{FedSage}, each hospital may have its own patient interaction network to track their physical contacts or co-diagnosis of disease; however, this graph may not be shared with others. How can we then collaboratively train, without sharing actual data, GNNs, when the subgraphs are distributed across multiple participants (i.e., clients)? The most straightforward way is to perform Federated Learning (FL) with GNNs, where each client individually trains a local GNN on the local data, while a central server aggregates locally updated GNN weights from multiple clients into one. 

However, an important challenge on it is how to deal with potentially \textit{missing edges} between subgraphs that are not captured by individual data owners, but may carry important information (See Figure~\ref{fig:concept} (A)). Recent subgraph FL methods~\citep{FedGNN, FedSage} tackle this problem by expanding the local subgraph from other subgraphs, as illustrated in Figure~\ref{fig:concept} (B). Specifically, they expand the local subgraph either by exactly augmenting the relevant nodes from the other subgraphs at the other clients~\citep{FedGNN}, or by estimating the nodes using the node information in the other subgraphs~\citep{FedSage}. However, such sharing of node information may compromise data privacy and can incur high communication costs.

\input{figures/concept}

Also, there exists a more important challenge that has been overlooked by the existing subgraph FL methods. We observe that they suffer from large performance degeneration (See Figure~\ref{fig:concept} right), due to the \textit{heterogeneity} among subgraphs, which is natural since subgraphs comprise different parts of a global graph. Specifically, two individual subgraphs -- for example, User 1 and 3 subgraphs in Communities A and B, respectively, in Figure~\ref{fig:concept} (A) -- are sometimes completely disjoint, having opposite properties. Meanwhile, two densely connected subgraphs form a community (e.g., User 1 and 2 subgraphs within the Community A of Figure~\ref{fig:concept} (A)), in which they share similar characteristics. However, it is challenging to consider such heterogeneity arising from the community structures of a graph in subgraph FL.

Motivated by this challenge, we introduce a novel problem of \emph{personalized subgraph FL}, whose goal is to jointly improve the interrelated local models trained on the interconnected local subgraphs, for instance, subgraphs belonging to the same community, by sharing weights among them (See Figure~\ref{fig:concept} (C)). However, realizing such selective weight sharing is challenging, since we do not know which subgraph each client has, due to its local accessibility. To address this issue, we use functional embeddings of GNNs on random graphs to obtain similarity scores between two local GNNs, and then use them to perform weighted averaging of the model parameters at the server. However, the similarity scores only tell how relevant each local model from the other clients is, but not which of the parameters are relevant. Thus we further learn and apply personalized sparse masks on the local GNN at each client to obtain only the subnetwork, relevant to the local subgraph. We refer to this framework as \textit{FEDerated Personalized sUBgraph learning} (FED-PUB).

We extensively validate our FED-PUB on six datasets with varying numbers of clients on both overlapping and disjoint subgraph FL scenarios. The experimental results show that ours significantly outperforms relevant baselines. Further analyses show that our functional embeddings can discover community structures among subgraphs, and the masking strategy localizes GNN parameters with respect to the subgraph of each client. Our contributions are as follows:
\vspace{-0.13in}
\begin{itemize}[itemsep=0.0mm, parsep=1pt, leftmargin=*]
    \item We introduce a novel problem of personalized subgraph FL, which aims at collaborative improvements of the related local models for subgraphs belonging to the same community, which has been relatively overlooked.
    \item We propose a novel framework for personalized subgraph FL, which performs weighted averaging of the local model parameters based on their functional similarities obtained without accessing the data, and learns sparse masks to select the relevant subnetworks for the given subgraphs.
    \item We validate our FED-PUB on six real-world datasets on both overlapping and non-overlapping node scenarios, demonstrating its effectiveness against relevant baselines.
\end{itemize}

%% file: figures/concept.tex
\begin{figure*}[t]
    \vspace{-0.05in}
    \begin{minipage}{0.66\textwidth}
    \centering
    \includegraphics[width=1.0\linewidth]{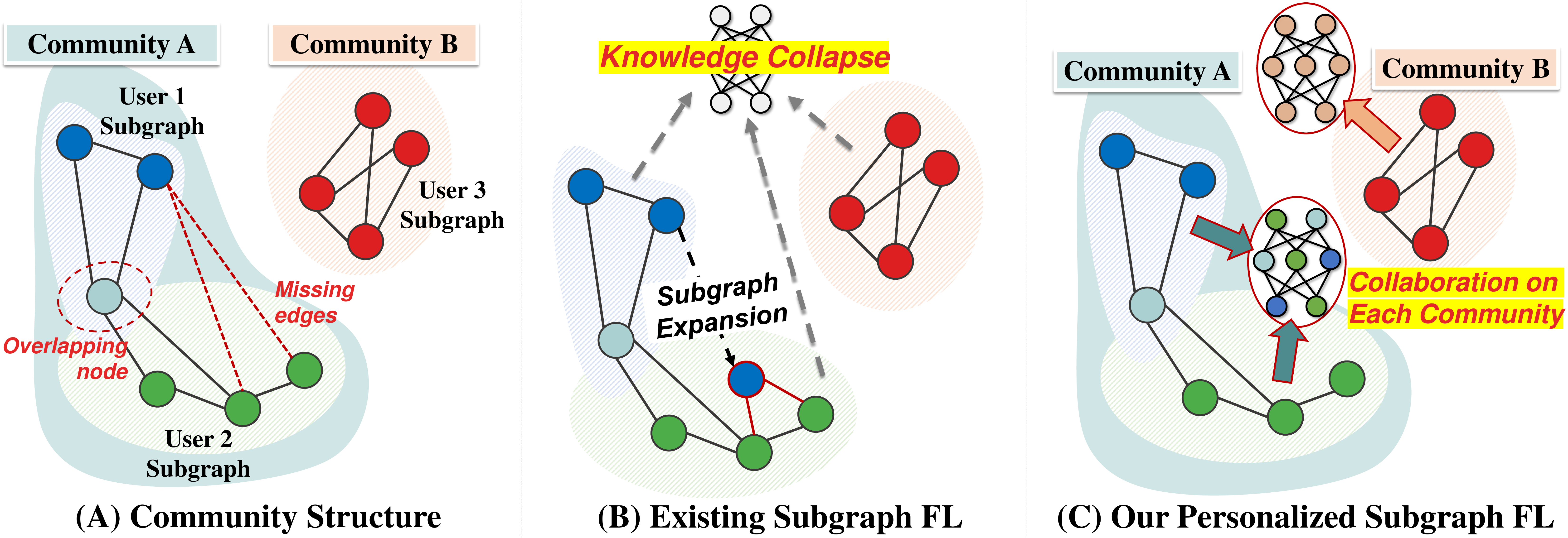}
    \end{minipage}
    \hfill
    \begin{minipage}{0.18\textwidth}
        \centering
        \includegraphics[width=1\linewidth]{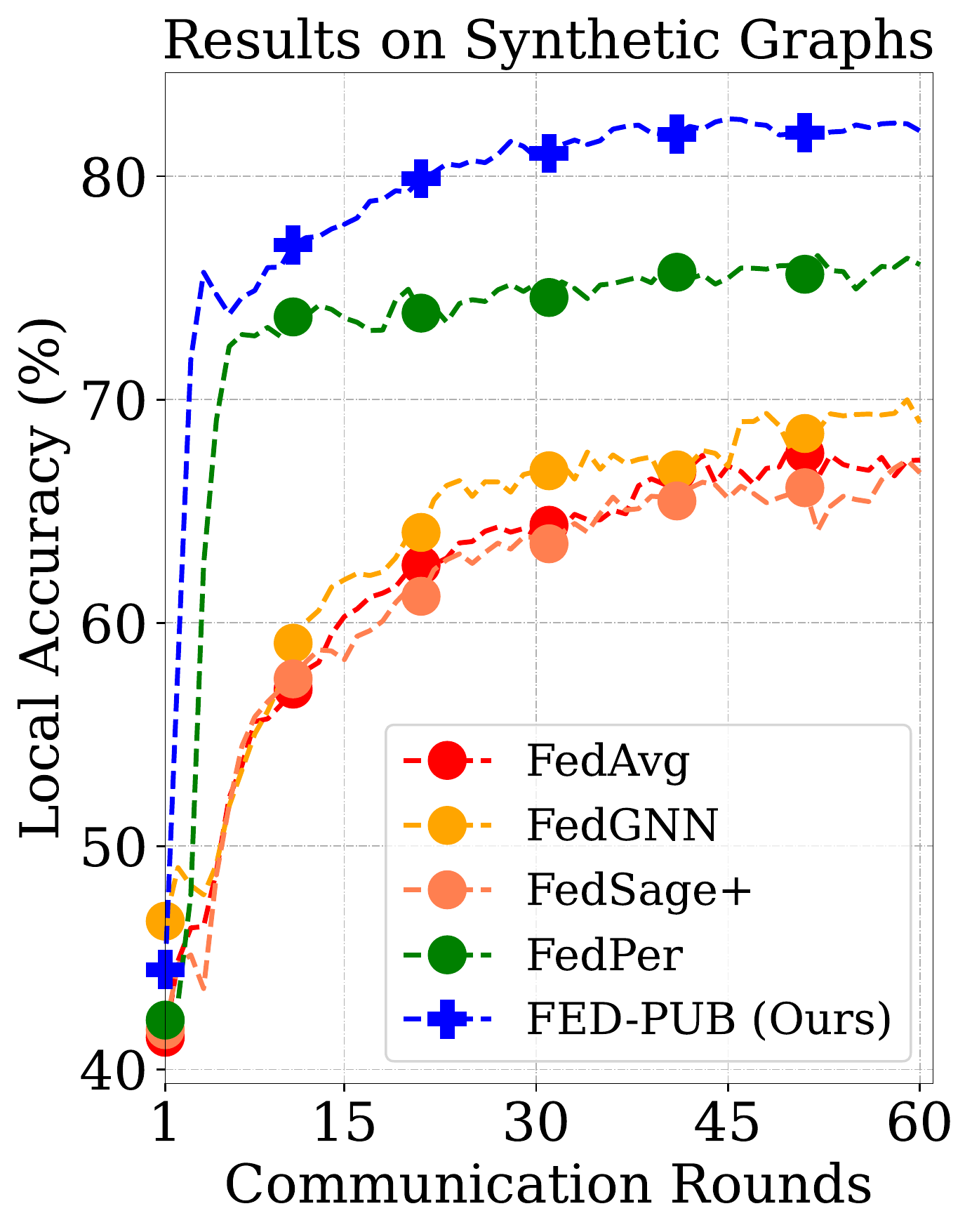}
    \end{minipage}
    \hfill
    \begin{minipage}{0.13\textwidth}
        \centering
        \includegraphics[width=1\linewidth]{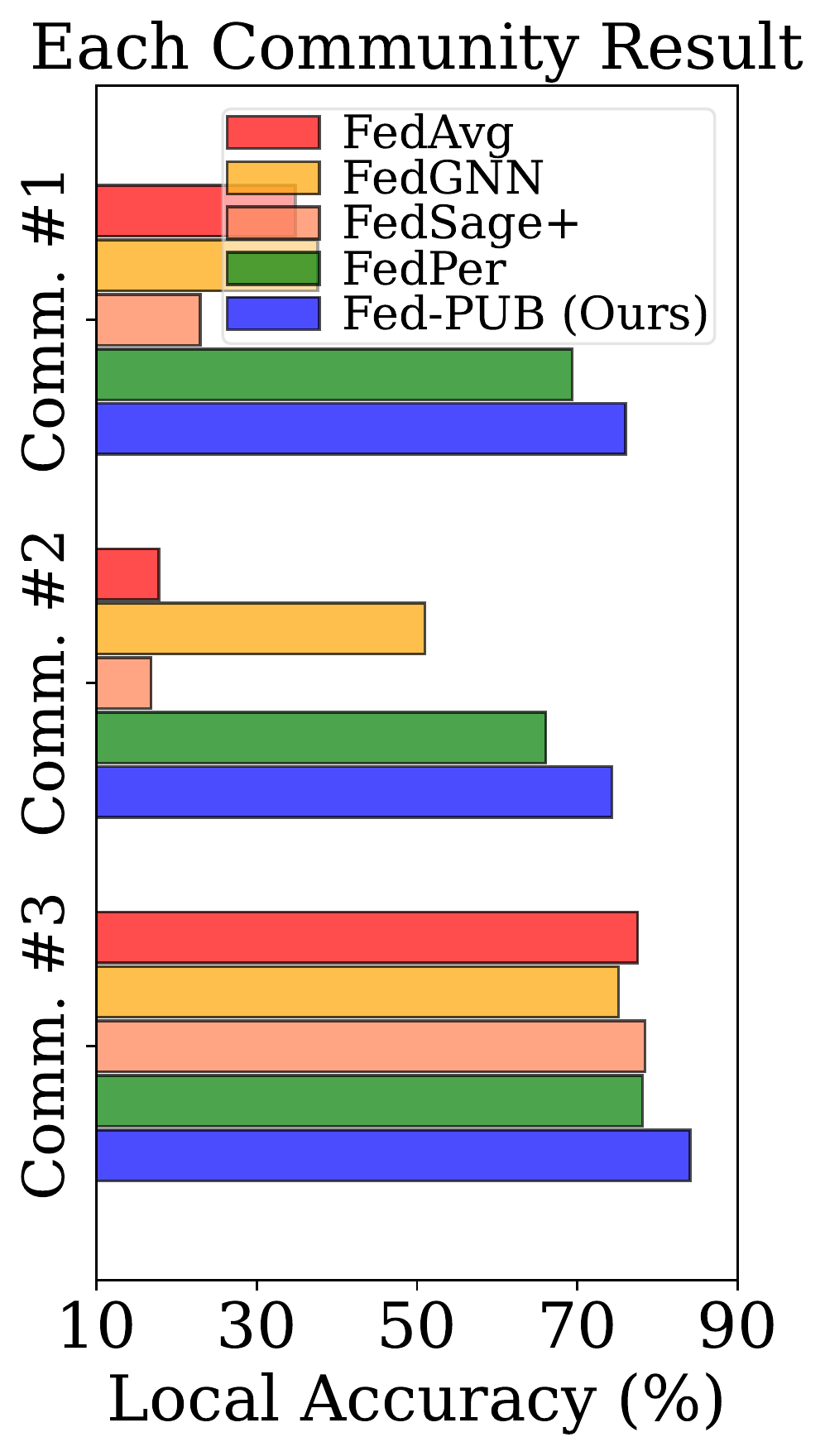}
    \end{minipage}
    \vspace{-0.13in}
    \caption{\small \textbf{(A) Local subgraphs} with overlapping nodes, missing edges, and community structures. \textbf{(B) Existing subgraph FL}~\citep{FedGNN, FedSage} expands local subgraphs to tackle the missing edge problem, but collapses the incompatible knowledge from heterogeneous subgraphs. \textbf{(C) Our personalized subgraph FL} focuses on the joint improvement of local models working on interrelated subgraphs, such as ones within the same community, by selectively sharing the knowledge across them. \textbf{(Right:) Knowledge collapse results}, where local models belonging to two small communities (Communities 1 and 2) suffer from performance degeneration by existing subgraph FL, FedGNN~\citep{FedGNN, FedPerGNN} and FedSage+~\citep{FedSage}. A personalized FL method, FedPer~\citep{arivazhagan2019federated}, also underperforms ours since it focuses on individual model's improvement without considering community structures.}
    \label{fig:concept}
    \vspace{-0.2in}
\end{figure*}

%% file: sections/3_related_work.tex
\section{Related Work}
\label{related_work} 
\vspace{-0.03in}

\textbf{Graph Neural Networks (GNNs)}, which aim to learn the representations of nodes, edges, and entire graphs, are an extensively studied topic~\citep{GNN, GNN/1, GNN/2, edge, gmt}. Most existing GNNs under a message passing scheme~\citep{MPNN} iteratively represent a node by aggregating features from its neighboring nodes as well as itself. For example, Graph Convolutional Network (GCN)~\citep{GCN} approximates the spectral graph convolutions~\citep{spectralgraph}, yielding a mean aggregation over neighboring nodes. Similarly, for each node, GraphSAGE~\citep{GraphSAGE} aggregates the features from its neighbors to update the node representation. While they lead to successes on node classification and link prediction tasks for a single graph, they are not directly applicable to real-world systems with locally distributed graphs, where graphs from different sources are not shared across participants, which gives rise to federated learning to train GNNs.

\textbf{Federated Learning}~\citep{FLsurvey} is essential for our distributed subgraph learning problem. To mention a few, FedAvg~\citep{McMahan2017CommunicationEfficientLO} locally trains a model for each client and then transmits the trained model to a server, while the server aggregates the model weights from local clients and then sends the aggregated model back to them. However, since the locally collected data may largely vary across different clients, heterogeneity is a crucial issue. To tackle this, FedProx~\citep{li2020federated} proposes the regularization term that minimizes the weight differences between local and global models, which prevents the local model from diverging to the local training data. However, when the local data is extremely heterogeneous, it is more appropriate to collaboratively train a personalized model for each client rather than learning a single global model. FedPer~\citep{arivazhagan2019federated} is such a method, which shares base layers while having local personalized layers for each client, to keep the local knowledge. Further, recent studies propose to distill the outputs from different clients~\citep{distil/2, distil/3, distil/4}, or directly minimize the differences of their model outputs~\citep{distil/1}. However, unlike the commonly studied image and text data, graph-structured data is defined by connections between instances, which yields additional challenges: missing edges, and community structures between private subgraphs.

\textbf{Graph Federated Learning.} 
Few recent studies suggest using the FL framework to collaboratively train GNNs without sharing graph data~\citep{he2021fedgraphnn, graphflsystem}, which can be broadly classified into subgraph- and graph-level methods. Graph-level FL methods assume that different clients have completely disjoint graphs (e.g., molecular graphs), and recent work~\citep{GCFL, SpreadGNN, graphfl} focuses on the heterogeneity among non-IID graphs (i.e., differences in graph labels across clients). Unlike the graph-level FL that has similar challenges to general FL scenarios, the subgraph-level FL we target has a unique graph-structural challenge: there exist missing yet probable links between subgraphs, since a subgraph is a part of a larger global graph. To deal with such a missing link problem among subgraphs, existing methods~\citep{FedGNN, FedSage, FedGCN} augment the nodes by requesting the node information in the other subgraphs, and then connecting the existing nodes with the augmented ones. However, this scheme may compromise data privacy constraints, and also increases communication overhead across clients, during the node information sharing process. Unlike them focusing on the problem of missing links, our subgraph FL method tackles the new problem with a completely different perspective by exploring subgraph communities~\citep{Community1, Community}, which are groups of densely connected subgraphs.

%% file: sections/4_definition.tex
\section{Problem Statement}

\label{definition}

We explain GNNs and FL, then define our novel problem of personalized subgraph FL lying at the intersection of them.

\paragraph{Graph Neural Networks} 
A graph $\mathcal{G} = (\mathcal{V}, \mathcal{E})$ consists of a set of $n$ nodes $\mathcal{V}$ and a set of $m$ edges $\mathcal{E}$ along with its node feature matrix $\bm{X} \in \mathbb{R}^{n \times d}$, where each row represents a $d$-dimensional node feature. $(u, v) \in \mathcal{E}$ represents an edge from a node $u$ to a node $v$. Then, Graph Neural Networks (GNNs)~\citep{GNN} represent nodes based on their neighborhoods and themselves, as follows: 
\begin{equation}
\fontsize{9pt}{9pt}\selectfont
    \bm{H}^{l+1}_v
    = \text{UPD}^{l}\left( \bm{H}^{l}_v, \text{AGG}^{l}\left(\left\{  \bm{H}^{l}_u: \forall u\in\mathcal{N}(v) \right\}\right) \right),
    \label{eq:mp}
\fontsize{9pt}{9pt}\selectfont
\end{equation}
where $\bm{H}^{l}_v$ is the features of the node $v$ at $l$-th layer, $\mathcal{N}(v)$ denotes a set of adjacent nodes of the node $v$: $\mathcal{N}(v) = \left\{ u \in \mathcal{V} \; | \; (u, v) \in \mathcal{E} \right\}$, $\text{AGG}$ aggregates the features of $v$'s neighbors, and $\text{UPD}$ updates the node $v$'s representation given its previous representation and the aggregated representations from its neighbors. $\bm{H}^{1}$ is initialized as $\bm{X}$.

\paragraph{Federated Learning} The goal of FL is to collaboratively train models with their local data. Let assume we have $K$ clients with locally collected data inaccessible from others: $\mathcal{D}_k=\{\bm{X}_i, \bm{y}_i\}_{i=1}^{N_k}$ for the $k$-th client, where $\bm{X}_i$ is a data instance, $\bm{y}_i$ is its corresponding class label, and $N_k$ is the number of data instances. Then, a popular FL algorithm, FedAvg~\citep{McMahan2017CommunicationEfficientLO}, works as follows:
\vspace{-0.075in}
\begin{enumerate}[itemsep=1.0mm, parsep=1pt, leftmargin=*]
    \item \textbf{(Initialization)} At the initial communication round $r=0$, the central server initializes the local model parameters of $K$ clients as the global parameters $\boldsymbol{\bar \theta}$, as follows: $\boldsymbol{\theta}_k^{(0)} \leftarrow \boldsymbol{\bar \theta}^{(0)}$ $\forall k$, where $\boldsymbol{\theta}_k^{(0)}$ is the $k$-th client parameters.
    \item \textbf{(Local Updates)} Each local model performs training on the local data $\mathcal{D}_k$ to minimize the task loss $\mathcal{L}(\mathcal{D}_k; \boldsymbol{\theta}_k^{(0)})$, and then updating the parameters: $\boldsymbol{\theta}^{(1)}_{k} \leftarrow \boldsymbol{\theta}^{(0)}_{k}-\eta\nabla\mathcal{L}$.
    \item \textbf{(Global Aggregation)} After local training, the server aggregates the local knowledge with respect to the number of training instances, i.e., $\boldsymbol{\bar \theta}^{(1)} \leftarrow \frac{N_{k}}{N} \sum_{k=1}^{K} \boldsymbol{\theta}_{k}^{(1)}$ with $N=\sum_k N_k$, and distributes the updated global parameters $\boldsymbol{\bar \theta}^{(1)}$ to local clients selected at the next round.
\end{enumerate}

\vspace{-0.1in}
It iterates between Step 2 and 3 until reaching the final round $R$, which \textit{shares only parameters without private local data}.

\input{figures/method}

\paragraph{Challenges in Subgraph FL} While the above FL works well on image and text data, due to the unique characteristics of graphs, there exist nontrivial challenges for applying this FL scheme to graph-structured data. In particular, unlike with an image domain where each instance $\bm{X}_i$ is independent to the other images, each node $v$ in a graph is always influenced by its relationships to adjacent nodes $\mathcal{N}(v)$. Moreover, a local graph $G_i$ could be a subgraph of a larger global graph $\mathcal{G}$: $G_i \subseteq \mathcal{G}$. In such a case, there could be missing edges between subgraphs in two different clients: $(u, v)$ with $u \in \mathcal{V}_i$ and $v \in \mathcal{V}_j$ for clients $i$ and $j$, respectively. To tackle this problem, existing methods~\citep{FedGNN, FedSage} estimate the nodes of a local subgraph $G_k$ based on the node information from subgraphs at other clients $G_i$ with $\forall i \ne k$, and then extend the existing nodes with the estimated ones. However, this augmentation scheme incurs high communication costs as it requires sharing node information across clients, which may also violate data privacy constraints~\citep{abadi2016deep}.

Yet, there exists a more challenging issue. Assume that we have a global graph consisting of all subgraphs. Then, there are \emph{communities} of such subgraphs~\citep{Community1, Community, Community2}, where subgraphs within the same community are more densely connected to each other than subgraphs outside the community. Formally, a global graph $\mathcal{G}$ can be decomposed into $T$ different communities: $C_i \subseteq \mathcal{G}$ $\forall i=1,...,T$, where $i$-th community $C_i = (\mathcal{V}_i, \mathcal{E}_i)$ consists of densely connected nodes. Then, in a subgraph FL problem, a local subgraph $G_j$ belongs to at least one community: $C_i=\bigcup_{j=1}^{J} G_{j}$. Note that, based on a network homophily~\citep{homophily}, densely connected subgraphs within the same community have similar properties, while subgraphs in two opposite communities are not. Such distributional heterogeneity of communities may lead the naive FL algorithm to collapse the incompatible knowledge from different communities.

\paragraph{Personalized Subgraph FL} To alleviate the above knowledge collapse issue, we aim to personalize the subgraph FL algorithm by performing personalized weight averaging and masking of local model parameters; thereby capturing the community structures among interrelated subgraphs. To be more formal, the objective of existing subgraph FL~\citep{FedGNN, FedSage, liu2021federated} is as follows: $\min_{\boldsymbol{\bar \theta}} \sum_{G_i \subseteq \mathcal{G}} \mathcal{L}(G_i; \boldsymbol{\bar \theta})$. However, finding a universal set of parameters (i.e., $\boldsymbol{\bar \theta}$) that works on all subgraphs will result in finding the suboptimal parameter set, since subgraphs in two different communities with sparse connections are extremely heterogeneous due to the network homophily. To address this limitation, we formulate a novel problem of personalized subgraph FL, formalized as follows:
\begin{align}
\fontsize{9.0pt}{9.0pt}\selectfont
    \min_{\left\{ \boldsymbol{\theta}_i, \boldsymbol{\mu}_i \right\}_{i=1}^K} & \sum_{G_i \subseteq \mathcal{G}} \mathcal{L}(G_i; \boldsymbol{\theta}_i, \boldsymbol{\mu}_i), \; \boldsymbol{\theta}_i \leftarrow  \boldsymbol{\mu}_i \odot \left( \sum_{j=1}^K \alpha_{ij} \boldsymbol{\theta}_j \right) \nonumber \\[1ex]
    & \text{with} \; \alpha_{ik} \gg \alpha_{il} \; \text{for} \; G_k \subseteq C \; \text{and} \; G_l \nsubseteq C,
    \label{eq:def}
\fontsize{9.0pt}{9.0pt}\selectfont
\end{align}
where $\boldsymbol{\theta}_i$ is the weight for subgraph $G_i$ belonging to community $C$. $\alpha_{ij}$ is a coefficient for weight aggregation between clients $i$ and $j$, which can promote collaborative learning across local models of interrelated subgraphs that belong to the same community, by assigning larger weights. Yet, this scalar coefficient $\alpha_{ij}$ cannot inform us which elements of the aggregated weight are relevant to subgraph $G_i$. Therefore, we further multiply it to the trainable sparse vector $\boldsymbol{\mu}_i$ with element-wise multiplication $\odot$, to shift and filter out irrelevant weights from subgraphs of heterogeneous communities. We will specify how to obtain $\alpha$ and $\boldsymbol{\mu}$ in Section~\ref{method}.

%% file: figures/method.tex
\begin{figure*}[t]
    \centering
    \includegraphics[width=1.0\linewidth]{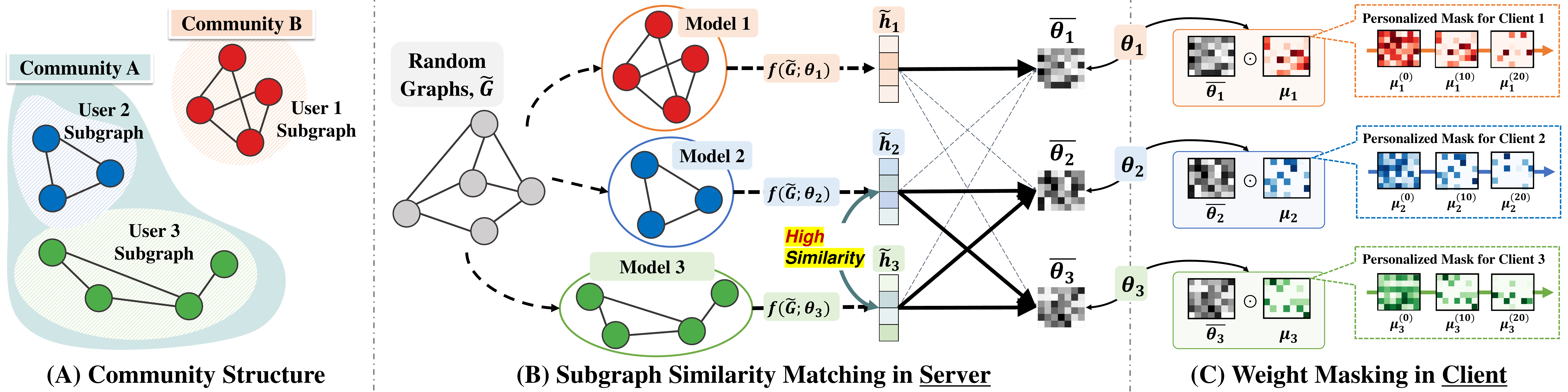}
    \vspace{-0.26in}
    \caption{\small \textbf{(A) Two communities}, where Community A and B consist of two and one subgraphs, respectively. \textbf{(B) Similarity Matching}: we first forward randomly generated graphs to models $f(\tilde G; \boldsymbol{\theta}_i)$, and obtain functional embeddings $\boldsymbol{\tilde{h}}_i$, which are then used to estimate subgraph similarities. Then, the similarities are used in weight aggregation, resulting in personalized model weights $\boldsymbol{\bar \theta}_i$. \textbf{(C) Weight Masking:} transmitted weights from the server to clients $\boldsymbol{\bar \theta}_i$ are masked and shifted by local masks $\boldsymbol{\mu}_i$ for localization to local subgraphs.}
    \vspace{-0.1in}
    \label{fig:method}
\end{figure*}

%% file: sections/5_methodology.tex
\begin{figure}
 \vspace{-0.05in}
 \centering
  \begin{minipage}{0.20\linewidth}
  \includegraphics[width=1.0\textwidth]{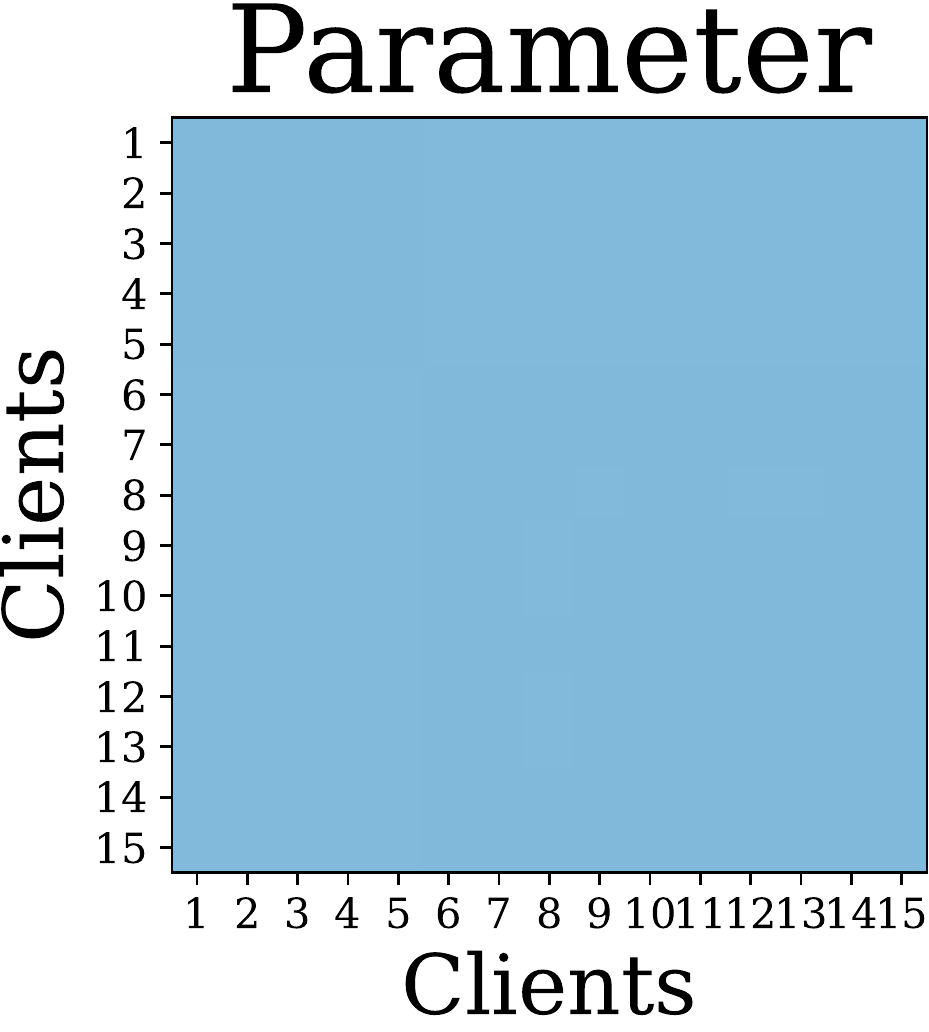}
  \end{minipage}
  \begin{minipage}{0.20\linewidth}
  \includegraphics[width=1.0\textwidth]{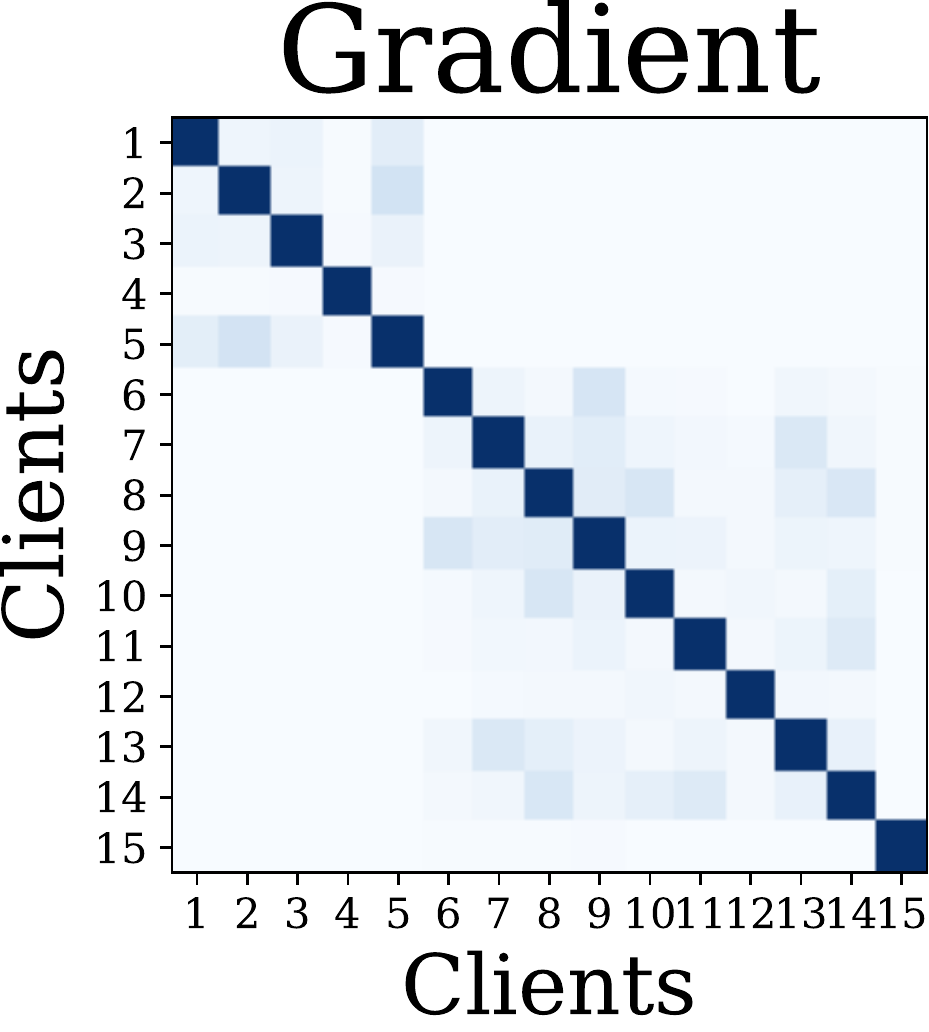}
  \end{minipage}
  \begin{minipage}{0.20\linewidth}
  \includegraphics[width=1.0\textwidth]{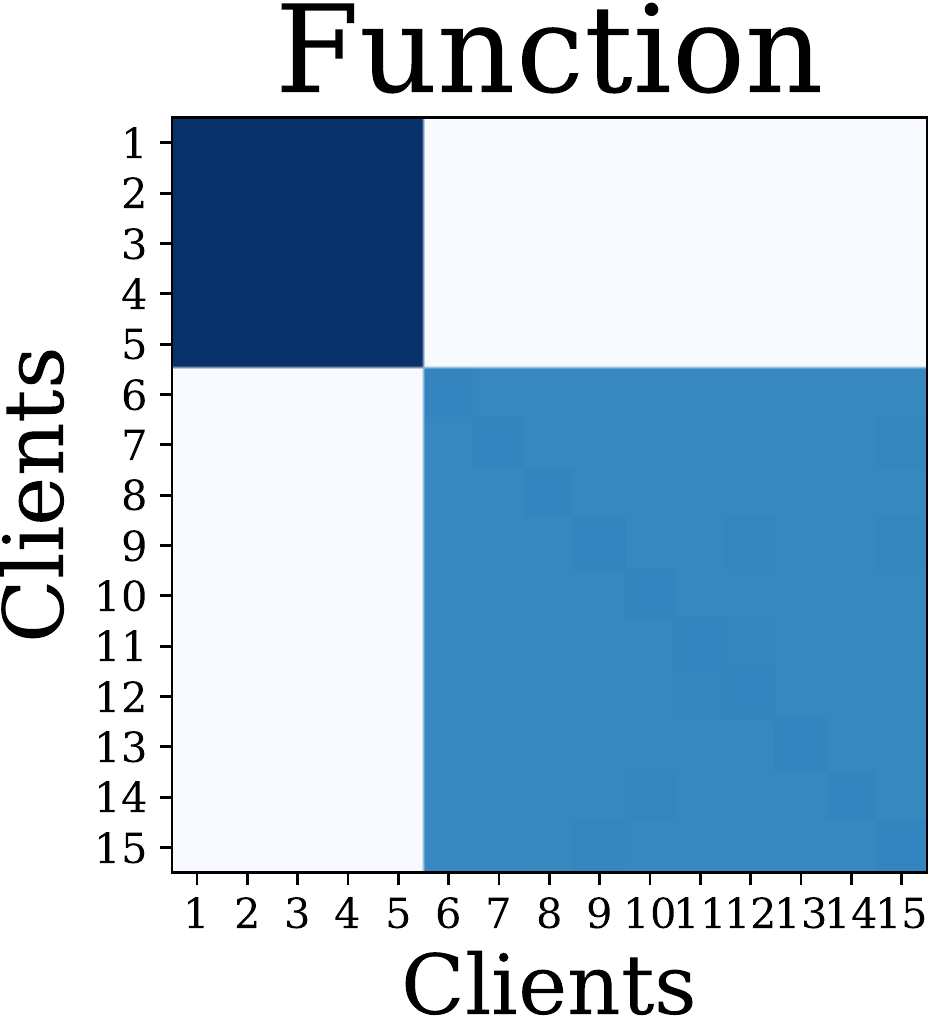}
  \end{minipage}
  \begin{minipage}{0.36\linewidth}
  \vspace{0.03in}
  \includegraphics[width=1.0\textwidth]{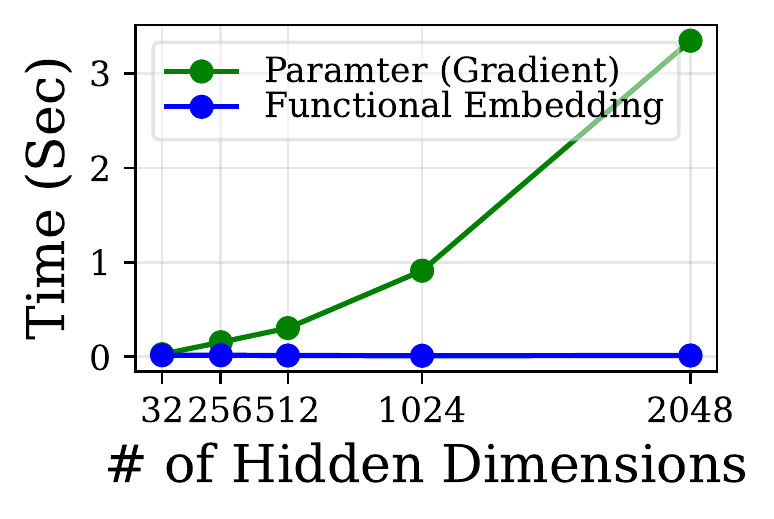}
  \end{minipage}
  \vspace{-0.15in}
  \caption{\small Effectiveness (left) and efficiencies (right) of different similarity measurements: parameter, gradient, and our function.}
  \label{fig:similarity}
  \vspace{-0.5in}
\end{figure}

\section{Federated Personalized Subgraph Learning}
\vspace{-0.05in}
\label{method}

To realize our approach in equation~\ref{eq:def}, we propose to compute subgraph similarities for discovering communities, and to mask weights from subgraphs in unrelated communities.

\vspace{-0.05in}
\subsection{Subgraph Similarity Estimation}
\label{method:community}
\vspace{-0.05in}

We aim at capturing the community consisting of a group of densely connected subgraphs. Note that, due to the network homophily where similar instances in the graph are more associated with each other~\citep{homophily}, subgraphs within the same community should be similar. Therefore, if one can measure subgraph similarities, we can group similar ones into the community. However, measuring the similarity between subgraphs is challenging since we do not know which subgraph each client has due to its local accessibility. To compute similarities \textit{only using the transmittable GNN parameters without accessing the local data}, we propose to approximate the similarities using auxiliary information obtained from GNNs working on subgraphs.

\vspace{-0.05in}
\paragraph{Model Parameters for Subgraph Similarities} 
To measure the similarity between local subgraphs without accessing them, we may use the model parameters as proxies, as follows: $S(i, j) = ({\boldsymbol{\theta}_i \cdot \boldsymbol{\theta}_j }) / ({ \| \boldsymbol{\theta}_i\|  \| \boldsymbol{\theta}_j\| }$), where $\boldsymbol{\theta}$ is a parameter flatten into a vector, and $S$ is a similarity measure. This may sound reasonable since the GNN trained on the subgraph will embed its knowledge into its parameters. However, this scheme has notable drawbacks that similarity measured in the high-dimensional space is not meaningful due to the curse of dimensionality~\citep{bellman1966dynamic}, and that the cost of calculating the similarity between parameters grows rapidly as the model size increases (See Figure~\ref{fig:similarity}).

\paragraph{Functional Embeddings for Subgraph Similarities} To address these limitations, we propose to measure the functional similarity of GNNs by feeding the same input to all GNN models and then calculating the similarities using their outputs, inspired by~\citet{tans}. The main intuition here is that we can consider the transformation defined with a neural network as a function, and we measure the functional similarity of two networks by the distance of their outputs for the same input. However, unlike the previous work, which uses Gaussian noises as inputs for image classification, we use random graphs as inputs as we work with GNNs. Formally, let $\tilde{G}=(\mathcal{\tilde{V}}, \mathcal{\tilde{E}})$ be a random community graph designed by a stochastic block model~\citep{SBM}, where subgraphs within the community have more edges between them than edges across the communities (See Appendix~\ref{appendix:sub:imple} for details). The similarity between two functions defined by GNNs at clients $i$ and $j$ is then defined as follows:
    $S(i, j) = \frac{\boldsymbol{\tilde h}_i \cdot \boldsymbol{\tilde h}_j }{ \| \boldsymbol{\tilde h}_i\|  \| \boldsymbol{\tilde h}_j\| }$,
where $\boldsymbol{\tilde h}$ is the averaged output of all node embeddings for input $\tilde{G}$ with average operation, $\texttt{AVG}$: $\boldsymbol{\tilde h}_i = \texttt{AVG}(f(\tilde{G}; \boldsymbol{\theta}_i))$. Note that this functional similarity is effective and efficient, compared to parameter and gradient similarities (See Figure~\ref{fig:similarity}). Also, it uses only parameters sent to the server, which does not compromise data privacy. For more discussions on variants of random graphs and similarity estimations, see Appendix~\ref{appendix:results:functional} and~\ref{appendix:results:similarity}.

\paragraph{Personalized Weight Aggregation}
With the similarity measure, $S(i, j)$, we now aim to share parameters between GNNs working on similar subgraphs, by using the weighted sum of model parameters across different clients~\cite{FedGraph, jeong2022factorized}. Note that entirely ignoring the model parameters from different communities may result in exploiting only the local objective while ignoring the globally useful weights, which results in suboptimal performance (See Appendix~\ref{appendix:results:explicit_community}). Therefore, we perform weighted averaging of local GNNs from all clients based on their functional similarities, as follows (Figure~\ref{fig:method} (B)):
\begin{equation}
    \boldsymbol{\bar \theta}_i \leftarrow \sum_{j=1}^K \alpha_{ij} \cdot \boldsymbol{\theta}_j, \quad \alpha_{ij} = \frac{\text{exp}(\tau \cdot S(i, j))}{\sum_k \text{exp}(\tau \cdot S(i, k))},
    \label{eq:agg}
\end{equation}
where $\alpha_{ij}$ is a normalized similarity between clients $i$ and $j$, and $\tau$ is a hyperparameter for scaling the unnormalized similarity score. Notably, increasing the value of $\tau$ (e.g., 10) will result in model averaging done almost exclusively among subgraphs detected as belonging to the same community.

This personalized scheme handles two challenges in subgraph FL. First, unlike the global weight aggregation which collapses the knowledge from heterogeneous communities, our subgraph FL allows the models belonging to different communities to obtain individual parameters that are beneficial for each of both communities. Also, missing edges (i.e., a lack of information sharing) between interconnected subgraphs, which are explicitly dealt with by expanding local subgraphs in existing works~\citep{FedGNN, FedSage}, could be implicitly handled by largely sharing the knowledge among GNNs of probably linked subgraphs within the same community (See Figure~\ref{fig:heatmap}, Figure~\ref{fig:neighbor}, and Appendix~\ref{appendix:results:missingedge} for results). This enhances data privacy while minimizing communication costs between subgraphs.

\subsection{Adaptive Weight Masking}
\label{method:mask}
Based on the previous similarity matching scheme, we can effectively group GNNs that belong to the same community; therefore, preventing the collapsing of irrelevant knowledge from opposite communities. However, the heterogeneity in subgraph FL is extremely severe due to the community structures (See Appendix~\ref{appendix:heterogeneity} for more discussions). Therefore, the previous scalar weighting scheme might be insufficient, since it considers only how much each local model from other clients is relevant, but not \emph{which} parameters are relevant. Thus we propose to select only the relevant parameters from the aggregated GNN weights transmitted from a server, similar to the existing weight masking literature~\cite{Mask/1, Mask/2, Mask/3}.

\paragraph{Personalized Parameter Masking} 
We aim to perform selective training and updating of models by modulating and masking their aggregated parameters using the sparse local masks (Figure~\ref{fig:method} (C)). To realize this on GNNs, we apply the local mask to the GNN weights, and their resulting weights are used for updating features of neighboring nodes during the message passing in Equation~\ref{eq:mp}. Formally, let $\boldsymbol{\mu}_i$ be a local mask for client $i$, which is a free variable and not shared. Then, our local GNN weight is obtained by modulating the weight from the server, as follows: $\boldsymbol{\theta}_i=\boldsymbol{\bar \theta}_i \odot \boldsymbol{\mu}_i$, where $\odot$ is an element-wise multiplication operation between the globally given weight $\boldsymbol{\bar \theta}_i$ and the local mask $\boldsymbol{\mu}_i$. Also, we initialize $\boldsymbol{\mu}_i$ as ones, in order to start training with the globally initialized GNNs without modification. We then further promote sparsity on the mask $\boldsymbol{\mu}$, to take two advantages. First, we can transmit only the partial parameters, that have not been sparsified at the client, to the server rather than sending all parameters, thus reducing communication costs. Also, if local masks are sufficiently sparse, local models can run faster, when zero-skipping operations are supported. To take these benefits from sparsity, we use $L_1$ regularizer on $\boldsymbol{\mu}_i$ when performing local optimization (See Appendix~\ref{appendix:sub:imple} for details on sparsification), formalized in equation~\ref{eq:loss}.

\input{tables/main_overlap}
\input{figures/fig_table_1}

\paragraph{Preventing Local Divergence with Proximal Term} As masks are trained only with limited local data without parameter sharing, they may be easily overfitted to the training instances in each client. To alleviate this issue, we adopt the proximal term proposed in~\citet{li2020federated} that regularizes the locally updated model $\boldsymbol{\theta}_i$ to be closer to the globally given model $\boldsymbol{\bar \theta}_i$, therefore, preventing the local model from extremely drifting to the local data distribution. To sum up, at $i$-th client, our objective function including sparsity and proximal terms with $L_1$ and $L_2$ losses is denoted as follows: 
\begin{equation}
    \min_{(\boldsymbol{\theta}_i, \boldsymbol{\mu}_i)} \mathcal{L}(G_i; \boldsymbol{\theta}_i, \boldsymbol{\mu}_i) + \lambda_1 \| \boldsymbol{\mu}_i \|_1 + \lambda_2 \| \boldsymbol{\theta}_i-\boldsymbol{\bar \theta}_i \|_2^2.
    \label{eq:loss}
\end{equation}
$\mathcal{L}$ is a certain loss function with hyperparameters $\lambda_1$, $\lambda_2$.

%% file: tables/main_overlap.tex
\begin{table*}[t]
\caption{\small \textbf{Results on the overlapping node scenario.} The reported results are mean and standard deviation over three different runs. The statistically significant performances ($p > 0.05$) are emphasized in bold.}
\vspace{-0.1in}
\label{tab:main:overlap}
\small
\centering
\resizebox{\textwidth}{!}{
\renewcommand{\arraystretch}{1.0}
\begin{tabular}{lccccccccca}
\toprule
& \multicolumn{3}{c}{\bf Cora} & \multicolumn{3}{c}{\bf CiteSeer} & \multicolumn{3}{c}{\bf Pubmed} & - \\
\cmidrule(l{2pt}r{2pt}){2-4} \cmidrule(l{2pt}r{2pt}){5-7} \cmidrule(l{2pt}r{2pt}){8-10} \cmidrule(l{2pt}r{2pt}){11-11}
\textbf{Methods} & \textbf{10 Clients} & \textbf{30 Clients} & \textbf{50 Clients} & \textbf{10 Clients} & \textbf{30 Clients} & \textbf{50 Clients} & \textbf{10 Clients} & \textbf{30 Clients} & \textbf{50 Clients} & - \\
\midrule

Local   & 73.98 $\pm$ 0.25 & 71.65 $\pm$ 0.12 & 76.63 $\pm$ 0.10 & 65.12 $\pm$ 0.08 & 64.54 $\pm$ 0.42 & 66.68 $\pm$ 0.44 & 82.32 $\pm$ 0.07 & 80.72 $\pm$ 0.16 & 80.54 $\pm$ 0.11 & - \\

\midrule

FedAvg  & 76.48 $\pm$ 0.36 & 53.99 $\pm$ 0.98 & 53.99 $\pm$ 4.53 & 69.48 $\pm$ 0.15 & 66.15 $\pm$ 0.64 & 66.51 $\pm$ 1.00 & 82.67 $\pm$ 0.11 & 82.05 $\pm$ 0.12 & 80.24 $\pm$ 0.35 & - \\
FedProx & 77.85 $\pm$ 0.50 & 51.38 $\pm$ 1.74 & 56.27 $\pm$ 9.04 & 69.39 $\pm$ 0.35 & 66.11 $\pm$ 0.75 & 66.53 $\pm$ 0.43 & 82.63 $\pm$ 0.17 & 82.13 $\pm$ 0.13 & 80.50 $\pm$ 0.46 & - \\
FedPer  & 78.73 $\pm$ 0.31 & 74.18 $\pm$ 0.24 & 74.42 $\pm$ 0.37 & 69.81 $\pm$ 0.28 & 65.19 $\pm$ 0.81 & 67.64 $\pm$ 0.44 & 85.31 $\pm$ 0.06 & 84.35 $\pm$ 0.38 & 83.94 $\pm$ 0.10 & - \\
GCFL    & 78.84 $\pm$ 0.26 & 73.41 $\pm$ 0.27 & 76.63 $\pm$ 0.16 & 69.48 $\pm$ 0.39 & 64.92 $\pm$ 0.18 & 65.98 $\pm$ 0.30 & 83.59 $\pm$ 0.25 & 80.77 $\pm$ 0.12 & 81.36 $\pm$ 0.11 & - \\
FedGNN  & 70.63 $\pm$ 0.83 & 61.38 $\pm$ 2.33 & 56.91 $\pm$ 0.82 & 68.72 $\pm$ 0.39 & 59.98 $\pm$ 1.52 & 58.98 $\pm$ 0.98 & 84.25 $\pm$ 0.07 & 82.02 $\pm$ 0.22 & 81.85	$\pm$ 0.10 & - \\
FedSage+ & 77.52 $\pm$ 0.46 & 51.99 $\pm$ 0.42 & 55.48 $\pm$ 11.5 & 68.75 $\pm$ 0.48 & 65.97 $\pm$ 0.02 & 65.93 $\pm$ 0.30 & 82.77 $\pm$ 0.08 & 82.14 $\pm$ 0.11 & 80.31 $\pm$ 0.68 & - \\ 

\midrule

FED-PUB (Ours)    & \textbf{79.60} $\pm$ 0.12 & \textbf{75.40} $\pm$ 0.54 & \textbf{77.84} $\pm$ 0.23 & \textbf{70.58} $\pm$ 0.20 & \textbf{68.33} $\pm$ 0.45 & \textbf{69.21} $\pm$ 0.30 & \textbf{85.70} $\pm$ 0.08 & \textbf{85.16} $\pm$ 0.10 & \textbf{84.84} $\pm$ 0.12 & - \\

\midrule
\midrule

& \multicolumn{3}{c}{\bf Amazon-Computer} & \multicolumn{3}{c}{\bf Amazon-Photo} & \multicolumn{3}{c}{\bf ogbn-arxiv} & \textbf{All} \\
\cmidrule(l{2pt}r{2pt}){2-4} \cmidrule(l{2pt}r{2pt}){5-7} \cmidrule(l{2pt}r{2pt}){8-10} \cmidrule(l{2pt}r{2pt}){11-11}
\textbf{Methods} & \textbf{10 Clients} & \textbf{30 Clients} & \textbf{50 Clients} & \textbf{10 Clients} & \textbf{30 Clients} & \textbf{50 Clients} & \textbf{10 Clients} & \textbf{30 Clients} & \textbf{50 Clients} & \textbf{Avg.} \\
\midrule

Local   & 88.50 $\pm$ 0.20 & 86.66 $\pm$ 0.00 & 87.04 $\pm$ 0.02 & 92.17 $\pm$ 0.12 & 90.16 $\pm$ 0.12 & 90.42 $\pm$ 0.15 & 62.52 $\pm$ 0.07 & 61.32 $\pm$ 0.04 & 60.04 $\pm$ 0.04 & 76.72 \\

\midrule

FedAvg  & 88.99 $\pm$ 0.19 & 83.37 $\pm$ 0.47 & 76.34 $\pm$ 0.12 & 92.91 $\pm$ 0.07 & 89.30 $\pm$ 0.22 & 74.19 $\pm$ 0.57 & 63.56 $\pm$ 0.02 & 59.72 $\pm$ 0.06 & 60.94 $\pm$ 0.24 & 73.38\\
FedProx & 88.84 $\pm$ 0.20 & 83.84 $\pm$ 0.89 & 76.60 $\pm$ 0.47 & 92.67 $\pm$ 0.19 & 89.17 $\pm$ 0.40 & 72.36 $\pm$ 2.06 & 63.52 $\pm$ 0.11 & 59.86 $\pm$ 0.16 & 61.12 $\pm$ 0.04 & 73.38 \\
FedPer  & 89.30 $\pm$ 0.04 & 87.99 $\pm$ 0.23 & 88.22 $\pm$ 0.27 & 92.88 $\pm$ 0.24 & 91.23 $\pm$ 0.16 & 90.92 $\pm$ 0.38 & 63.97 $\pm$ 0.08 & 62.29 $\pm$ 0.04 & 61.24 $\pm$ 0.11 & 78.42 \\
GCFL    & 89.01 $\pm$ 0.22 & 87.24 $\pm$ 0.09 & 87.02 $\pm$ 0.22 & 92.45 $\pm$ 0.10 & 90.58 $\pm$ 0.11 & 90.54 $\pm$ 0.08 & 63.24 $\pm$ 0.02 & 61.66 $\pm$ 0.10 & 60.32 $\pm$ 0.01 & 77.61\\
FedGNN  & 88.15 $\pm$ 0.09 & 87.00 $\pm$ 0.10 & 83.96 $\pm$ 0.88 & 91.47 $\pm$ 0.11 & 87.91 $\pm$ 1.34 & 78.90 $\pm$ 6.46 & 63.08 $\pm$ 0.19 & 60.09 $\pm$ 0.04 & 60.51 $\pm$ 0.11 & 73.66\\
FedSage+ & 89.24 $\pm$ 0.15 & 81.33 $\pm$ 1.20 & 76.72 $\pm$ 0.39 & 92.76 $\pm$ 0.05 & 88.69 $\pm$ 0.99 & 72.41 $\pm$ 1.36 & 63.24 $\pm$ 0.02 & 59.90 $\pm$ 0.12 & 60.95 $\pm$ 0.09 & 73.12 \\ 

\midrule

FED-PUB (Ours)    & \textbf{89.98} $\pm$ 0.08 & \textbf{89.15} $\pm$ 0.06 & \textbf{88.76} $\pm$ 0.14 & \textbf{93.22} $\pm$ 0.07 & \textbf{92.01} $\pm$ 0.07 & \textbf{91.71} $\pm$ 0.11 & \textbf{64.18} $\pm$ 0.04 & \textbf{63.34} $\pm$ 0.12 & \textbf{62.55} $\pm$ 0.12 & \textbf{79.53} \\

\bottomrule

\end{tabular}
}
\end{table*}

%% file: figures/fig_table_1.tex
\begin{figure*}
    \small
    \centering
    \vspace{-0.1in}
    \begin{minipage}{0.16\linewidth}
        \includegraphics[width=1.0\textwidth]{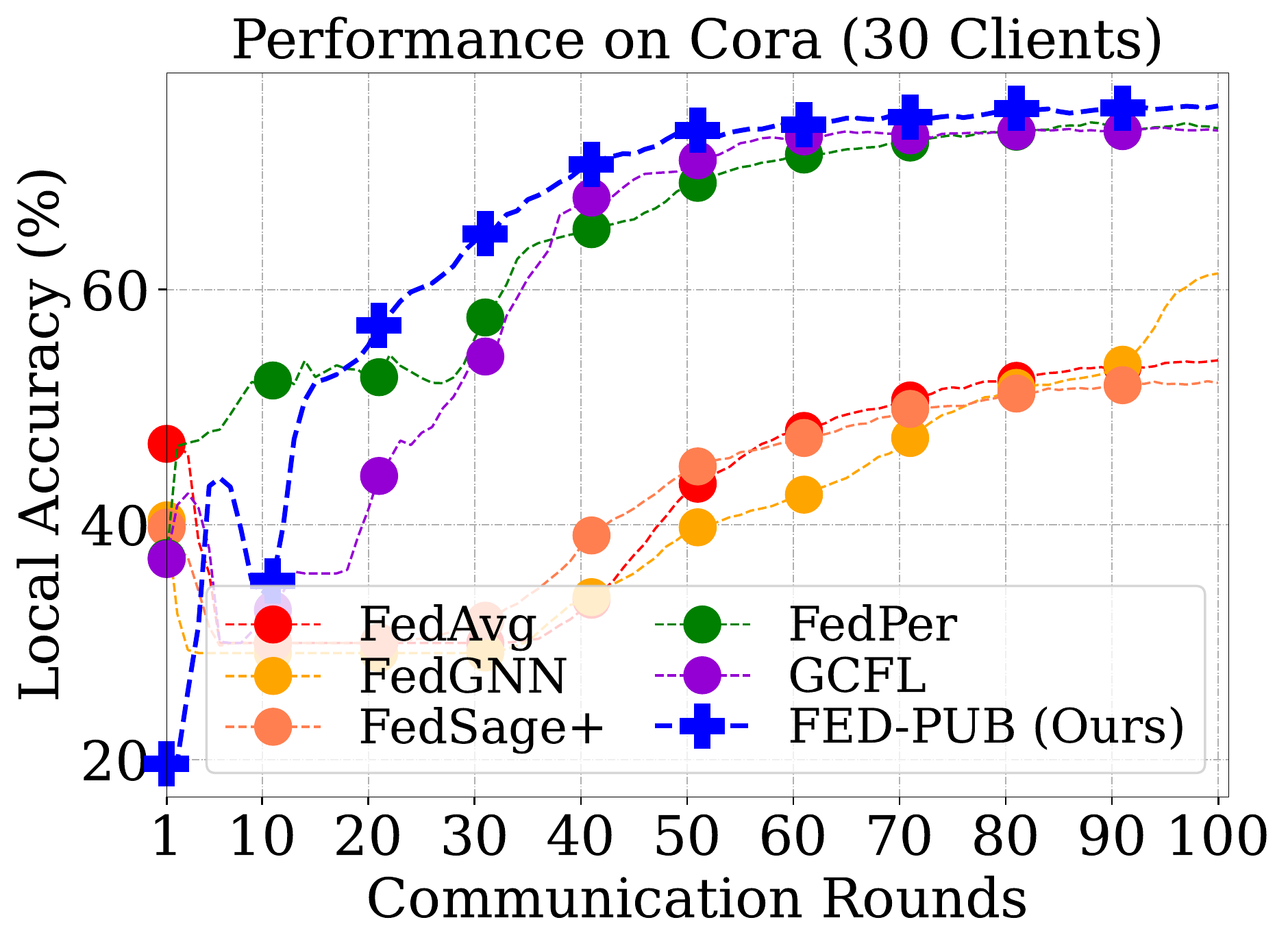}
        \vspace{-0.2in}
        \subcaption{\scriptsize Cora}
    \end{minipage}
    \begin{minipage}{0.16\linewidth}
        \includegraphics[width=1.0\textwidth]{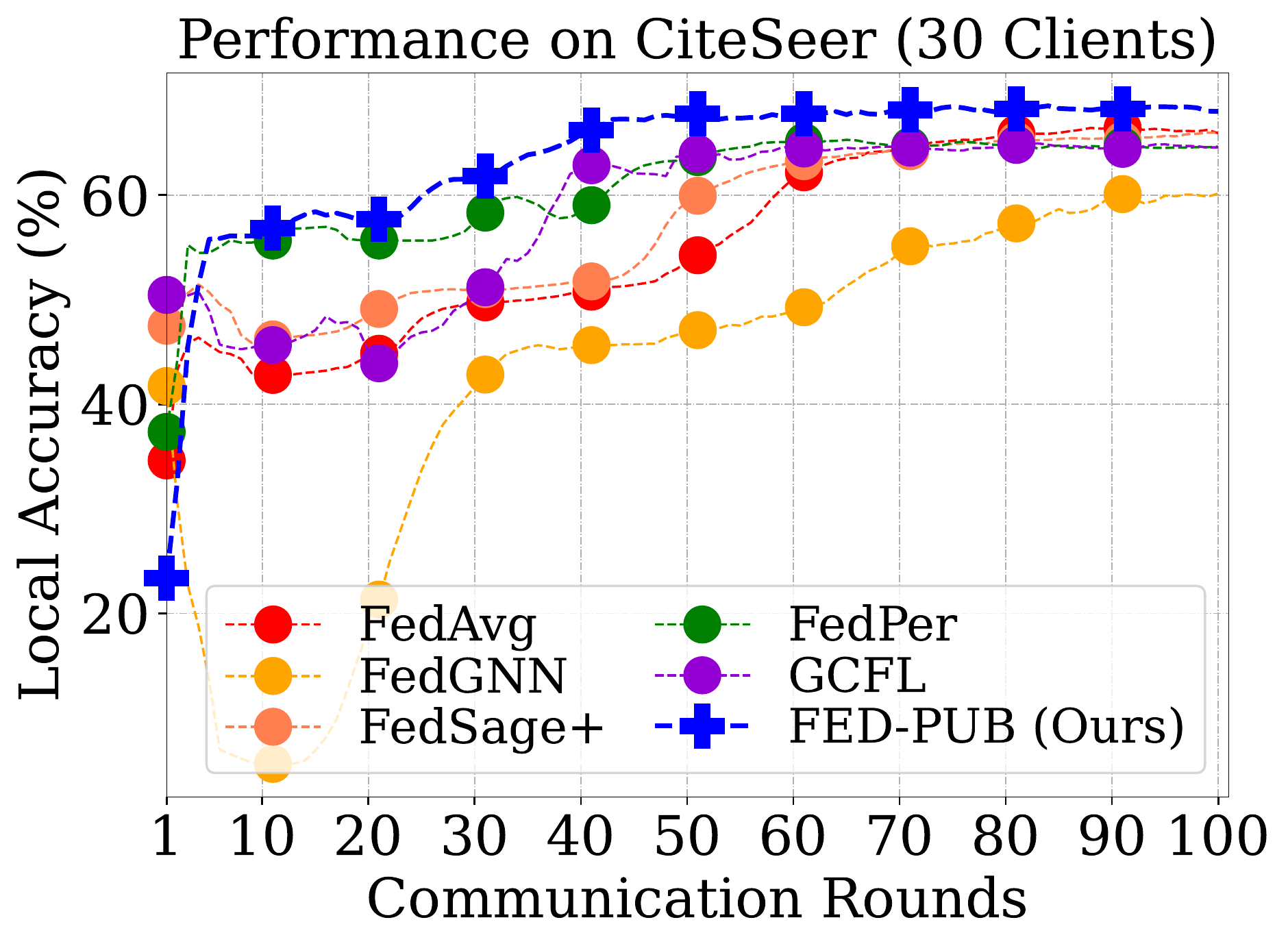}
        \vspace{-0.2in}
        \subcaption{\scriptsize CiteSeer}
    \end{minipage}
    \begin{minipage}{0.16\linewidth}
        \includegraphics[width=1.0\textwidth]{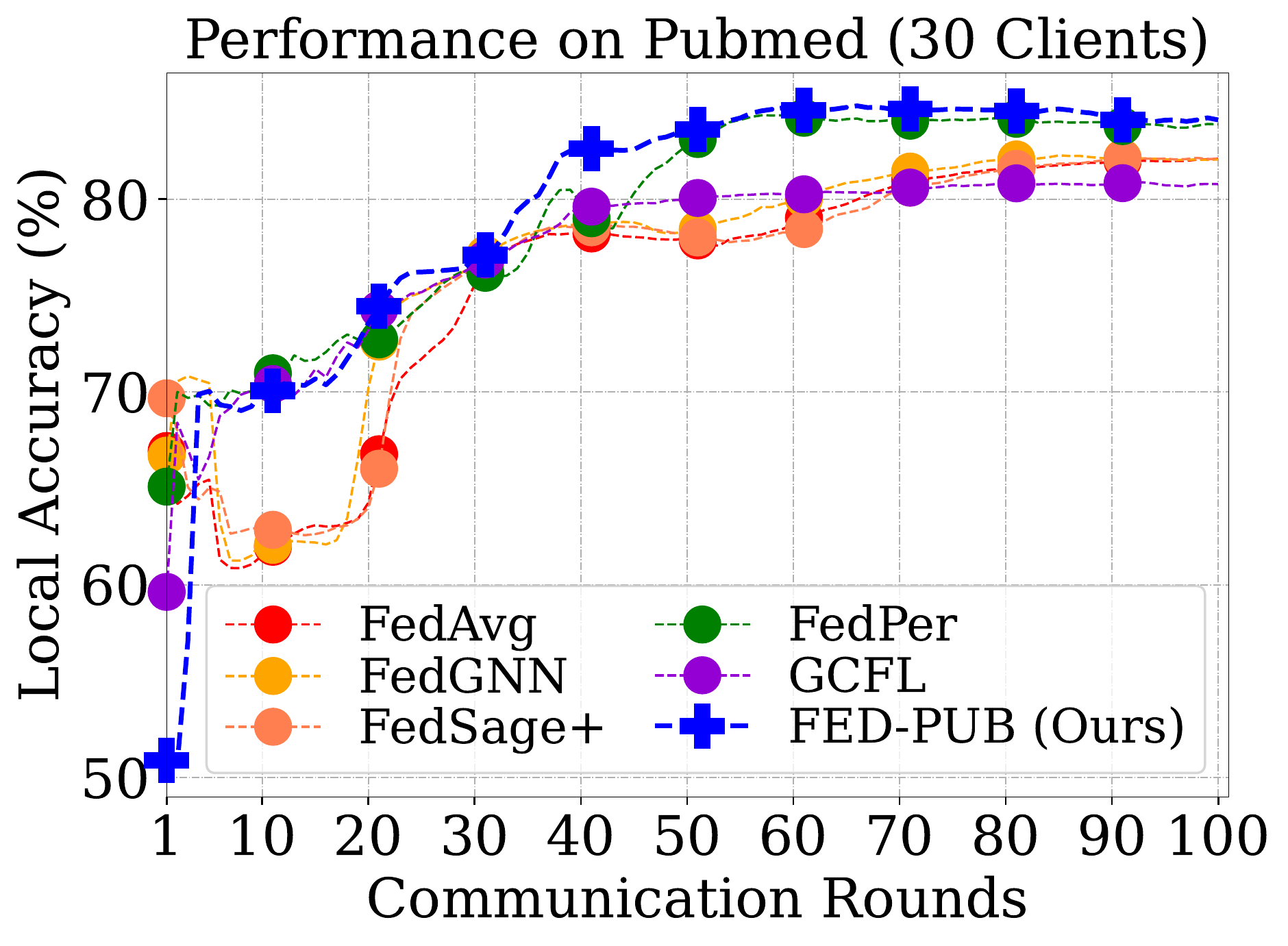}
        \vspace{-0.2in}
        \subcaption{\scriptsize Pubmed}
    \end{minipage}
    \begin{minipage}{0.16\linewidth}
        \includegraphics[width=1.0\textwidth]{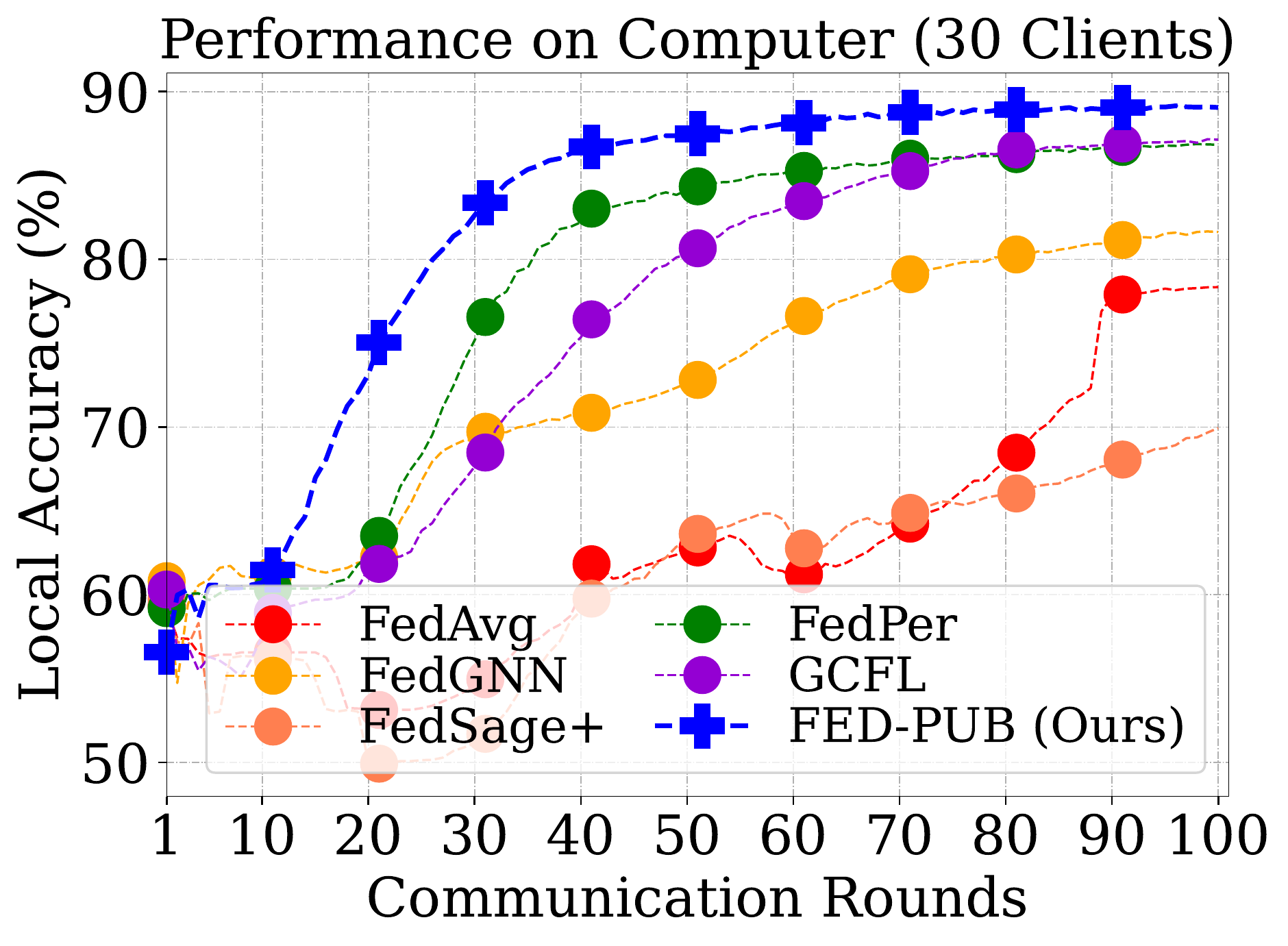}
        \vspace{-0.2in}
        \subcaption{\scriptsize Computer}
    \end{minipage}
    \begin{minipage}{0.16\linewidth}
        \includegraphics[width=1.0\textwidth]{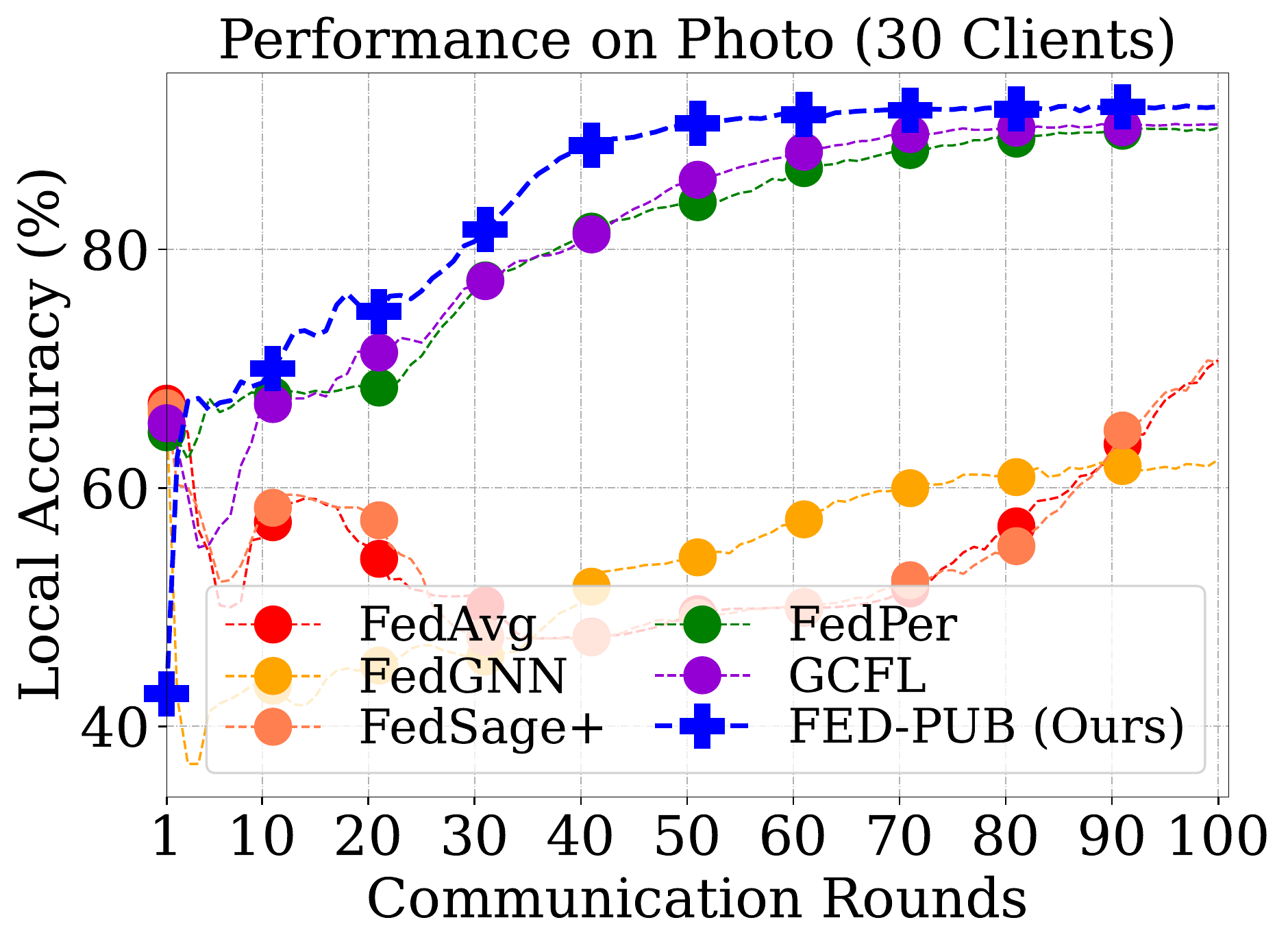}
        \vspace{-0.2in}
        \subcaption{\scriptsize Photo}
    \end{minipage}
    \begin{minipage}{0.16\linewidth}
        \includegraphics[width=1.0\textwidth]{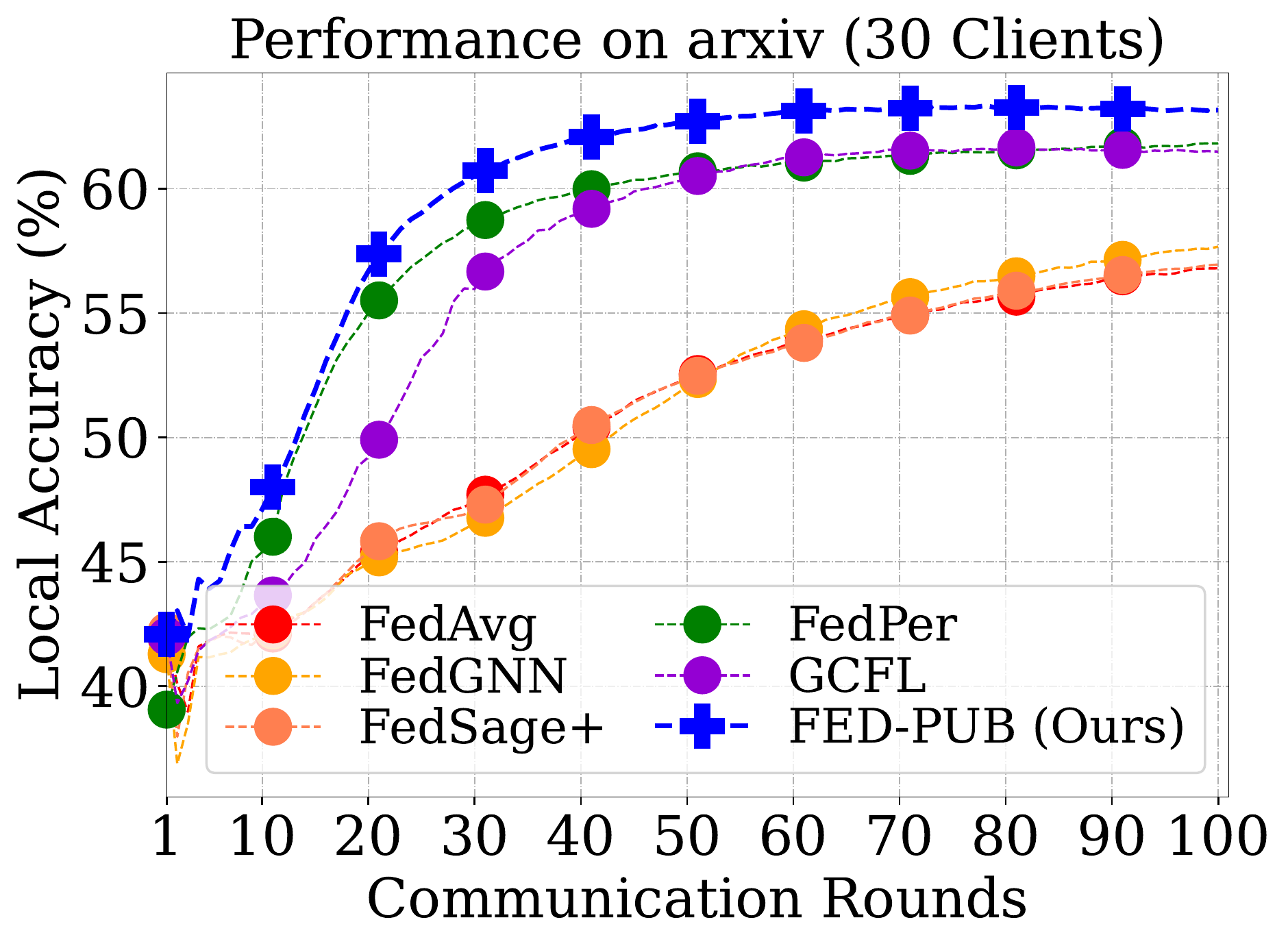}
        \vspace{-0.2in}
        \subcaption{\scriptsize ogbn-arxiv}
    \end{minipage}
    \vspace{-0.13in}
    \caption{\textbf{Convergence plots for the overlapping node scenario.} We visualize accuracies on 100 communication rounds with 30 clients.}
    \label{fig:convergence:overlap}
    \vspace{-0.125in}
\end{figure*}

%% file: sections/6_experiment.tex
\section{Experiments}
\label{experiments}

We validate our FED-PUB on six datasets under both the overlapping and disjoint subgraph scenarios mainly on node classification and additionally on link prediction tasks.

\subsection{Experimental Setups}
\label{experiments:datasets}

\paragraph{Datasets} Following the experimental setup from~\citet{FedSage}, we construct distributed subgraphs by dividing the dataset into a certain number of participants, as each FL participant has a subgraph that is a part of an original graph. Specifically, we use six datasets: Cora, CiteSeer, Pubmed and ogbn-arxiv for citation graphs~\citep{planetoid, ogb}; Computer and Photo for product graphs~\citep{amazon, amnazonsubset}. We then divide their graphs using the METIS graph partitioning algorithm~\citep{Karypis95metis}. Note that, unlike the Louvain algorithm~\citep{blondel2008fast}, used in~\citet{FedSage}, that requires to further merge partitioned subgraphs into particular numbers of subgraphs since it cannot specify the number of subsets (i.e., clients for FL), the METIS algorithm can specify the number of subsets, thus making more reasonable experimental settings (See Appendix~\ref{appendix:Louvain}). For the non-overlapping node scenario where there are no duplicate nodes between subgraphs, we use the output from the METIS as it provides non-overlapping partitions. For the overlapping scenario where nodes are duplicated among subgraphs, we randomly sample the subsets (i.e., subgraphs) of the partitioned graph multiple times. See Appendix~\ref{appendix:sub:data} for more details.

\input{tables/main_nonoverlap}
\input{figures/fig_table_2}

\vspace{-0.05in}
\paragraph{Baselines and Our Model} \textbf{1) FedAvg}~\citep{McMahan2017CommunicationEfficientLO} and \textbf{2) FedProx}~\citep{li2020federated}: The most popular FL baselines. \textbf{3) FedPer}~\citep{arivazhagan2019federated}: A personalized FL baseline without sharing personalized layers. \textbf{4) FedGNN (FedPerGNN)}\footnote{FedGNN is extended to FedPerGNN, and their core algorithms of averaging gradients of all clients are exactly the same.}~\citep{FedGNN, FedPerGNN} and \textbf{5) FedSage+}~\citep{FedSage}: Subgraph FL baselines which we mainly target. \textbf{6) GCFL}~\citep{GCFL}: A graph FL baseline which targets completely disjoint graphs for graph-level FL as in clustered FL~\citep{sattler2020clustered}, adopted for subgraph FL. \textbf{7) Local}: A baseline that locally trains models without weight sharing. \textbf{8) FED-PUB}: Our personalized subgraph FL, which includes similarity matching and weight masking. See Appendix~\ref{appendix:sub:model} for details.

\vspace{-0.05in}
\paragraph{Implementation Details} We use two layer GCNs~\citep{GCN} as the base GNN for all models. We perform FL over 100 communication rounds for Cora, CiteSeer and Pubmed datasets, while 200 rounds for Computer, Photo and arxiv, considering the dataset size. The local training epoch is selected in the range of $\left\{ 1, 2, 3 \right\}$ depending on the dataset size (e.g., Computer is three while CiteSeer is one)\footnote{We found that communication rounds and local epochs are important factors to prevent overfitting of models.}. We use the Adam optimizer~\cite{Adam} for optimization. We measure the node classification accuracy on subgraphs on the client-side, and then average the performances across all clients. We provide more details in Appendix~\ref{appendix:sub:imple}.

\subsection{Experimental Results}

\paragraph{Main Results} 
Table~\ref{tab:main:overlap} shows node classification results on the overlapping subgraph scenario, in which our FED-PUB outperforms all baselines with statistical significance ($p > 0.05$). Specifically, while FedGNN and FedSage+ are two pioneer works for subgraph FL, they are inferior to personalized FL methods including ours, especially at the larger number of clients. This is surprising as they share node information between clients for handling the missing edge problem, yet we suppose such inferior performance comes from naive averaging of local weights without consideration of community structures. While personalized FL baselines including FedPer and GCFL show decent performance by alleviating the knowledge collapse issue between subgraphs with local parameterization or clustering, they still underperform ours as they are not concerned with aggregation between similar subgraphs that form a community (i.e., GCFL uses a bi-partitioning scheme, which iteratively divides a group of subgraphs within the same community into two disjoint sets). We then further conduct experiments on the disjoint subgraph scenario (i.e., non-overlapping scenario), which makes the subgraph FL problem more heterogeneous. As shown in Table~\ref{tab:main:nonoverlap}, the proposed FED-PUB consistently outperforms all existing baselines in such a challenging scenario, demonstrating the efficacy of ours.

\input{figures/heatmap}
\input{figures/neighbor}

\vspace{-0.05in}
\paragraph{Fast Convergence} 
As shown in Figure~\ref{fig:convergence:overlap} and \ref{fig:convergence:nonoverlap}, our FED-PUB converges rapidly, compared against baselines. We conjecture that this is because ours can accurately identify subgraphs forming the community and then share weights substantially across them for promoting the joint improvement. Also, ours can mask out subgraph-irrelevant weights received from the server for localization to local subgraphs. We demonstrate those two points in the next two paragraphs.

\vspace{-0.05in}
\paragraph{Community Detection}
We aim to show whether FED-PUB can group subgraphs comprising a community during personalized weight aggregation. Note that, if two different subgraphs have many missing edges or have similar label distributions, we usually consider those two as within the same community~\citep{Community1, Community, Community2}. Thereby, as shown in Figure~\ref{fig:heatmap} (a) and (b), there are four different communities by the interval of five, and the last two communities further comprise a larger community. Then, as shown in Figure~\ref{fig:heatmap} (c) and (d), our FED-PUB detects obvious four communities at the first few rounds, and then captures the larger yet somewhat less-obvious community consisting of two smaller communities.

\vspace{-0.05in}
\paragraph{Ablation Study} 
To analyze the contribution of each component, we conduct ablation studies. As shown in Figure~\ref{fig:ablation}, we observe that each of our similarity matching and weight masking schemes significantly improves the performance from the naive FedAvg, while the performance is much improved when using both together. However, the benefit from each component is different between overlapping and non-overlapping scenarios. In particular, in the former scenario where a group of densely overlapped subgraphs comprises an obvious community, similarity matching for discovering community structures is more beneficial since capturing the community would promote the joint improvement of subgraphs belonging to the same community. However, in the non-overlapping scenario, two individual subgraphs become more heterogeneous, thus selectively using the aggregated parameters from the server with personalized weight masks improves the performance substantially. See Appendix~\ref{appendix:heterogeneity} for more discussions on heterogeneity with local masks.

\vspace{-0.05in}
\paragraph{Communication Efficiency}
Another notable advantage of using sparse masks is that we can reduce the communication costs at every FL round, as well as the model size for faster runtime. In particular, as reported in Table~\ref{fig:efficiency}, existing subgraph FL methods require more than two times larger communications costs, measured by adding both the client-to-server and server-to-client costs, compared against the naive FedAvg. This is because they require to transfer additional node information between clients for estimating the probable nodes on each subgraph. In contrast, our FED-PUB has significantly lower communication costs and lower model sizes by using sparse masks on model weights: transmitting and running with only the partial parameters not sparsified at the client. Further, we can manage the trade-off between the model sparsity and the performance by controlling the hyperparameter for sparsity regularization, $\lambda_1$ (See Appendix~\ref{appendix:lambda_analysis} for more results on hyperparameters).

\vspace{-0.05in}
\paragraph{Varying Local Epochs} 
As shown in Figure~\ref{fig:epoch}, when we increase the number of communication rounds and the local steps, local models diverge to their local subgraphs (i.e., overfitting), due to the small number of training instances and the direct connection between training and test nodes: struggling to generalize to the test instances. However, our FED-PUB with a proximal term in equation~\ref{eq:loss} alleviates this issue, therefore, maintaining the highest local performance. Notably, the performance with five local epochs is inferior to the performance of one epoch, which indicates that increasing the local epochs does not always bring advantages, and properly tuning them is important for subgraph FL.

\input{tables/link_prediction.tex}

\vspace{-0.05in}
\paragraph{Handling Missing Edges}
The missing edge problem, where two interconnected subgraphs cannot directly share the knowledge between them due to a lack of edges, is a unique challenge in subgraph FL (See Appendix~\ref{appendix:results:missingedge} for more discussions). To tackle this, existing subgraph FL explicitly augments nodes and edges to enable the information flow between interconnected subgraphs. Meanwhile, our FED-PUB implicitly shares weights across similar subgraphs within the same community. To measure their efficacies, we evaluate the performance on the neighboring subgraph, which has the most missing edges to the local subgraph for each client, based on its local model weight. Specifically, in Figure~\ref{fig:neighbor}, (Neighbor) denotes the subgraph performance evaluated by its neighbor model, while (Local) denotes the performance by its own local model. Therefore, high performances on the (Neighbor) measure indicate two associated subgraphs share meaningful knowledge despite not having actual edges between them, thereby alleviating the missing edge problem. As shown in Figure~\ref{fig:neighbor}, FED-PUB achieves superior performance on neighboring subgraphs against subgraph FL baselines. This result verifies that our FED-PUB has an advantage on the missing edge problem by sharing meaningful knowledge between subgraphs having potential missing edges without explicitly augmenting them.

\paragraph{Imbalanced Subgraphs}
As explained in Appendix~\ref{appendix:sub:data} and reported in Table~\ref{tab:appendix:data}, each subgraph is of similar size in our main experiments. However, real-world subgraphs might have variances in size; therefore, in Table~\ref{tab:imbalanced}, we further conduct experiments on the imbalance node scenarios where different subgraphs have different numbers of nodes. To do this, we create the imbalanced dataset by equally dividing the entire graph into several subgraphs and then merging some of them for making imbalanced subgraphs. More specifically, for our ten client setting in Table~\ref{tab:imbalanced}, we first partition an original graph into 20 subgraphs. Then, we merge each of the five, three, two, two, and two subgraphs into one larger subgraph. As shown in Table~\ref{tab:imbalanced}, our FED-PUB outperforms all baselines, which demonstrates the advantages of FED-PUB in the more realistic setting.

\paragraph{Link Prediction Results}
In addition to the extensive experiments on the node classification task, we further perform experiments on the link prediction task. In this link prediction task, we use the cross-entropy loss in Equation~\ref{eq:loss}, which is the same as the loss function for the node classification task yet the target value is binary. Further, during training, we sample negative edges with random sampling whose sizes are the same as the number of positive edges in the same batch. For evaluation, we measure the link prediction performance with ROC-AUC as a metric on all subgraphs, and then report their averaged result. Note that, for the other experimental setups, we follow the experimental settings of node classification tasks described in Section~\ref{appendix:sub:imple}. First of all, as shown in Figure~\ref{tab:link_prediction}, our FED-PUB consistently outperforms all baselines on the link prediction task similar to node classification results. We further visualize the convergence plots in Figure~\ref{fig:link_prediction}, to see whether our FED-PUB can still rapidly converge over the link prediction task. In Figure~\ref{fig:link_prediction}, we observe that FED-PUB converges rapidly, compared to baselines. These two results further demonstrate the applicability of FED-PUB to other subgraph tasks.

%% file: tables/main_nonoverlap.tex
\begin{table*}[t]
\caption{\small \textbf{Results on the non-overlapping node scenario.} The reported results are mean and standard deviation over three different runs. The statistically significant performances ($p > 0.05$) are emphasized in bold.}
\vspace{-0.1in}
\label{tab:main:nonoverlap}
\small
\centering
\resizebox{\textwidth}{!}{
\renewcommand{\arraystretch}{1.0}
\begin{tabular}{lccccccccca}
\toprule
& \multicolumn{3}{c}{\bf Cora} & \multicolumn{3}{c}{\bf CiteSeer} & \multicolumn{3}{c}{\bf Pubmed} & -  \\
\cmidrule(l{2pt}r{2pt}){2-4} \cmidrule(l{2pt}r{2pt}){5-7} \cmidrule(l{2pt}r{2pt}){8-10} \cmidrule(l{2pt}r{2pt}){11-11}
\textbf{Methods} & \textbf{5 Clients} & \textbf{10 Clients} & \textbf{20 Clients} & \textbf{5 Clients} & \textbf{10 Clients} & \textbf{20 Clients} & \textbf{5 Clients} & \textbf{10 Clients} & \textbf{20 Clients} & -  \\
\midrule

Local   & 81.30 $\pm$ 0.21 & 79.94 $\pm$ 0.24 & 80.30 $\pm$ 0.25 & 69.02 $\pm$ 0.05 & 67.82 $\pm$ 0.13 & 65.98 $\pm$ 0.17 & 84.04 $\pm$ 0.18 & 82.81 $\pm$ 0.39 & 82.65 $\pm$ 0.03 & - \\

\midrule

FedAvg  & 74.45 $\pm$ 5.64 & 69.19 $\pm$ 0.67 & 69.50 $\pm$ 3.58 & 71.06 $\pm$ 0.60 & 63.61 $\pm$ 3.59 & 64.68 $\pm$ 1.83 & 79.40 $\pm$ 0.11 & 82.71 $\pm$ 0.29 & 80.97 $\pm$ 0.26 & - \\
FedProx & 72.03 $\pm$ 4.56 & 60.18 $\pm$ 7.04 & 48.22 $\pm$ 6.81 & 71.73 $\pm$ 1.11 & 63.33 $\pm$ 3.25 & 64.85 $\pm$ 1.35 & 79.45 $\pm$ 0.25 & 82.55 $\pm$ 0.24 & 80.50 $\pm$ 0.25 & - \\
FedPer  & 81.68 $\pm$ 0.40 & 79.35 $\pm$ 0.04 & 78.01 $\pm$ 0.32 & 70.41 $\pm$ 0.32 & 70.53 $\pm$ 0.28 & 66.64 $\pm$ 0.27 & 85.80 $\pm$ 0.21 & 84.20 $\pm$ 0.28 & 84.72 $\pm$ 0.31 & - \\
GCFL    & 81.47 $\pm$ 0.65 & 78.66 $\pm$ 0.27 & 79.21 $\pm$ 0.70 & 70.34 $\pm$ 0.57 & 69.01 $\pm$ 0.12 & 66.33 $\pm$ 0.05 & 85.14 $\pm$ 0.33 & 84.18 $\pm$ 0.19 & 83.94 $\pm$ 0.36 & - \\
FedGNN  & 81.51 $\pm$ 0.68 & 70.12 $\pm$ 0.99 & 70.10 $\pm$ 3.52 & 69.06 $\pm$ 0.92 & 55.52 $\pm$ 3.17 & 52.23 $\pm$ 6.00 & 79.52 $\pm$ 0.23 & 83.25 $\pm$ 0.45 & 81.61 $\pm$ 0.59 & - \\
FedSage+ & 72.97 $\pm$ 5.94 & 69.05 $\pm$ 1.59 & 57.97 $\pm$ 12.6 & 70.74 $\pm$ 0.69 & 65.63 $\pm$ 3.10 & 65.46 $\pm$ 0.74 & 79.57 $\pm$ 0.24 & 82.62 $\pm$ 0.31 & 80.82 $\pm$ 0.25 & - \\ 

\midrule

FED-PUB (Ours)    & \textbf{83.70} $\pm$ 0.19 & \textbf{81.54} $\pm$ 0.12 & \textbf{81.75} $\pm$ 0.56 & \textbf{72.68} $\pm$ 0.44 & \textbf{72.35} $\pm$ 0.53 & \textbf{67.62} $\pm$ 0.12 & \textbf{86.79} $\pm$ 0.09 & \textbf{86.28} $\pm$ 0.18 & \textbf{85.53} $\pm$ 0.30 & - \\

\midrule
\midrule

& \multicolumn{3}{c}{\bf Amazon-Computer} & \multicolumn{3}{c}{\bf Amazon-Photo} & \multicolumn{3}{c}{\bf ogbn-arxiv} & \textbf{All} \\
\cmidrule(l{2pt}r{2pt}){2-4} \cmidrule(l{2pt}r{2pt}){5-7} \cmidrule(l{2pt}r{2pt}){8-10} \cmidrule(l{2pt}r{2pt}){11-11}
\textbf{Methods} & \textbf{5 Clients} & \textbf{10 Clients} & \textbf{20 Clients} & \textbf{5 Clients} & \textbf{10 Clients} & \textbf{20 Clients} & \textbf{5 Clients} & \textbf{10 Clients} & \textbf{20 Clients} & \textbf{Avg.} \\
\midrule

Local   & 89.22 $\pm$ 0.13 & 88.91 $\pm$ 0.17 & 89.52 $\pm$ 0.20 & 91.67 $\pm$ 0.09 & 91.80 $\pm$ 0.02 & 90.47 $\pm$ 0.15 & 66.76 $\pm$ 0.07 & 64.92 $\pm$ 0.09 & 65.06 $\pm$ 0.05 & 79.57\\

\midrule

FedAvg  & 84.88 $\pm$ 1.96 & 79.54 $\pm$ 0.23 & 74.79 $\pm$ 0.24 & 89.89 $\pm$ 0.83 & 83.15 $\pm$ 3.71 & 81.35 $\pm$ 1.04 & 65.54 $\pm$ 0.07 & 64.44 $\pm$ 0.10 & 63.24 $\pm$ 0.13 & 74.58\\
FedProx & 85.25 $\pm$ 1.27 & 83.81 $\pm$ 1.09 & 73.05 $\pm$ 1.30 & 90.38 $\pm$ 0.48 & 80.92 $\pm$ 4.64 & 82.32 $\pm$ 0.29 & 65.21 $\pm$ 0.20 & 64.37 $\pm$ 0.18 & 63.03 $\pm$ 0.04 & 72.84\\
FedPer  & 89.67 $\pm$ 0.34 & 89.73 $\pm$ 0.04 & 87.86 $\pm$ 0.43 & 91.44 $\pm$ 0.37 & 91.76 $\pm$ 0.23 & 90.59 $\pm$ 0.06 & 66.87 $\pm$ 0.05 & 64.99 $\pm$ 0.18 & 64.66 $\pm$ 0.11 & 79.94\\
GCFL    & 89.07 $\pm$ 0.91 & 90.03 $\pm$ 0.16 & 89.08 $\pm$ 0.25 & 91.99 $\pm$ 0.29 & 92.06 $\pm$ 0.25 & 90.79 $\pm$ 0.17 & 66.80 $\pm$ 0.12 & 65.09 $\pm$ 0.08 & 65.08 $\pm$ 0.04 & 79.90\\
FedGNN  & 88.08 $\pm$ 0.15 & 88.18 $\pm$ 0.41 & 83.16 $\pm$ 0.13 & 90.25 $\pm$ 0.70 & 87.12 $\pm$ 2.01 & 81.00 $\pm$ 4.48 & 65.47 $\pm$ 0.22 & 64.21 $\pm$ 0.32 & 63.80 $\pm$ 0.05 & 75.23\\
FedSage+ & 85.04 $\pm$ 0.61 & 80.50 $\pm$ 1.30 & 70.42 $\pm$ 0.85 & 90.77 $\pm$ 0.44 & 76.81 $\pm$ 8.24 & 80.58 $\pm$ 1.15 & 65.69 $\pm$ 0.09 & 64.52 $\pm$ 0.14 & 63.31 $\pm$ 0.20 & 73.47\\ 

\midrule

FED-PUB (Ours)    & \textbf{90.74} $\pm$ 0.05 & \textbf{90.55} $\pm$ 0.13 & \textbf{90.12} $\pm$ 0.09 & \textbf{93.29} $\pm$ 0.19 & \textbf{92.73} $\pm$ 0.18 & \textbf{91.92} $\pm$ 0.12 & \textbf{67.77} $\pm$ 0.09 & \textbf{66.58} $\pm$ 0.08 & \textbf{66.64} $\pm$ 0.12 & \textbf{81.59}\\

\bottomrule
\end{tabular}
}
\end{table*}

%% file: figures/fig_table_2.tex
\begin{figure*}
    \small
    \centering
    \vspace{-0.1in}
    \begin{minipage}{0.16\linewidth}
        \includegraphics[width=1.0\textwidth]{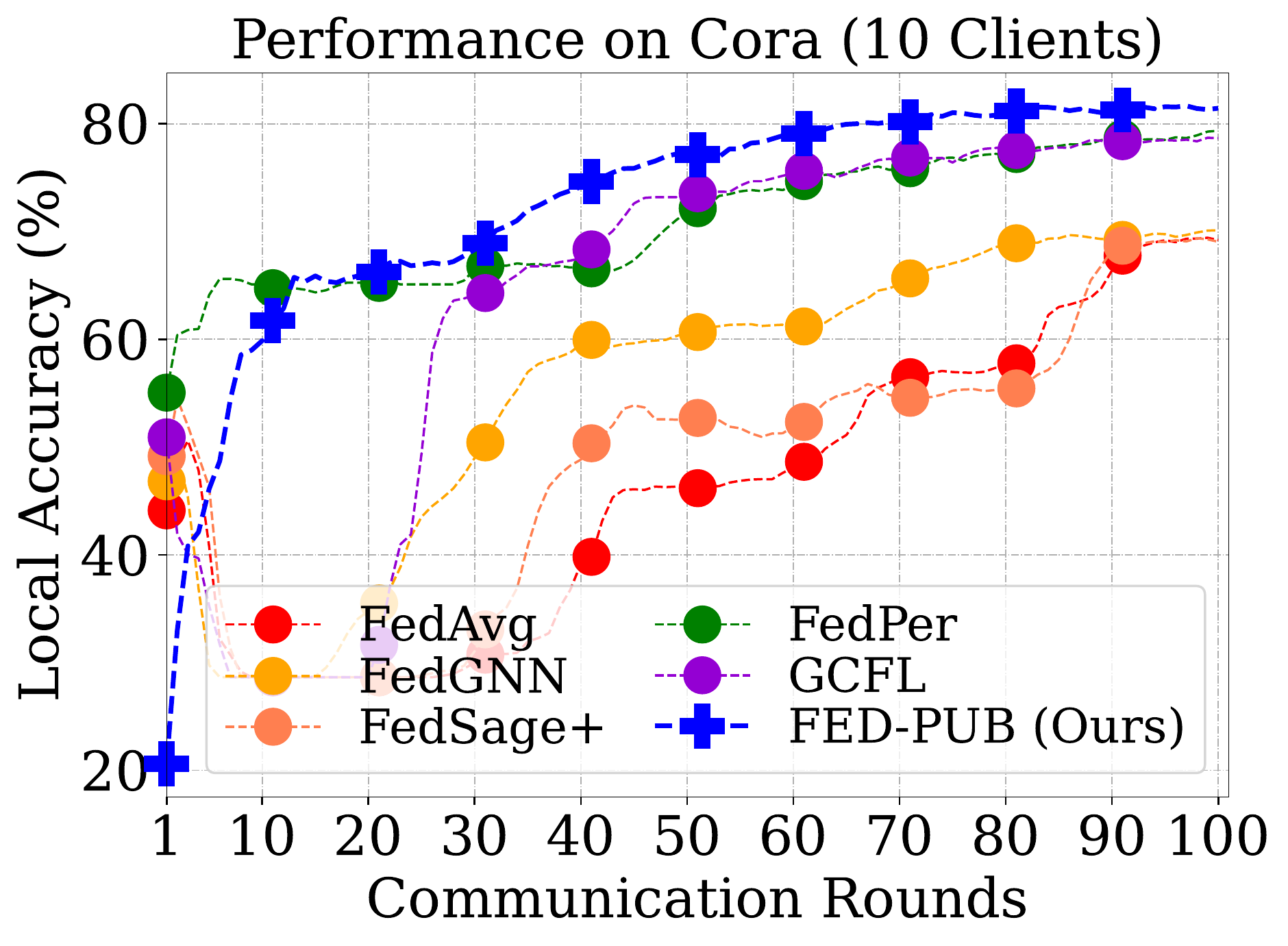}
        \vspace{-0.2in}
        \subcaption{\scriptsize Cora}
    \end{minipage}
    \begin{minipage}{0.16\linewidth}
        \includegraphics[width=1.0\textwidth]{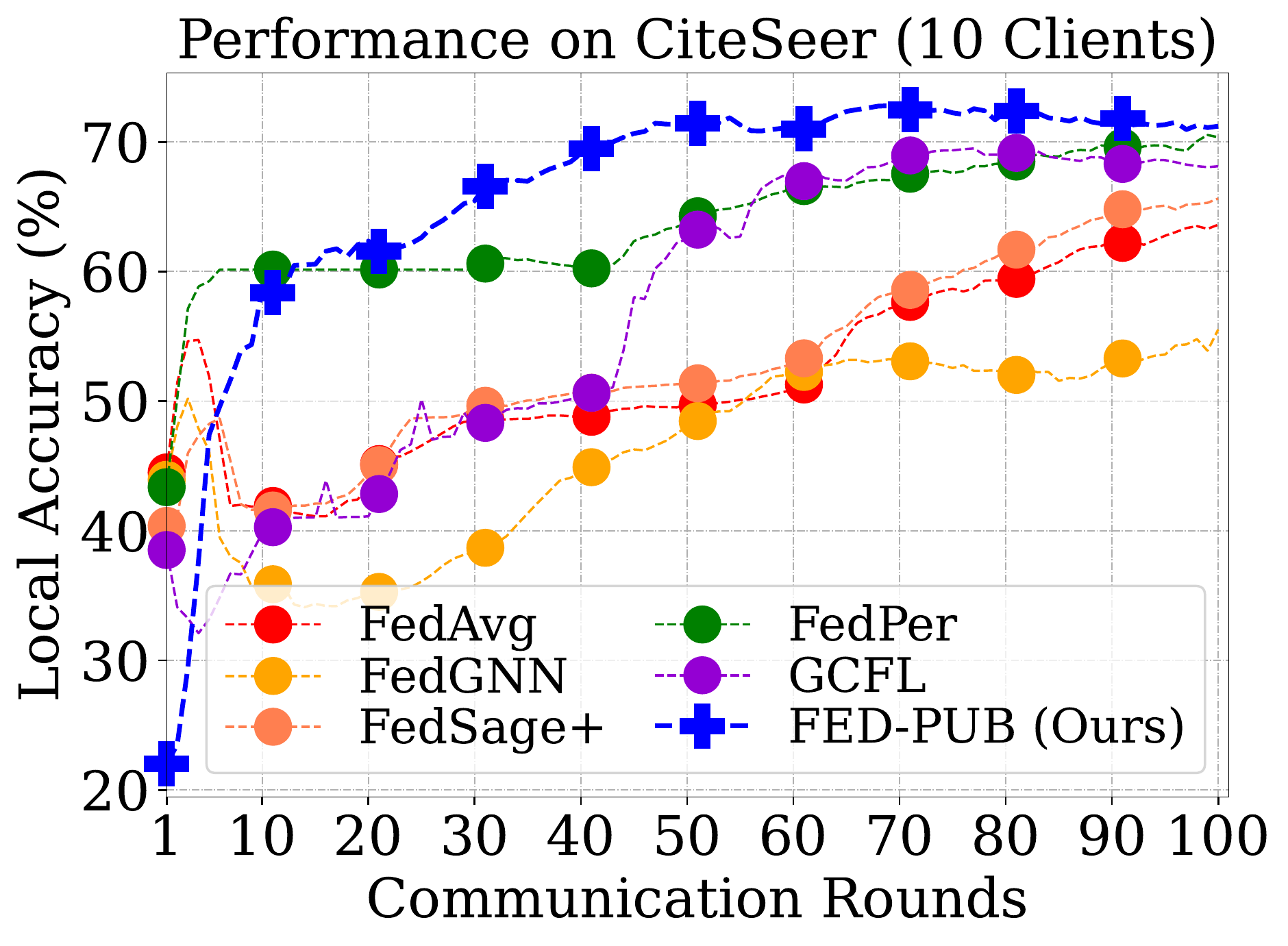}
        \vspace{-0.2in}
        \subcaption{\scriptsize CiteSeer}
    \end{minipage}
    \begin{minipage}{0.16\linewidth}
        \includegraphics[width=1.0\textwidth]{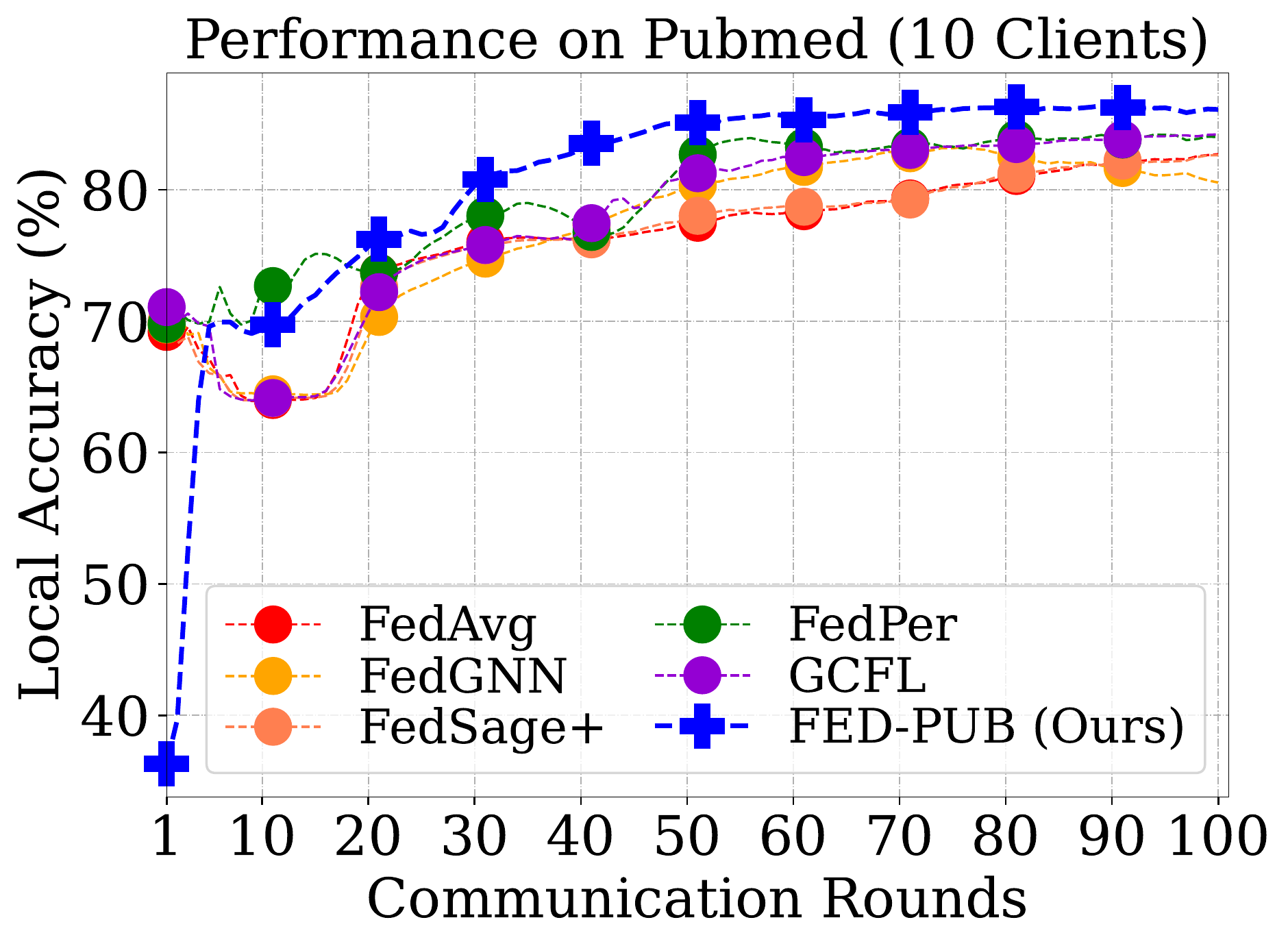}
        \vspace{-0.2in}
        \subcaption{\scriptsize Pubmed}
    \end{minipage}
    \begin{minipage}{0.16\linewidth}
        \includegraphics[width=1.0\textwidth]{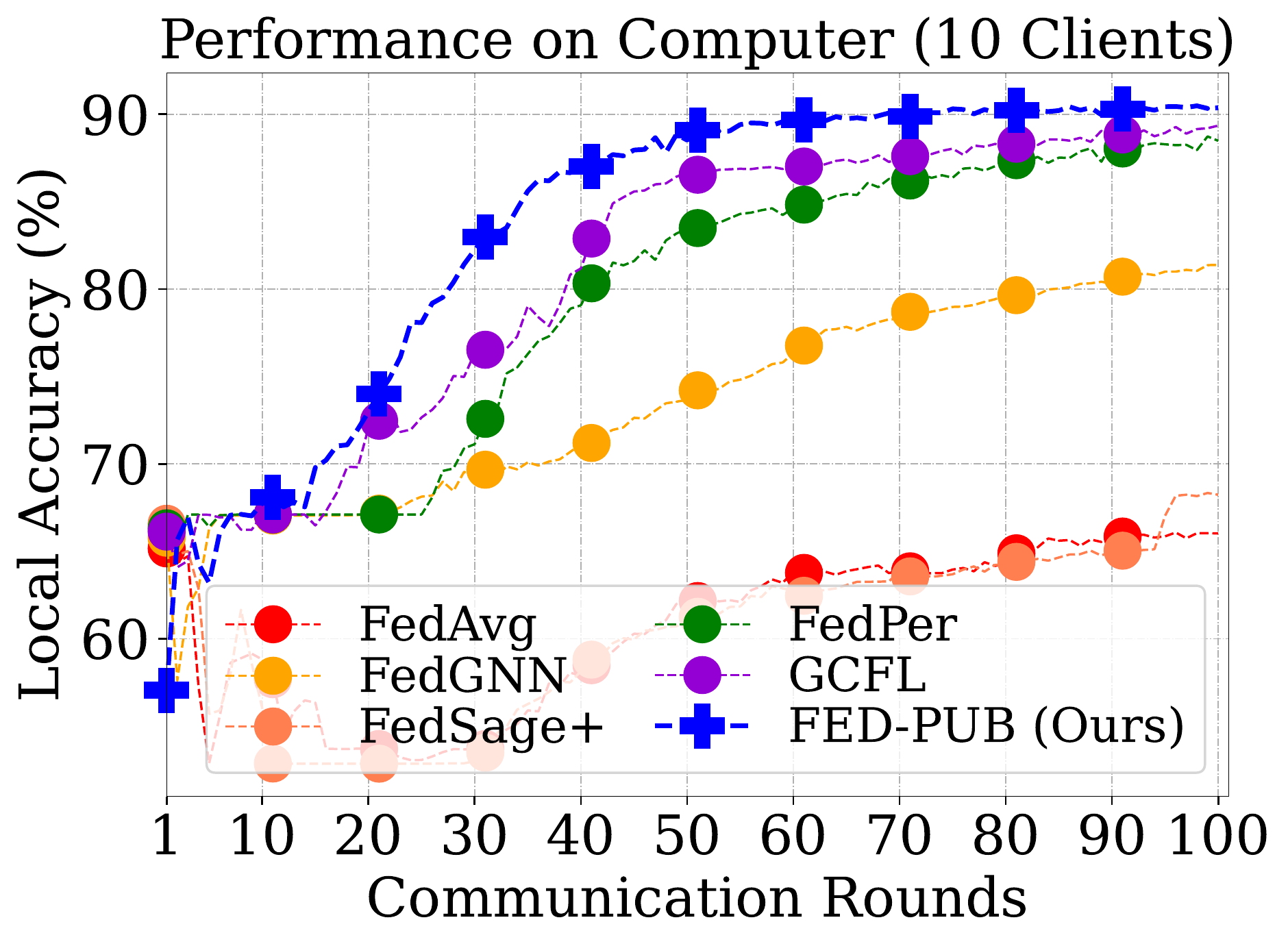}
        \vspace{-0.2in}
        \subcaption{\scriptsize Computer}
    \end{minipage}
    \begin{minipage}{0.16\linewidth}
        \includegraphics[width=1.0\textwidth]{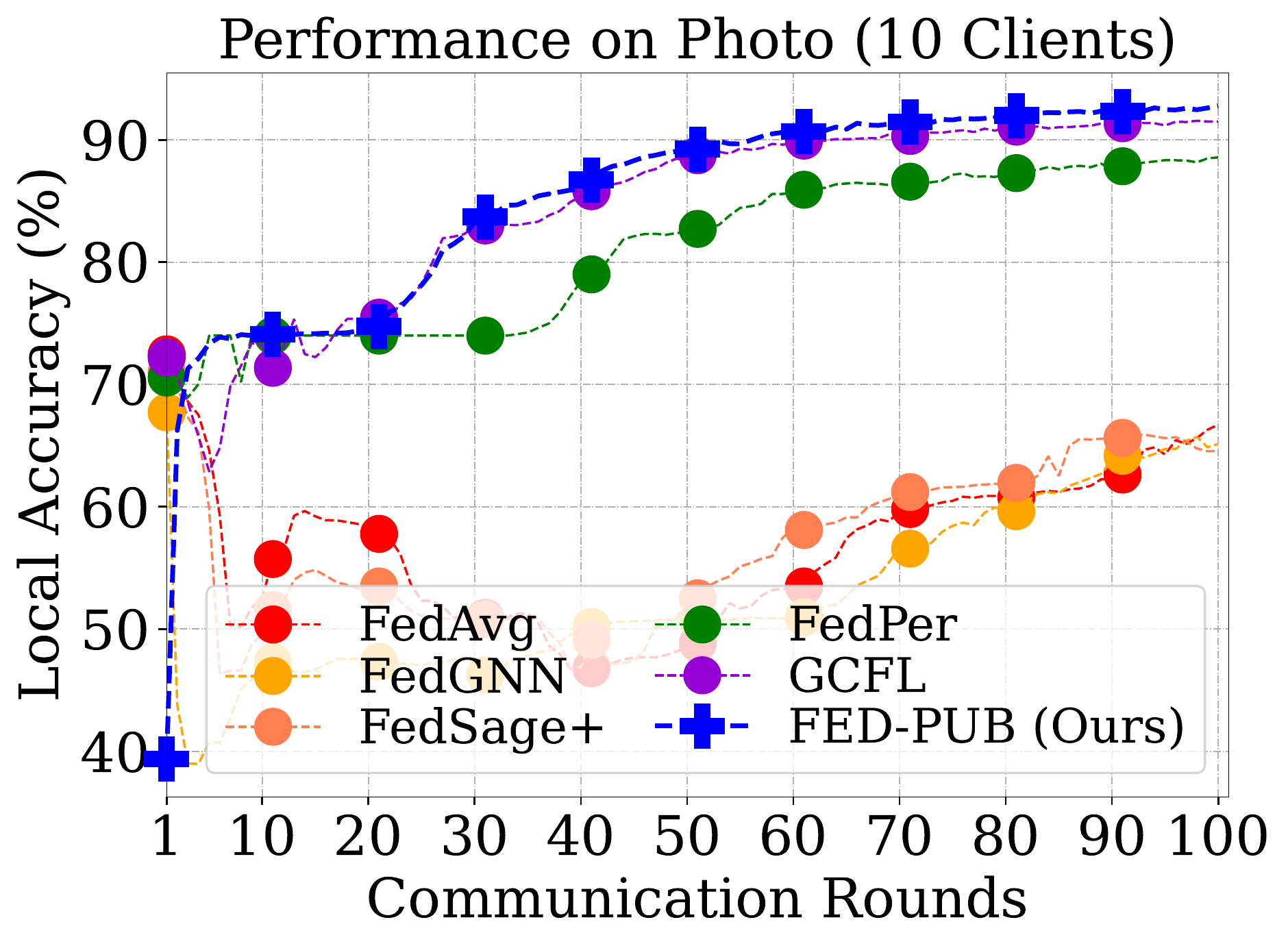}
        \vspace{-0.2in}
        \subcaption{\scriptsize Photo}
    \end{minipage}
    \begin{minipage}{0.16\linewidth}
        \includegraphics[width=1.0\textwidth]{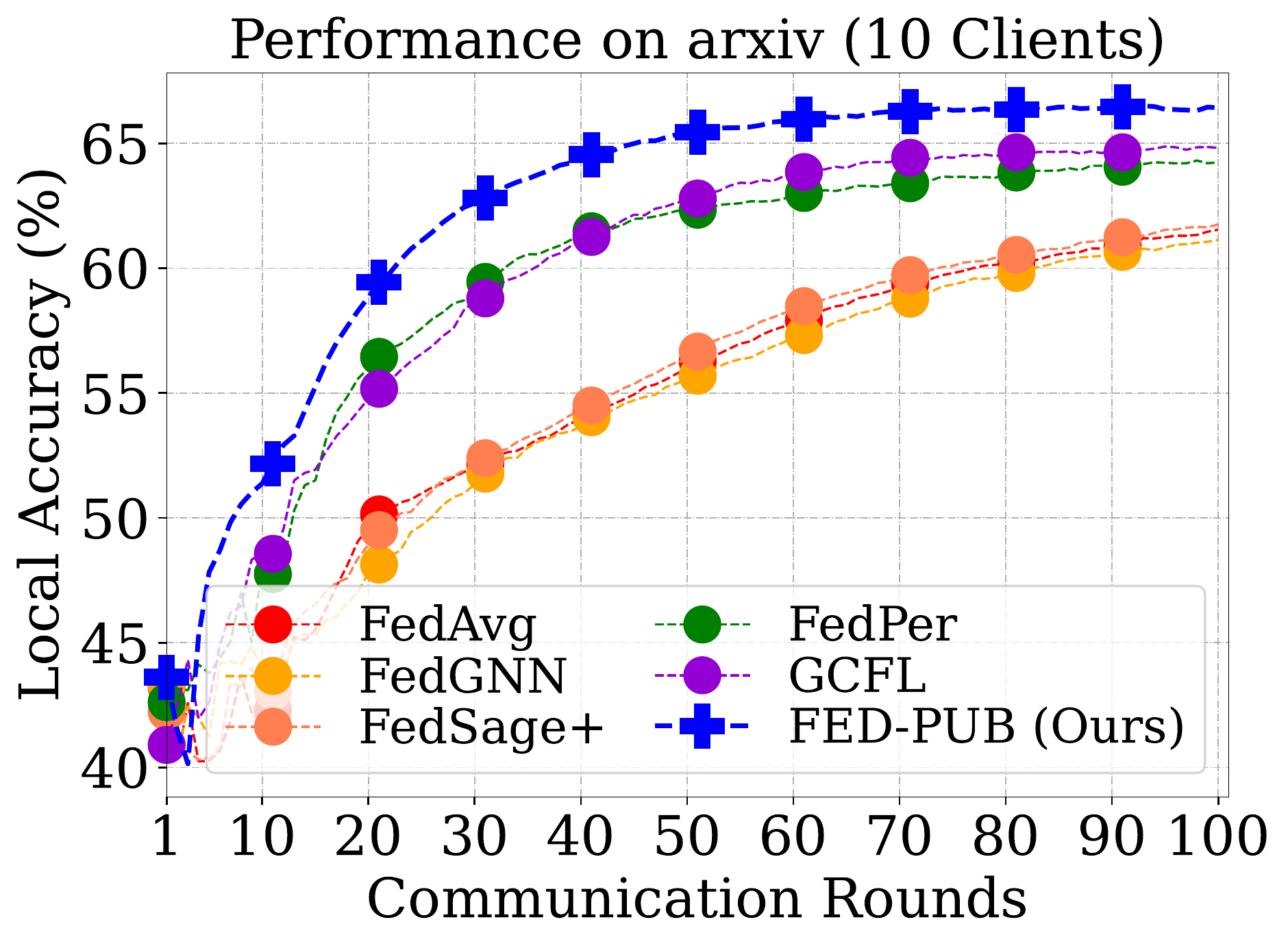}
        \vspace{-0.2in}
        \subcaption{\scriptsize ogbn-arxiv}
    \end{minipage}
    \vspace{-0.13in}
    \caption{\textbf{Convergence plots for the non-overlapping node scenario.} We visualize accuracies on 100 FL rounds with 10 clients.}
    \label{fig:convergence:nonoverlap}
    \vspace{-0.125in}
\end{figure*}

%% file: figures/heatmap.tex
\begin{figure*}[!t]
    \begin{minipage}{0.615\linewidth}
        \begin{minipage}{0.24\linewidth}
            \centerline{\includegraphics[width=0.975\linewidth]{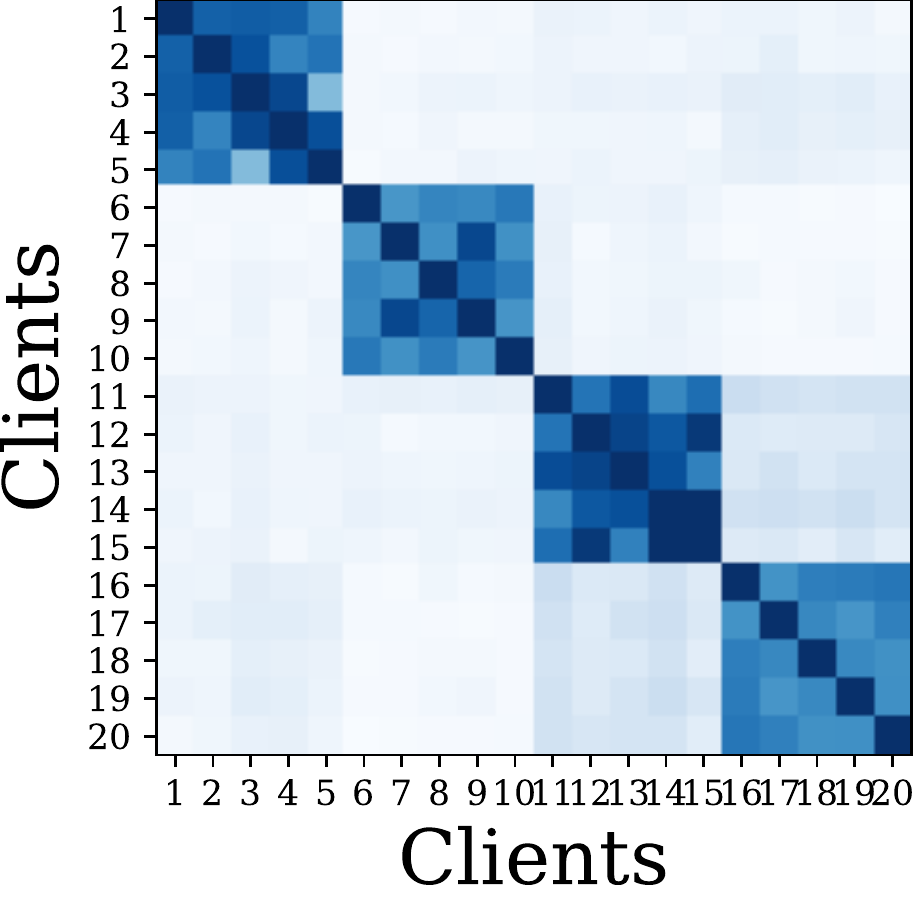}}
            \vspace{-0.05in}
            \subcaption{\scriptsize Missing edges}
        \end{minipage}
        \begin{minipage}{0.24\linewidth}
            \centerline{\includegraphics[width=0.975\linewidth]{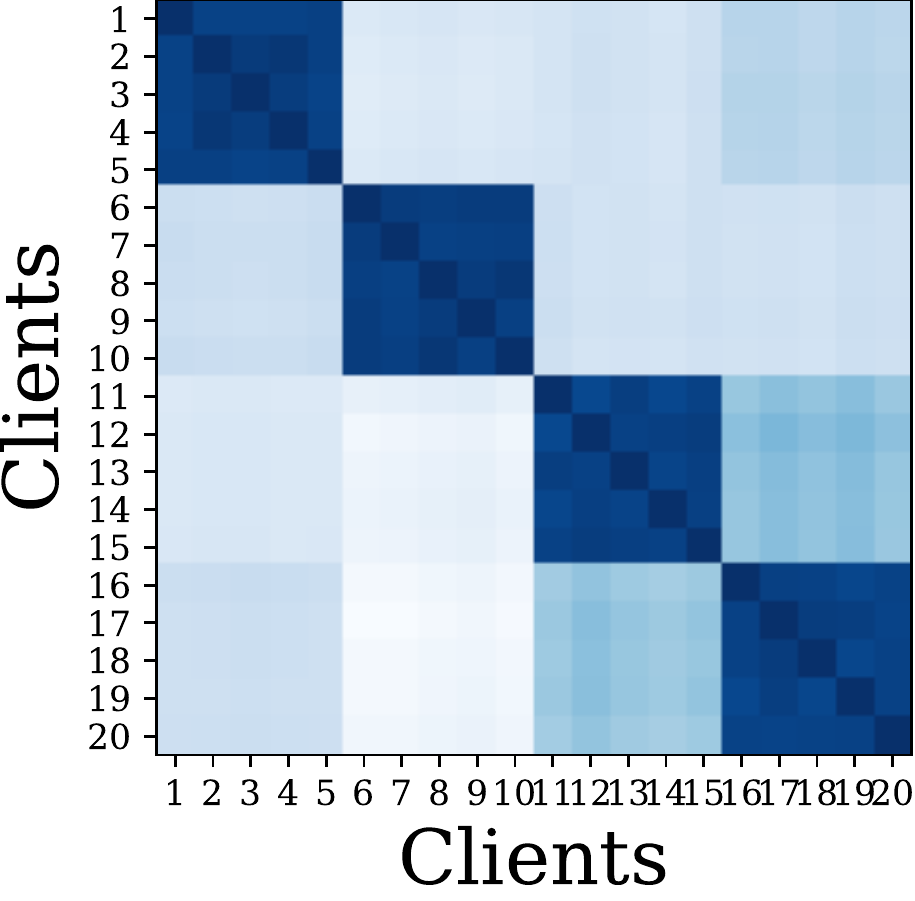}}
            \vspace{-0.05in}
            \subcaption{\scriptsize Label similarity}
        \end{minipage}
        \begin{minipage}{0.24\linewidth}
            \centerline{\includegraphics[width=0.975\linewidth]{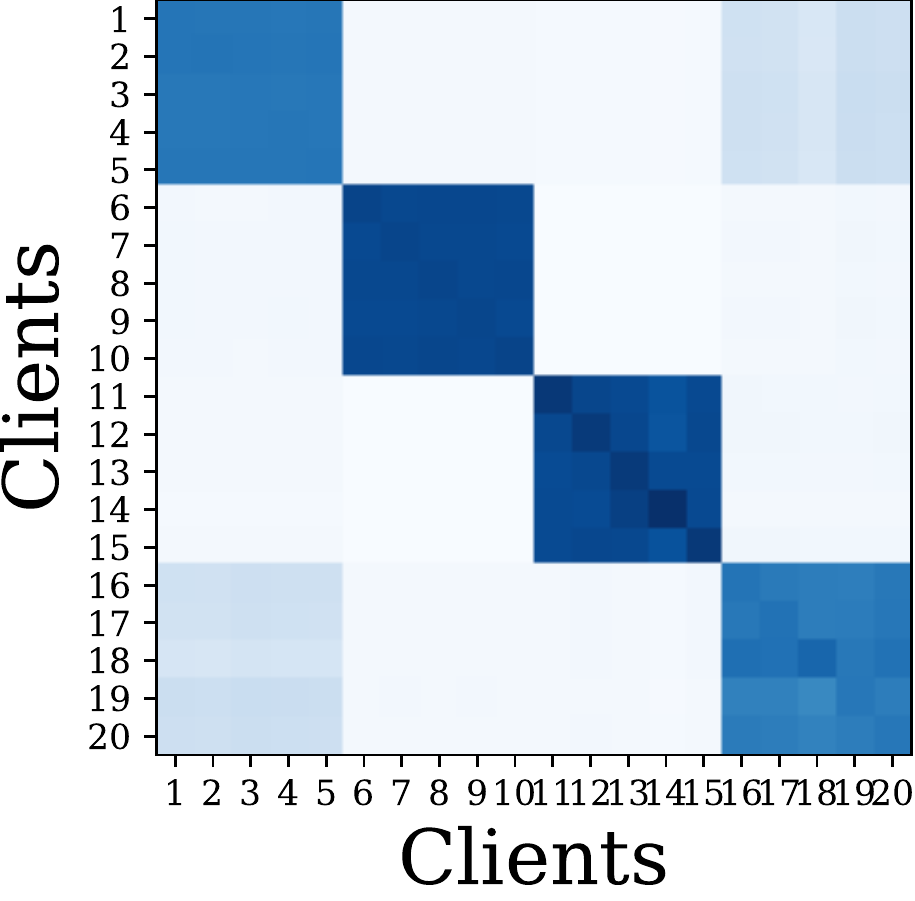}}
            \vspace{-0.05in}
            \subcaption{\scriptsize Round at 5}
        \end{minipage}
        \begin{minipage}{0.24\linewidth}
            \centerline{\includegraphics[width=0.975\linewidth]{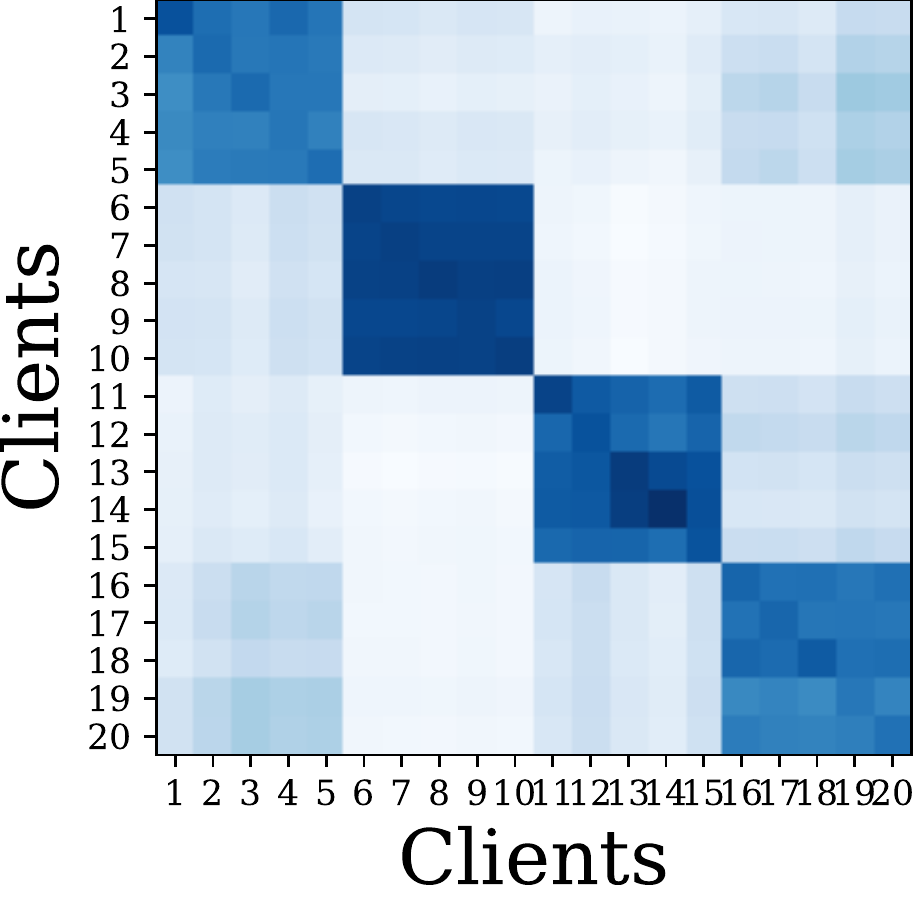}}
            \vspace{-0.05in}
            \subcaption{\scriptsize Round at 30}
        \end{minipage}
        \centering
        \vspace{-0.1in}
        \caption{\small \textbf{Heatmaps of community structures} on overlapping node scenario with Cora (20 clients). Darker color indicates many missing edges between subgraphs (a) or high similarities in labels (b). (c) and (d) are functional similarities by FED-PUB.}
        \label{fig:heatmap}
    \end{minipage}
    \hfill
    \begin{minipage}{0.36\linewidth}
        \vspace{-0.03in}
        \begin{minipage}{0.49\linewidth}
            \centering
            \includegraphics[width=1\linewidth]{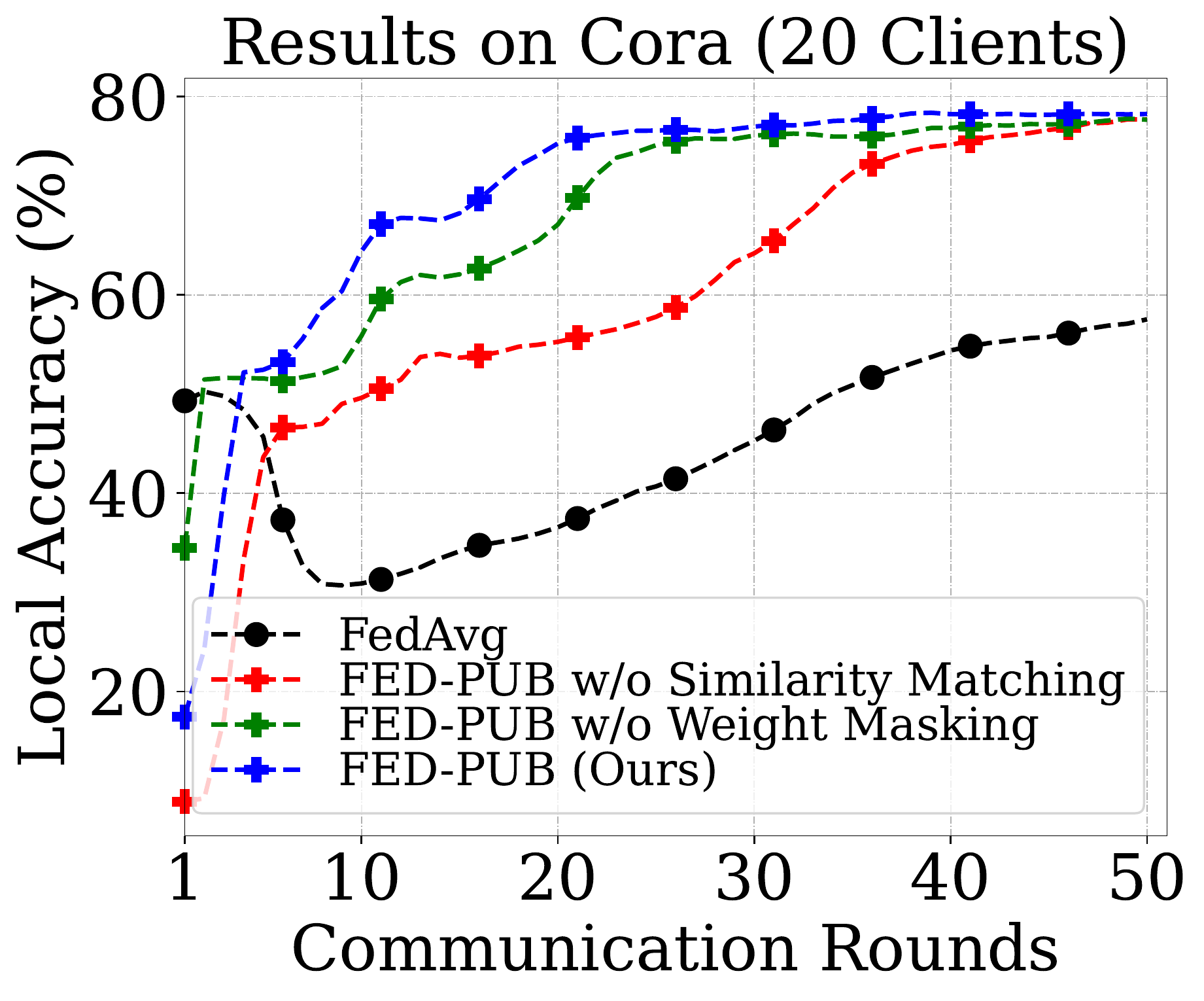}
            \vspace{-0.23in}
            \subcaption{\scriptsize Overlapping}
        \end{minipage}
        \begin{minipage}{0.49\linewidth}
            \centering
            \includegraphics[width=1\linewidth]{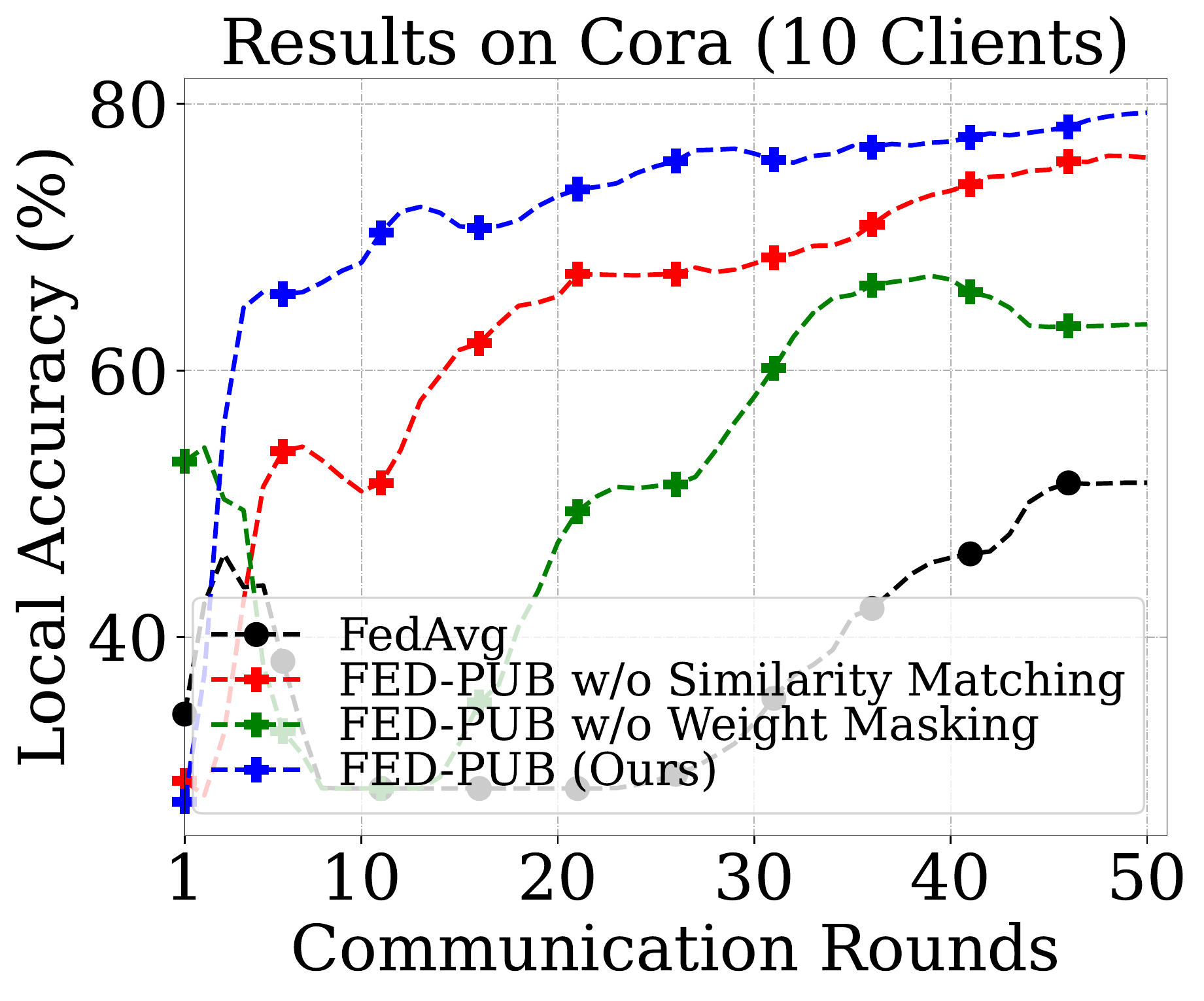}
            \vspace{-0.23in}
            \subcaption{\scriptsize Non-overlapping}
        \end{minipage}
        \vspace{-0.1in}
        \caption{\small \textbf{Ablation studies} of the proposed FED-PUB on both overlapping (a) and non-overlapping (b) subgraph scenarios, on the Cora dataset.}
    \label{fig:ablation}
    \end{minipage}
\end{figure*}

%% file: figures/neighbor.tex
\begin{figure*}[!t]
    \begin{minipage}{0.435\linewidth}
        \resizebox{\linewidth}{!}{
            \renewcommand{\arraystretch}{0.95}
            \begin{tabular}{l c c c}
                \toprule
                \midrule
                 \textbf{Model} & \textbf{Acc. [\%]} & \textbf{Model Size [\%]} & \textbf{Cost [\%]} \\
                \midrule
                     FedAvg & 76.48 $\pm$ 0.36 & 100.00 $\pm$ 0.00 & 100.00 $\pm$ 0.00 \\
                \midrule
                     FedGNN & 70.63 $\pm$ 0.83 & 100.00 $\pm$ 0.00 & 214.94 $\pm$ 0.00 \\
                     FedSage+ & 77.52 $\pm$ 0.46 & 100.00 $\pm$ 0.00 & 276.84 $\pm$ 0.00 \\
                     GCFL & 78.84 $\pm$ 0.26 & 100.00 $\pm$ 0.00 & 100.00 $\pm$ 0.00 \\
                 \midrule
                     \textbf{Ours} ($\lambda_1$=$9e$-$1$) & 77.36 $\pm$ 0.99 & \textbf{25.13} $\pm$ 0.34 & \textbf{37.70} $\pm$ 0.56 \\
                     \textbf{Ours} ($\lambda_1$=$7e$-$1$) & 79.46 $\pm$ 0.41 & 42.59 $\pm$ 1.33 & 63.89 $\pm$ 1.99 \\
                     \textbf{Ours} ($\lambda_1$=$5e$-$1$) & \textbf{79.89} $\pm$ 0.12 & 57.07 $\pm$ 0.52 & 85.61 $\pm$ 0.78\\
                \midrule
                \bottomrule
            \end{tabular}
        }
        \vspace{-0.095in}
        \captionof{table}{\textbf{Analyses on efficiencies} of communication costs and model sizes with sparse masks on Cora with 10 clients.}
        \label{fig:efficiency}
    \end{minipage}
    \hfill
    \begin{minipage}{0.265\linewidth}
        \centering
        \vspace{-0.03in}
        \includegraphics[width=1\linewidth]{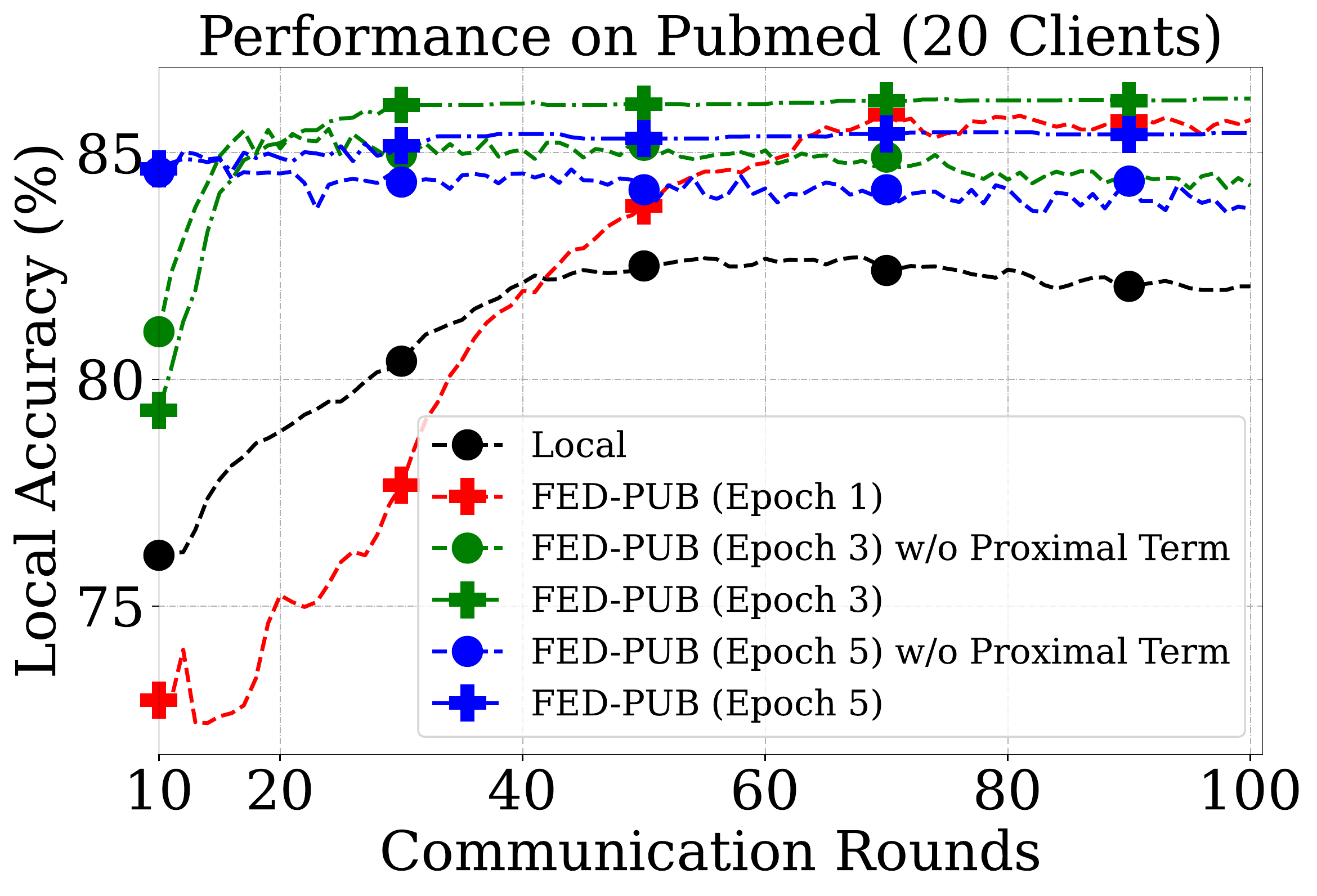} 
        \vspace{-0.27in}
        \caption{\textbf{Performances by varying the local epochs}.}
        \label{fig:epoch}
    \end{minipage}
    \hfill
    \begin{minipage}{0.265\linewidth}
        \centering
        \vspace{-0.03in}
        \includegraphics[width=1\linewidth]{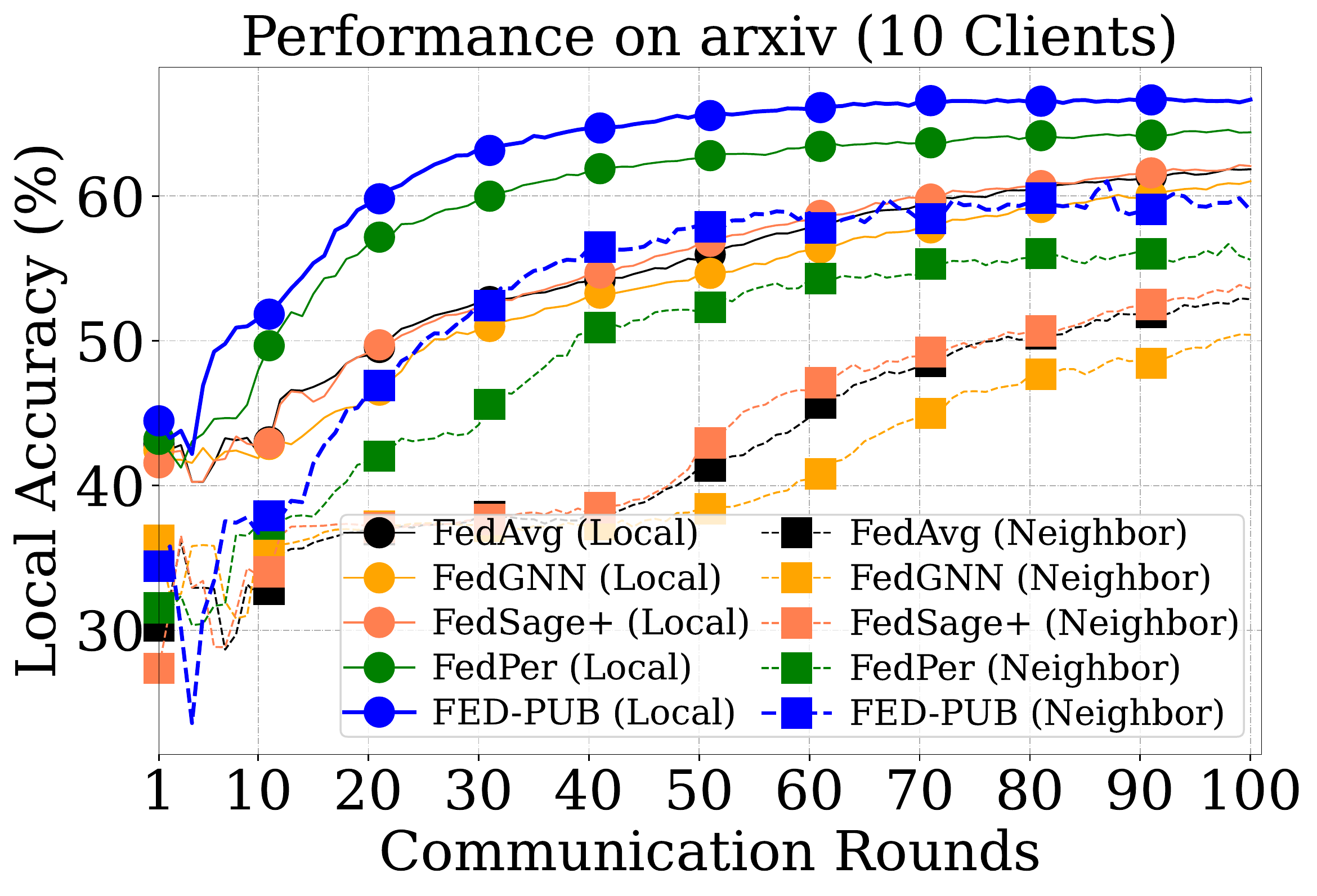}
        \vspace{-0.27in}
        \caption{\textbf{Performances on interrelated neighboring subgraphs}.}
        \label{fig:neighbor}
    \end{minipage}
    \vspace{-0.1in}
    \label{fig:analysis}
\end{figure*}

%% file: tables/link_prediction.tex
\begin{figure*}[t]
    \small
    \centering
    \begin{minipage}{0.30\textwidth}
    \resizebox{\textwidth}{!}{
        \renewcommand{\arraystretch}{1.01}
        \begin{tabular}{lccc}
            \toprule
            Methods & 5 Clients & 10 Clients \\
            \midrule
            Local & 90.49 & 89.58 \\
            \midrule
            FedAvg & 86.04 & 82.76 \\
            FedProx & 84.75 & 82.20 \\ 
            FedPer & 91.33 & 89.06 \\
            FedSage+ & 84.25 & 84.38 \\
            GCFL & 90.36 & 83.10 \\ 
            \midrule
            FED-PUB (Ours) & 91.76 & 91.04 \\
            \bottomrule
        \end{tabular}
    }
    \vspace{-0.06in}
    \caption{\small \textbf{Link prediction results} on ogbn-arxiv with clients of 5 and 10.}
    \label{tab:link_prediction}
    \end{minipage}
    \hfill
    \begin{minipage}{0.28\textwidth}
    \vspace{-0.03in}
        \includegraphics[width=1.0\textwidth]{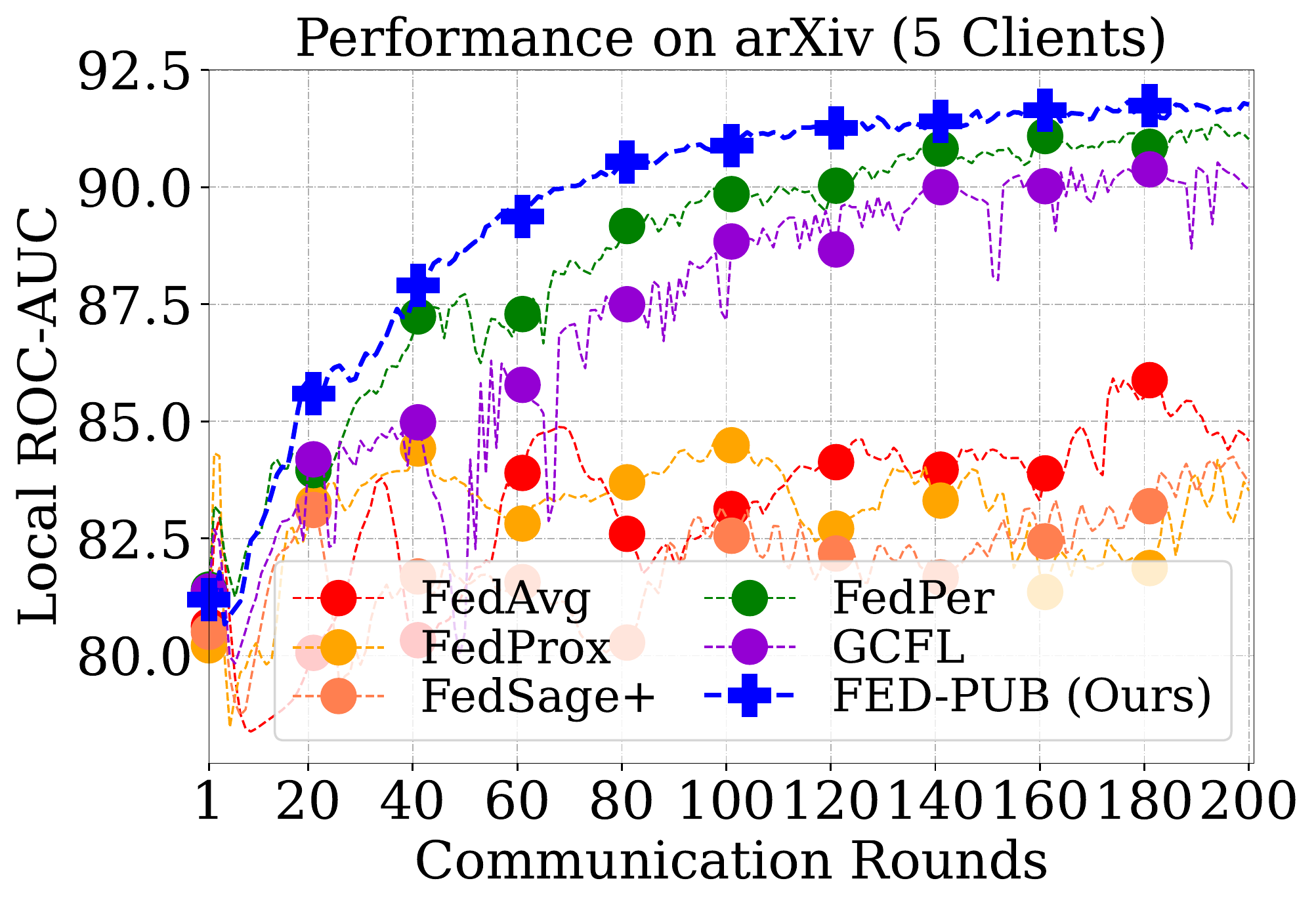}
    \vspace{-0.275in}
    \caption{\small \textbf{Convergence plots for link prediction} with 5 clients.}
    \label{fig:link_prediction}
    \end{minipage}
    \hfill
    \begin{minipage}{0.40\textwidth}
    \resizebox{\textwidth}{!}{
        \begin{tabular}{lccc}
            \toprule
            Methods & Cora & CiteSeer & PubMed \\
            \midrule
            Local & 83.19 ± 0.53 & 69.68 ± 0.38 & 83.88 ± 0.17 \\
            \midrule
            FedAvg & 68.18 ± 0.66 & 66.71 ± 1.54 & 83.08 ± 0.21 \\
            FedProx & 65.70 ± 1.89 & 68.17 ± 1.74 & 83.07 ± 0.28 \\ 
            FedPer & 82.06 ± 1.34 & 70.20 ± 0.60 & 85.85 ± 0.18 \\
            FedGNN & 72.72 ± 0.56 & 65.03 ± 1.18 & 81.60 ± 0.38 \\
            FedSage+ & 68.42 ± 0.80 & 66.22 ± 0.47 & 83.17 ± 0.13 \\
            GCFL & 82.72 ± 0.40 & 69.82 ± 0.87 & 84.82 ± 0.28 \\ 
            \midrule
            FED-PUB (Ours) & \textbf{85.41} ± 0.19 & \textbf{73.30} ± 0.13 & \textbf{86.44} ± 0.45 \\
            \bottomrule
        \end{tabular}
    }
    \vspace{-0.06in}
    \caption{\textbf{Results of the imbalance node scenario} on the non-overlapping node setting with 10 clients.}
    \label{tab:imbalanced}
    \end{minipage}
    \vspace{-0.12in}
\end{figure*}

%% file: sections/7_conclusion.tex
\section{Conclusion}
\label{conclusion}
In this work, we introduced a novel problem of personalized subgraph FL, which focuses on the joint improvement of local GNNs working on interrelated subgraphs (e.g. subgraphs belonging to the same community) by selectively utilizing knowledge from other models. The proposed personalized subgraph FL is highly challenging due to 1) the difficulty in computing similarities between local subgraphs that are only locally accessible, and 2) the problem of knowledge collapse among local GNNs that work on heterogeneous subgraphs during weight aggregation. To this end, we proposed a novel personalized subgraph FL framework, called FEDerated Personalized sUBgraph learning (FED-PUB), which estimates similarities between subgraphs using functional embeddings of their GNN models on random graphs, and uses them to perform a weighted average of the local models for each client. Further, we mask out globally given weights to focus on only the relevant subnetwork for each community and client. We extensively validated our FED-PUB framework on multiple benchmark datasets with overlapping and non-overlapping subgraphs, on which our FED-PUB significantly outperforms relevant baselines. Further analyses show the effectiveness of our similarity matching method for capturing the community structures, and also our weight masking strategy for tackling the subgraph heterogeneity.

\section*{Acknowledgements}
We thank the anonymous reviewers for their constructive comments. This work was supported by the Institute of Information \& communications Technology Planning \& Evaluation (IITP) grant funded by the Korea government (MSIT) (No.2019-0-00075, Artificial Intelligence Graduate School Program (KAIST)), the Engineering Research Center Program through the National Research Foundation of Korea (NRF) funded by the Korea Government (MSIT) (NRF-2018R1A5A1059921), and Samsung Research.

%% file: sections/8_appendix.tex
\section{Algorithms}
\label{appendix:algo}

\begin{figure*}[h!]
    \vspace{-0.2in}
    \begin{minipage}[t]{.495\linewidth}
        \begin{algorithm}[H]
        \small
    	\caption{\textbf{FED-PUB} Client Algorithm}
    	    \begin{algorithmic}[1]
    	        \label{algo:fedpub_client}
                \STATE  $R$: the number of rounds, $E$: the number of epochs, $K$: the number of clients, $G_i$: local subgraph for client $i$, $f_i$: model function for client $i$, $\boldsymbol{\theta}_i$: model parameters for client $i$, $\boldsymbol{\mu}_i$: weight masking parameters for client $i$, $S(\cdot)$: similarity matching function, and $\tau$: scaling factor for similarity matching. \\
                \vspace{0.215in}
                \STATE   \textbf{Function} RunClient($\boldsymbol{\bar{\theta}}_{i}$)
                    \STATE $\boldsymbol{\theta}_{i} \leftarrow \boldsymbol{\bar{\theta}}_i \odot \boldsymbol{\mu}_i $ 
                    \FOR{each local epoch $e$ from $1$ to $E$}
                        \STATE $\boldsymbol{\theta}_i \leftarrow \boldsymbol{\theta}_i-\eta\nabla\mathcal{L}(G_i; \boldsymbol{\theta}_i, \boldsymbol{\mu}_i)$
                    \ENDFOR
                    \STATE \textbf{return} $\boldsymbol{\theta}_i$
        	\end{algorithmic}
    	\end{algorithm}
    \end{minipage}
    \begin{minipage}[t]{.495\linewidth}
        \begin{algorithm}[H]
            \small
            \caption{\textbf{FED-PUB} Server Algorithm}
        	\begin{algorithmic}[1]
        	    \label{algo:fedpub_server}
        	   
                \STATE   \textbf{Function} RunServer()
                    \STATE initialize $\boldsymbol{\bar{\theta}}^{(1)} $
                    \FOR{each round $r=1,2,\dots, R$}
                        \FOR{$\forall i~\textbf{\mbox{in parallel}}$}
                            \IF{$r = 1$} 
                                \STATE $\boldsymbol{\theta}_i^{(r+1)} \leftarrow \text{RunClient} (\boldsymbol{\bar{\theta}}^{(r)})$
                            \ELSE
                                
                                \STATE $\boldsymbol{\bar{\theta}}_{i}^{(r)} \leftarrow \sum_{j=0}^K \frac{\text{exp}({\tau \cdot S(i, j)})}{\sum_{k=0}^K \text{exp}(\tau \cdot S(i, k))}  \boldsymbol{\theta}_j$
                                \STATE $\boldsymbol{\theta}_i^{(r+1)} \leftarrow \text{RunClient} (\boldsymbol{\bar{\theta}}_{i}^{(r)})$
                            \ENDIF 
                    	\ENDFOR
                    \ENDFOR 
                
        	\end{algorithmic}
    	\end{algorithm}
    \end{minipage}
    
\end{figure*}

In this section, we provide algorithms for the proposed subgraph similarity estimation and adaptive weight masking methods in our FED-PUB framework. In particular, weight masking, performed in the client, is shown in Algorithm~\ref{algo:fedpub_client}. Also, similarity matching, performed in the server, is shown in Algorithm~\ref{algo:fedpub_server}.

\section{Experimental Setups}
\label{appendix:setup}

In this section, we first provide the descriptions of six different benchmark datasets that we use, along with their preprocessing setups for federated learning and their statistics in Subsection~\ref{appendix:sub:data}. Then, we explain the baselines and our proposed FED-PUB in detail in Subsection~\ref{appendix:sub:model}. Lastly, we further describe the implementation details for experiments on synthetic and real-world graphs, as well as additional experimental details on functional similarities and sparse masks in Subsection~\ref{appendix:sub:imple}.

\begin{table*}[t]
\caption{\small \textbf{Dataset statistics.} We report the number of nodes, edges, classes, clustering coefficient, and heterogeneity for the original graph and its split subgraphs on overlapping and disjoint node scenarios. Ori denotes the original graph, and Cli denotes the number of clients.}
\vspace{-0.075in}
\label{tab:appendix:data}
\small
\centering
\resizebox{\textwidth}{!}{
\renewcommand{\arraystretch}{0.9}
\begin{tabular}{lcccccccccccc}
\toprule
\multicolumn{13}{l}{\bf \emph{Overlapping node scenario}} \\
\midrule
& \multicolumn{4}{c}{\bf Cora} & \multicolumn{4}{c}{\bf CiteSeer} & \multicolumn{4}{c}{\bf Pubmed}  \\
\cmidrule(l{2pt}r{2pt}){2-5} \cmidrule(l{2pt}r{2pt}){6-9} \cmidrule(l{2pt}r{2pt}){10-13}
& \textbf{Ori} & \textbf{10 Cli} & \textbf{30 Cli} & \textbf{50 Cli} & \textbf{Ori} & \textbf{10 Cli} & \textbf{30 Cli} & \textbf{50 Cli} & \textbf{Ori} & \textbf{10 Cli} & \textbf{30 Cli} & \textbf{50 Cli} \\
\midrule

\# Classes & \multicolumn{4}{c}{7} & \multicolumn{4}{c}{6} & \multicolumn{4}{c}{3} \\
\# Nodes   & 2,485 & 621 & 207 & 124 & 2,120 & 530 & 177 & 106 & 19,717 & 4,929 & 1,643 & 986 \\
\# Edges   & 10,138 & 1,249 & 379 & 215 & 7,358 & 889 & 293 & 170  & 88,648 & 10,675 & 3,374 & 1,903 \\
Clustering Coefficient & 0.238 & 0.133 & 0.129 & 0.125 & 0.170 & 0.088 & 0.087 & 0.096 & 0.060 & 0.035 & 0.034 & 0.035 \\
Heterogeneity & N/A & 0.297 & 0.567 & 0.613 & N/A & 0.278 & 0.494 & 0.547 & N/A & 0.210 & 0.383 & 0.394 \\

\midrule

& \multicolumn{4}{c}{\bf ogbn-arxiv} & \multicolumn{4}{c}{\bf Amazon-Computer} & \multicolumn{4}{c}{\bf Amazon-Photo} \\
\cmidrule(l{2pt}r{2pt}){2-5} \cmidrule(l{2pt}r{2pt}){6-9} \cmidrule(l{2pt}r{2pt}){10-13}
& \textbf{Ori} & \textbf{10 Cli} & \textbf{30 Cli} & \textbf{50 Cli} & \textbf{Ori} & \textbf{10 Cli} & \textbf{30 Cli} & \textbf{50 Cli} & \textbf{Ori} & \textbf{10 Cli} & \textbf{30 Cli} & \textbf{50 Cli} \\
\midrule

\# Classes & \multicolumn{4}{c}{40} & \multicolumn{4}{c}{10} & \multicolumn{4}{c}{8} \\
\# Nodes   & 169,343 & 42,336 & 14,112 & 8,467 & 13,381 & 3,345 & 1,115 & 669 & 7,487 & 1,872 & 624 & 374 \\
\# Edges   & 2,315,598 & 282,083 & 83,770 & 44,712 & 491,556 & 59,236 & 16,684 & 8,969 & 238,086 & 29,223 & 8,735 & 4,840 \\
Clustering Coefficient & 0.226 & 0.177 & 0.185 & 0.191 & 0.351 & 0.337 & 0.348 & 0.359 & 0.410 & 0.380 & 0.391 & 0.410  \\
Heterogeneity & N/A & 0.315 & 0.606 & 0.615 & N/A & 0.327 & 0.577 & 0.614 & N/A & 0.306 & 0.696 & 0.684 \\

\midrule
\midrule

\multicolumn{13}{l}{\bf \emph{Non-overlapping node scenario}} \\
\midrule
& \multicolumn{4}{c}{\bf Cora} & \multicolumn{4}{c}{\bf CiteSeer} & \multicolumn{4}{c}{\bf Pubmed}  \\
\cmidrule(l{2pt}r{2pt}){2-5} \cmidrule(l{2pt}r{2pt}){6-9} \cmidrule(l{2pt}r{2pt}){10-13}
& \textbf{Ori} & \textbf{5 Cli} & \textbf{10 Cli} & \textbf{20 Cli} & \textbf{Ori} & \textbf{5 Cli} & \textbf{10 Cli} & \textbf{20 Cli} & \textbf{Ori} & \textbf{5 Cli} & \textbf{10 Cli} & \textbf{20 Cli} \\
\midrule

\# Classes &  \multicolumn{4}{c}{7} & \multicolumn{4}{c}{6} & \multicolumn{4}{c}{3} \\
\# Nodes   & 2,485 & 497 & 249 & 124 & 2,120 & 424 & 212 & 106 & 19,717 & 3,943 & 1,972 & 986 \\
\# Edges   & 10,138 & 1,866 & 891 & 422 & 7,358 & 1,410 & 675 & 326 & 88,648 & 16,374 & 7,671 & 3,607 \\
Clustering Coefficient & 0.238 & 0.250 & 0.259 & 0.263 & 0.170 & 0.175 & 0.178 & 0.180 & 0.060 & 0.063 & 0.066 & 0.067 \\
Heterogeneity & N/A & 0.590 & 0.606 & 0.665 & N/A & 0.517 & 0.541 & 0.568 & N/A & 0.362 & 0.392 & 0.424 \\

\midrule

& \multicolumn{4}{c}{\bf ogbn-arxiv} & \multicolumn{4}{c}{\bf Amazon-Computer} & \multicolumn{4}{c}{\bf Amazon-Photo} \\
\cmidrule(l{2pt}r{2pt}){2-5} \cmidrule(l{2pt}r{2pt}){6-9} \cmidrule(l{2pt}r{2pt}){10-13}
& \textbf{Ori} & \textbf{5 Cli} & \textbf{10 Cli} & \textbf{20 Cli} & \textbf{Ori} & \textbf{5 Cli} & \textbf{10 Cli} & \textbf{20 Cli} & \textbf{Ori} & \textbf{5 Cli} & \textbf{10 Cli} & \textbf{20 Cli} \\
\midrule

\# Classes &  \multicolumn{4}{c}{40} & \multicolumn{4}{c}{10} & \multicolumn{4}{c}{8} \\
\# Nodes   & 169,343 & 33,869 & 16,934 & 8,467 & 13,381 & 2,676 & 1,338 & 669 & 7,487 & 1,497 & 749 & 374 \\
\# Edges   & 2,315,598 & 410,948 & 182,226 & 86,755 & 491,556 & 84,480 & 36,136 & 15,632 & 238,086 & 43,138 & 19,322 & 8,547  \\
Clustering Coefficient & 0.226 & 0.247 & 0.259 & 0.269 & 0.351 & 0.385 & 0.398 & 0.418 & 0.410 & 0.437 & 0.457 & 0.477 \\
Heterogeneity & N/A & 0.593 & 0.615 & 0.637 & N/A & 0.604 & 0.612 & 0.647 & N/A & 0.684 & 0.681 & 0.751 \\

\bottomrule

\end{tabular}
}
\vspace{-0.1in}
\end{table*}

\subsection{Datasets}
\label{appendix:sub:data}

We report statistics of six different benchmark datasets~\citep{planetoid, ogb, amazon, amnazonsubset}, such as Cora, CiteSeer, Pubmed, and ogbn-arxiv for citation graphs; Computer and Photo for amazon product graphs, which we use in our experiments, for both the overlapping and non-overlapping node scenarios in Table~\ref{tab:appendix:data}. Specifically, in Table~\ref{tab:appendix:data}, we report the number of nodes, edges, classes, and clustering coefficient for each subgraph, but also the heterogeneity between the subgraphs. Note that, to measure the clustering coefficient, which indicates how much nodes are clustered together, for the individual subgraph, we first calculate the clustering coefficient~\citep{watts1998collective} for all nodes, and then average them. On the other hand, to measure the heterogeneity, which indicates how disjointed subgraphs are dissimilar, we calculate the median Jenson-Shannon divergence of label distributions between all pairs of subgraphs.

For dataset splits, we randomly sample 20\% nodes for training, 35\% for validation, and 35\% for testing, for all datasets except for the arxiv dataset. This is because the arxiv dataset has a relatively larger number of nodes compared to the other datasets, as reported in Table~\ref{tab:appendix:data}. Therefore, for this dataset, we randomly sample 5\% nodes for training, the remaining half of the nodes for validation, and the other nodes for testing.

We then describe how to partition the original graph into multiple subgraphs, whose number is the same as the number of clients (i.e., FL participants). In general, we use the METIS graph partitioning algorithm~\citep{Karypis95metis} to divide the original graph into multiple subgraphs, which can control the number of disjoint subgraphs as parameters. Consequently, in the non-overlapping node scenario, the disjoint subgraph for each client is directly obtained by the output of the METIS algorithm (i.e., if we set the parameter value for METIS as 10, then we can obtain 10 different disjoint subgraphs, each of which is given to each client). On the other hand, in the overlapping node scenario where nodes are duplicated across different subgraphs, we first divide the original graph into 2, 6, and 10 disjoint subgraphs for 10 clients, 30 clients, and 50 clients, respectively, with the METIS algorithm. After that, in each split subgraph, we randomly sample half of the nodes and their associated edges, and then use them as the subgraph for one particular client. This procedure is performed five times to generate five different yet overlapped subgraphs, per one split subgraph obtained from METIS.

\subsection{Baselines and Our Model}
\label{appendix:sub:model}

\begin{enumerate}[itemsep=0.9mm, parsep=1pt, leftmargin=*]

    \item \textbf{FedAvg}: This method~\citep{McMahan2017CommunicationEfficientLO} is the FL baseline, where each client locally updates a model and sends it to a server, while the server aggregates the locally updated models with respect to their numbers of training samples and transmits the aggregated model back to the clients.
    
    \item \textbf{FedProx}: This method~\citep{li2020federated} is the FL baseline, which prevents the local model from drifting to the local data by minimizing weight differences between local and global models.
    
    \item \textbf{FedPer}: This method~\citep{arivazhagan2019federated} is the personalized FL baseline, which shares only the base layers, while keeping the personalized classification layers in the local side.
    
    \item \textbf{FedGNN}: This method~\citep{FedGNN} is the subgraph FL baseline, which expands local subgraphs by augmenting the relevant nodes from other clients. In the original paper, if two nodes in two different clients have exactly the same neighboring nodes, this method transmits and augments the nodes having the same neighborhoods in other clients with nodes in the original client. In our non-overlapping node scenario, since nodes are disjoint across clients, we measure the similarities between nodes of different clients and augment them having the similarity above the threshold (e.g., 0.5).
    
    \item \textbf{FedSage+}: This method~\citep{FedSage} is the subgraph FL baseline, which expands local subgraphs by generating additional nodes with the local graph generator. To train the graph generator, each client first receives node representations from other clients, and then calculates the gradient of distances between the local node features and the other client's node representations. Then, the gradient is sent back to other clients, which is then used to train the graph generator.
    
    \item \textbf{GCFL}: This method~\citep{GCFL} is the graph FL baseline, which targets completely disjoint graphs (e.g., molecular graphs) as in image tasks. In particular, it uses the bi-partitioning scheme, which divides a set of clients into two disjoint client groups based on their gradient similarities. Then, the model weights are only shared between grouped clients having similar gradients, after partitioning. Note that this bi-partitioning mechanism is similar to the mechanism proposed in clustered-FL~\citep{sattler2020clustered} for image classification, and we adopt this for our subgraph FL.
    
    \item \textbf{Local}: This method is the non-FL baseline, which only locally trains the model for each client without weight sharing.
    
    \item \textbf{FED-PUB}: This is our FEDerated Personalized sUBgraph learning (FED-PUB) framework, which not only estimates the similarities between subgraphs based on their models' functional embeddings for discovering community structures, but also adaptively masks the received weights from the server to filter irrelevant weights from heterogeneous communities.

\end{enumerate}

\subsection{Implementation Details}
\label{appendix:sub:imple}

\paragraph{Implementation Details on Functional Embeddings} The functional embeddings are key ingredients in the proposed FED-PUB framework, to capture community structures of interconnected subgraphs leveraged in personalized weight aggregation (See Section~\ref{method:community}). To obtain such functional embeddings, the input of GNNs is important, which we randomly generates via a stochastic block model~\citep{SBM}. Specifically, we first sample five individual subgraphs, each of which has 100 nodes, in which the probability of edges within the single graph is 0.1, while the probability of edges between different graphs is 0.01. Also, we initialize the node features with the normal distribution of 0.0 mean and 1.0 variance. Note that, in practice, this randomly sampled graph is initialized on the server-side at once, and the server distributes it to all clients. Then, the client calculates its model's functional embedding, and then transmits it to the server. However, the effect is the same even if we calculate the functional embeddings on the server-side, which is up to the FL system design.

\vspace{-0.05in}
\paragraph{Implementation Details on Sparse Masks}
As described in Section~\ref{method:mask}, we propose to sparsify the local personalized mask $\boldsymbol{\mu}_i$ for each client $i$, for taking the benefits in communication and prediction costs. In this paragraph, we additionally provide the detailed implementation specifications on sparse masks during training and test phases of our FED-PUB. First, in training, we regularize the local mask $\boldsymbol{\mu}_k$ to be sparse by minimizing the $L_1$ Norm of it along with its scaling parameter $\lambda_2$ to the local loss $\mathcal{L}$, represented in equation~\ref{eq:loss}. However, this regularization scheme might not be enough to exactly make a subset of local masks zero. Therefore, in the test phase, we use the threshold scheme, where elements (neurons) of $\boldsymbol{\mu}_k$ below a certain threshold (i.e., $\lambda_2$) are set to zero. By doing so, we can transmit only the partial parameters to the server, but also can predict with only the partial parameters; therefore, effectively reducing both communication and prediction costs.

\vspace{-0.05in}
\paragraph{Common Implementation Details for Experiments}
For all experiments, we stack two layers of Graph Convolutional Network (GCN)~\citep{GCN} and one linear classifier layer on top of them. Regarding hyperparameters, the number of hidden dimensions is set to 128, and the learning rate is set to 0.001. All models are optimized with Adam optimizer~\citep{Adam}. Also, all clients participate in the federated learning at every round. For all experiments about our FED-PUB framework, we set $\lambda_1$ and $\lambda_2$ values for $L_1$ and $L_2$ losses in equation~\ref{eq:loss} for sparsity and proximal terms as 0.001. While we can tune such two scaling hyperparameters, we observe that those default values show satisfactory performances across all datasets without specific tuning to each dataset (See Appendix~\ref{appendix:lambda_analysis} for more analyses).

\vspace{-0.05in}
\paragraph{Implementation Details on Synthetic Graph Experiments}
We perform two experiments on synthetic graphs, which are shown in Figure~\ref{fig:concept} and Figure~\ref{fig:similarity}. In particular, in the experiment of Figure~\ref{fig:concept}, there are three communities that have different label distributions (e.g., nodes in the first community have label 0, whereas nodes in the last community have label 2), and three communities consist of 5/5/40 non-overlapped subgraphs with 50 clients. In communities, each subgraph consists of 30 nodes, and the edges between two nodes in the same community are sampled from the probability of 0.5. Meanwhile, the edges between two nodes in different communities are sampled from the probability of 0.1. Similarly, in the experiment of Figure~\ref{fig:similarity}, there are two communities that have different label distributions, and two communities have 5/15 non-overlapped subgraphs with 20 clients. In communities, each subgraph consists of 30 nodes, and the edges between two subgraphs within the same community are sampled from the probability of 0.7. Meanwhile, the edges between two subgraphs from different communities are sampled from the probability of 0.01. For all experiments, the number of local epochs is set to 3, and the number of total FL rounds is set to 100. In our FED-PER including its variants of using parameter and gradient for subgraph similarity estimation, the scaling hyperparameter (i.e., $\tau$) for the similarity in equation~\ref{eq:agg} is set to 10.

\vspace{-0.05in}
\paragraph{Implementation Details on Real-World Graph Experiments} 
Regarding relatively small datasets, namely Cora, CiteSeer and PubMed, we set the number of local training epoch as 1, and the number of total rounds as 100. For larger datasets, such as Computer, Photo and arxiv, we set the number of total rounds as 200, while the number of local epochs is set to 2 for Photo and arxiv, and set to 3 for Computer. In the overlapping node scenario, we set the similarity scaling hyperparameter (i.e., $\tau$) as 5 for all our models. Meanwhile, we set the similarity scaling hyperparameter (i.e., $\tau$) as 3 in the non-overlapping node scenario for all our models. We observe that, the larger $\tau$ value performs better for the overlapping node scenario, in which different subgraphs are easily grouped together, compared to the disjoint node scenario. Finally, we report the test performance of all models at the best validation epoch, and the performance is measured by the node classification accuracy.

\vspace{-0.05in}
\paragraph{Computing Resources} For all experiments, we use PyTorch~\citep{pytorch} and PyTorch Geometric~\citep{pyg} as deep learning libraries. We use two types of GPUs: GeForce RTX 2080 Ti and TITAN XP, for training models. Note that the runtime of our framework depends on the number of workers for processing clients' jobs in parallel. In general, we use 10 or 20 workers (i.e., simultaneously training 10 or 20 local models for 10 or 20 clients), and, based on 10 workers, the single run of our FED-PUB for training 50 clients with 1 local epoch and 100 total rounds takes less than 2 hours.

\section{Additional Experimental Results}
\label{appendix:results}

In this section, we provide additional experimental results on the sensitivity analyses of hyperparameters in Section~\ref{appendix:lambda_analysis}; varying the graph partitioning schemes in Section~\ref{appendix:Louvain} and~\ref{appendix:random}; varying the random graph inputs in Section~\ref{appendix:results:functional}; and varying the similarity estimation schemes in Section~\ref{appendix:results:similarity}. In addition to them, we also analyze the heterogeneity in subgraph FL in Section~\ref{appendix:heterogeneity} and its relationship to the graph size in Section~\ref{appendix:graph_size}, as well as the impact of missing edges to the task performance in Section~\ref{appendix:results:missingedge}.

\subsection{Results on Varying Scaling Hyperparameters in Loss Function}
\label{appendix:lambda_analysis}

\input{figures/fig_hyper_params}

In Table~\ref{fig:appendix_lambda}, we explore the effects of hyperparameters $\lambda_1$ and $\lambda_2$ on the Cora dataset with the overlapping node scenario, where the number of local epochs is set as 2 and the number of clients is set as 10. In particular, $\lambda_1$ value can control the degree of the model sparsity; therefore, to see its efficacy, we fix $\lambda_2$ value while varying $\lambda_1$, and then measure both the model sparsity and performance. As shown in Table~\ref{fig:appendix_lambda} left, higher $\lambda_1$ values significantly increase the model sparsity, meanwhile, the model performance is slightly decreased. This result indicates that we should consider the trade-off between the sparsity and the model performance when selecting $\lambda_1$ value. On the other hand, $\lambda_2$ value is designed to prevent the excessive knowledge drift to the local subgraph distribution, and, to verify its effectiveness, we fix $\lambda_1$ value while varying $\lambda_2$. As shown in Table~\ref{fig:appendix_lambda} right, small lambda values lead to performance degeneration, meanwhile, choosing the sufficiently large $\lambda_2$ values (e.g., 1e-1) would yield high performance. Further, we observe that the sparsity does not depend on $\lambda_2$ value in Table~\ref{fig:appendix_lambda} right, which suggests that the effects of $\lambda_1$ and $\lambda_2$ are orthogonal and complementary.

\begin{wraptable}{t}{0.45\textwidth}
    \centering
    \vspace{-0.16in}
    \caption{\small Results on experimental settings of Louvain graph partitioning algorithms, following~\citet{FedSage}.}
    \vspace{-0.1in}
    \resizebox{0.44\textwidth}{!}{
    \begin{tabular}{lccc}
        \toprule
        Methods & Cora & CiteSeer & PubMed \\
        \midrule
        Local & 78.56 ± 0.27 & 64.06 ± 0.09 & 84.07 ± 0.17 \\
        \midrule
        FedAvg & 71.83 ± 0.40 & 69.23 ± 0.71 & 82.47 ± 0.32 \\
        FedProx & 72.09 ± 0.29 & 67.66 ± 0.97 & 82.68 ± 0.34 \\ 
        FedPer & 80.13 ± 0.50 & 66.28 ± 1.22 & 85.02 ± 0.23 \\
        FedGNN & 76.59 ± 0.66 & 61.21 ± 1.46 & 82.67 ± 0.26 \\
        FedSage+ & 72.20 ± 0.60 & 68.40 ± 0.61 & 82.76 ± 0.09 \\
        GCFL & 78.55 ± 0.38 & 64.20 ± 0.31 & 84.62 ± 0.31 \\ 
        \midrule
        FED-PUB (Ours) & \textbf{82.68} ± 0.13 & \textbf{69.45} ± 0.75 & \textbf{86.20} ± 0.11 \\
        \bottomrule
    \end{tabular}
    }
    \label{tab:Louvain}
    \vspace{-0.1in}
\end{wraptable}

\subsection{Results on Louvain Graph Partitioning Algorithm}
\label{appendix:Louvain}

To validate our FED-PUB framework on different graph partitioning settings for subgraph FL, we use another experimental setup from~\citet{FedSage}, which uses Louvain algorithm~\citep{blondel2008fast} to partition the entire graph into several subgraphs for FL clients. Before explaining experimental results, we would like to point out that there is a drawback in the Louvain algorithm presented in~\citet{FedSage}, unlike the METIS algorithm~\citep{Karypis95metis} that we use, for subgraph FL scenarios. Specifically, since the Louvain algorithm cannot specify the number of graph partitions, the number of subgraphs on the CiteSeer dataset is 38, where three of them have less than ten nodes. Then, based on those 38 disjoint subgraphs, to generate the particular number of clients (e.g., 10), \citet{FedSage} randomly merge the different subgraphs without considering their graph properties. Therefore, even though each partitioned subgraph has its unique structural role/characteristic, the reconstructed 10 subgraphs from the original 38 subgraphs have mixed properties (i.e., two incompatible subgraphs could be merged), which is suboptimal and might be unrealistic. However, as described in the Datasets paragraph of Section~\ref{experiments:datasets}, the METIS that we use can specify the number of subgraph partitions; therefore, METIS is more appropriate when making the experimental settings for subgraph FL.

As shown in Table~\ref{tab:Louvain}, we conduct experiments with the Louvain graph partitioning algorithm~\citep{blondel2008fast, FedSage}, on Cora, CiteSeer, and PubMed datasets with the number of clients as 10. The results show that our FED-PUB consistently outperforms all the other baselines on this different graph partitioning setting, which further concretizes the effectiveness of our FED-PUB framework.

\newpage

\begin{wraptable}{t}{0.32\textwidth}
    \centering
    \vspace{-0.16in}
    \caption{\small Results on experimental settings of the random graph partitioning.}
    \vspace{-0.1in}
    \resizebox{0.31\textwidth}{!}{
    \begin{tabular}{lc}
        \toprule
        Methods & CiteSeer with 10 Clients \\
        \midrule
        Local & 44.27 ± 1.05 \\
        \midrule
        FedAvg & 60.84 ± 0.80 \\
        FedProx & 59.38 ± 1.66 \\ 
        FedPer & 60.04 ± 0.93 \\
        FedGNN & 54.64 ± 1.67 \\
        FedSage+ & 61.03 ± 0.11 \\
        GCFL & 53.15 ± 1.82 \\ 
        \midrule
        FED-PUB (Ours) & \textbf{63.63} ± 0.86 \\
        \bottomrule
    \end{tabular}
    }
    \label{tab:random}
    \vspace{-0.1in}
\end{wraptable}

\subsection{Results on Random Graph Partitioning Algorithm}
\label{appendix:random}

One might be curious about experimental results on the uniform partitions of graphs, instead of splitting the graph with sophisticated partitioning algorithms (e.g., METIS and Louvain algorithms). Therefore, in this subsection, we explain why the random graph partitioning setting is unrealistic, and further show the performances on this random setting. To be specific, if we partition the entire graph of the CiteSeer dataset into different subgraphs uniformly at random, the number of nodes of each subgraph is larger than the number of edges (e.g., 211 nodes yet 72 edges per subgraph, thus some nodes do not have any edges), which is uncommon in practice. Nonetheless, we further perform experiments on the random split setting with 10 different clients on the CiteSeer dataset. As shown in Table~\ref{tab:random}, the gap between baselines and our model is reduced compared to the non-overlapping and overlapping scenarios in Table~\ref{tab:main:overlap} and Table~\ref{tab:main:nonoverlap}. This is because there are no specific community structures in this random graph partitioning setting; however, our FED-PUB still consistently outperforms all baselines.

\subsection{Analyses on Distribution Shifts Between Subgraphs with Sparse Masks}
\label{appendix:heterogeneity}

To see the distributional shifts between subgraphs in the subgraph FL task, we measure distributional differences of labels between subgraphs with the Jenson-Shannon divergence on the Cora dataset with 20 different clients over the overlapping and non-overlapping scenarios. Then, the experimental results show that the distance (i.e., divergence value) among subgraphs within the same community is 0.384, meanwhile, the distance between subgraphs belonging to different communities is 0.639 for the non-overlapping node scenario. On the other hand, for the overlapping node scenario, the distance among subgraphs within the same community is 0.047, meanwhile, the distance between subgraphs belonging to different communities is 0.528. Thus, these results confirm that the heterogeneity of subgraphs even within the same community is extremely larger in the non-overlapping setup (0.384) compared to the overlapping setup (0.047).

Then, from the above results, we can further argue that personalized weight aggregation based on similarity matching for discovering community structures is not sufficient in disjoint subgraph FL problems, since the model weight received from the completely heterogeneous subgraphs might not be meaningful to the local subgraph task, especially in the non-overlapping setting. However, in this extremely heterogeneous scenario, a personalized weight masking scheme is obviously helpful, since it can filter out irrelevant information transmitted from the other heterogeneous subgraphs, while allowing the model to maintain the locally helpful information in its parameters with sparse local masks. This claim and result on the heterogeneity are also aligned with the results in Figure~\ref{fig:ablation}: ablation study, which shows that the personalized weight masking scheme brings huge performance improvements in the non-overlapping setting (i.e., high heterogeneity between subgraphs), whereas the personalized weight aggregation scheme is more beneficial in the overlapping setting with low heterogeneity.

Lastly, to more closely look at the efficacy of sparse masks in subgraph FL, we empirically analyze whether they can indeed filter out irrelevant weights received from the heterogeneous communities and subgraphs. To do so, we measure how many parameters are shared between the two most dissimilar (i.e., heterogeneous) subgraphs, as well as between the two most similar subgraphs, for the Cora dataset with 20 clients on the non-overlapping node scenario. For the two most similar subgraphs within the same community, 75\% parameters are shared. Meanwhile, for the two heterogeneous subgraphs from two opposite communities, 73\% parameters are filtered by sparse masks (i.e., only 27\% parameters are shared). These results demonstrate that sparse masks can prevent the knowledge collapse from subgraphs of heterogeneous communities.

\begin{wraptable}{t}{0.47\textwidth}
    \centering
    \vspace{-0.45in}
    \caption{\small Results on Cora, CiteSeer, and PubMed datasets on the non-overlapping scenario, with the number of clients of 3.}
    \vspace{-0.12in}
    \resizebox{0.465\textwidth}{!}{
    \renewcommand{\arraystretch}{0.82}
    \begin{tabular}{lccc}
        \toprule
        Methods & Cora & CiteSeer & PubMed \\
        \midrule
        Local & 81.73 ± 0.44 & 68.16 ± 0.25 & 84.81 ± 0.40 \\
        \midrule
        FedAvg & 78.77 ± 0.13 & 69.34 ± 0.23 & 85.29 ± 0.20 \\
        FedProx & 78.91 ± 0.21 & 69.54 ± 0.27 & 85.59 ± 0.18 \\ 
        FedPer & 82.29 ± 0.13 & 69.80 ± 0.33 & 85.34 ± 0.16 \\
        FedGNN & 82.36 ± 0.62 & 67.79 ± 0.49 & 85.57 ± 0.13 \\
        FedSage+ & 77.79 ± 1.96 & 69.35 ± 0.12 & 85.63 ± 0.22 \\
        GCFL & 82.67 ± 0.74 & 68.85 ± 0.58 & 86.20 ± 0.15 \\ 
        \midrule
        FED-PUB (Ours) & \textbf{84.45} ± 0.23 & \textbf{70.66} ± 0.34 & \textbf{86.74} ± 0.16 \\
        \bottomrule
    \end{tabular}
    }
    \label{tab:client:3}
    \vspace{-0.2in}
\end{wraptable}

\subsection{Analyses on Local Graph Size vs Heterogeneity}
\label{appendix:graph_size}

To see how much severe the heterogeneity issues are in terms of the number of clients, we first analyze the exact amount of heterogeneities with respect to the client numbers. In particular, following the reported statistics in Table~\ref{tab:appendix:data}, when we increase the number of clients in both the overlapping and non-overlapping node scenarios, the heterogeneity across subgraphs becomes more severe and problematic for subgraph FL, and thus this becomes an important issue to tackle. Note that, in this work, we address this problem with the sparse local masks described in Section~\ref{method:mask} and Appendix~\ref{appendix:heterogeneity}.

Note that one might be further curious about whether our FED-PUB is still effective when the heterogeneity issue is less significant. To analyze this, we further conduct the experiment in the setting where the number of clients is 3 on Cora, CiteSeer, and PubMed datasets with the non-overlapping node scenario. As shown in Table~\ref{tab:client:3}, compared to the results in Table~\ref{tab:main:nonoverlap} with client numbers of 5, 10, and 20, the performance gaps between our FED-PUB and baselines are much reduced. However, we can clearly observe that our FED-PUB consistently outperforms all baselines with large margins even when the number of clients is small, since there still exists incompatible knowledge across clients, which our FED-PUB effectively handles with personalized weight aggregation and local weight masking schemes.

\begin{wraptable}{t}{0.375\textwidth}
    \centering
    \vspace{-0.18in}
    \caption{\small Results on varying the graph inputs for functional embeddings, over overlapping and non-overlapping node scenarios with 20 clients on Cora.}
    \vspace{-0.1in}
    \resizebox{0.37\textwidth}{!}{
    \renewcommand{\tabcolsep}{2.5mm}
    \renewcommand{\arraystretch}{1.0}
    \begin{tabular}{lccc}
        \toprule
        Graphs & Overlapping & Non-Overlapping \\
        \midrule
        SBM & 0.937 & 0.810 \\
        ER & 0.920 & 0.712 \\
        One & 0.822 & 0.656 \\
        Feature & 0.897 & 0.632 \\
        \bottomrule
    \end{tabular}
    }
    \label{tab:random:graphs}
    \vspace{-0.2in}
\end{wraptable}

\subsection{Results on Varying Graph Inputs for Functional Embeddings}
\label{appendix:results:functional}

As described in Section~\ref{appendix:sub:imple}, to obtain the functional embedding of GNNs, we use the same random graphs for all clients, where random graphs are initialized by the stochastic block model~\citep{SBM} with node features initialized by the normal distribution. We note the underlying assumption on using random graphs is that such randomness may not yield any bias on the functional space, unlike existing node features of the certain subgraph. In other words, we expect that our random graphs are helpful for effectively capturing the similarities and community structures among subgraphs without having bias on any particular graph structures.

In this subsection, to experimentally validate the above statement, we further compare various graph inputs used for calculating the functional embeddings, listed as follows: 1) SBM denotes the random graphs generated by the Stochastic Block Model (SBM) like ours; 2) ER denotes the random graphs generated by the Erdos-Renyi (ER) model~\citep{randomgraph}; 3) One denotes the random graph having only one node; 4) Feature denotes the graph where node features are initialized by the existing ones in the client. We then measure the performances of those four schemes by calculating the correlation coefficient between label distributions and estimated similarities of subgraphs (i.e., the high correlation coefficient means that the estimated similarities from functional embeddings are similar to the actual label distributions) on the Cora dataset of non-overlapping and overlapping node scenarios with 20 clients, which are reported in Table~\ref{tab:random:graphs}. Specifically, as shown in Table~\ref{tab:random:graphs}, compared to the One scheme that uses only one node for calculating the functional embeddings, SBM and ER schemes that use more large numbers of randomly initialized nodes can accurately capture the similarities between subgraphs. This result demonstrates that a sufficient amount of randomness is required to capture the model’s functional space. Also, compared to the Feature scheme that uses existing node representations to calculate the functional embeddings, SBM and ER random models show superiority in capturing similarities among subgraphs, which verifies that randomness indeed helps obtain accurate functional embeddings of models without incurring bias.

\begin{wraptable}{t}{0.36\textwidth}
    \centering
    \vspace{-0.17in}
    \caption{\small Results on varying the similarity calculation schemes: parameter, gradient, label, and our functional embedding, on the overlapping node scenario with 30 clients of the Cora dataset.}
    \vspace{-0.1in}
    \resizebox{0.35\textwidth}{!}{
    \begin{tabular}{lcccc}
        \toprule
        & \multicolumn{4}{c}{Rounds} \\
        \cmidrule(l{2pt}r{2pt}){2-5}
        Model & 20 & 40 & 60 & 80 \\
        \midrule
        FedAvg & 29.94 & 32.69 & 47.84 & 52.42 \\
        \midrule
        Parameter & 29.94 & 35.89 & 47.03 & 52.28 \\
        Gradient & 33.93 & 51.09 & 52.77 & 58.14 \\
        Label & 65.97 & 74.31 & 76.50 & 76.82 \\
        \midrule
        Function (FED-PUB) & 67.82 & 73.51 & 74.66 & 75.90 \\
        \bottomrule
    \end{tabular}
    }
    \label{tab:varying:similarity}
    \vspace{-0.15in}
\end{wraptable}

\subsection{Results on Varying Similarity Estimation Mechanisms}
\label{appendix:results:similarity}

As shown in Figure~\ref{fig:similarity}, our functional embeddings are not only effective but also efficient in capturing similarities between subgraphs, compared against using the parameter and gradient similarities. Additionally, one might consider using label distributions as the proxy for similarity estimation between local subgraphs; however, this scheme may violate the privacy constraint of FL since subgraph labels are private local data stored in the client. Nevertheless, to see its actual performances, we additionally conduct experiments with the label similarities that are calculated by distributional differences between subgraph labels, on the Cora dataset of the overlapping node scenario with the number of clients as 30, and then compare the results with our functional similarities at 20, 40, 60, and 80 rounds.

As reported in Table~\ref{tab:varying:similarity}, we can observe that the models, which utilize the parameter and gradient for subgraph similarities, are inferior to our functional and label similarity schemes. This is because they struggle to discover similar subgraphs within the community due to the curse of dimensionality~\cite{bellman1966dynamic} (See Figure~\ref{fig:similarity}). On the other hand, even though the label similarity model uses privacy-sensitive local information (i.e., label distributions of clients), the performance of our FED-PUB that utilizes the functional embeddings from the privacy-free random graphs is similar to the performance of the label model. Therefore, along with the results in Figure~\ref{fig:heatmap}, these comparison results on similarity schemes further verify the effectiveness of our functional embedding in capturing similarities among subgraphs, for identifying their communities.

\begin{table*}[t]
    \centering
    \caption{\small Results on the overlapping node scenario with 10 clients (top) and the non-overlapping node scenario with 30 clients (bottom), where we report results with mean and standard deviation over three different runs.}
    \vspace{-0.08in}
    \resizebox{0.99\textwidth}{!}{
    \renewcommand{\arraystretch}{0.95}
    \begin{tabular}{lcccccc}
        \toprule
        Methods & Cora & CiteSeer & PubMed & Computer & Photo & obgn-arxiv \\
        \midrule
        \midrule
        \textit{\textbf{Overlapping Node Scenario}} \\
        \midrule
        FED-PUB with Explicit Community & 80.45 ± 0.73 & 69.50 ± 0.20 & 84.76 ± 0.14 & 90.31 ± 0.06 & 92.67 ± 0.08 & 64.56 ± 0.12 \\
        FED-PUB with Implicit Community & \textbf{81.54} ± 0.12 & \textbf{72.35} ± 0.53 & \textbf{86.28} ± 0.18 & \textbf{90.55} ± 0.13 & \textbf{92.73} ± 0.18 & \textbf{66.58} ± 0.08 \\
        \midrule
        \midrule
        \textit{\textbf{Non-Overlapping Node Scenario}} \\
        \midrule
        FED-PUB with Explicit Community & \textbf{76.59} ± 0.39 & 67.57 ± 0.51 & 83.20 ± 0.45 & 87.84 ± 0.42 & 91.26 ± 0.25 & 61.52 ± 0.06 \\
        FED-PUB with Implicit Community & 75.40 ± 0.54 & \textbf{68.33} ± 0.45 & \textbf{85.16} ± 0.10 & \textbf{89.15} ± 0.06 & \textbf{92.01} ± 0.07 & \textbf{63.34} ± 0.12 \\
        \bottomrule
    \end{tabular}
    }
    \label{tab:explicit_community}
    \vspace{-0.2in}
\end{table*}

\subsection{Analyses on Implicit and Explicit Communities in Weight Aggregation}
\label{appendix:results:explicit_community}
As formalized in Equation~\ref{eq:agg} and described in Section~\ref{method:community}, we implicitly model the community structures by performing weight aggregation over all available clients. However, one can alternatively perform the explicit weight aggregation, by grouping similar subgraphs within the community first and then performing weight aggregation exclusively among clients discovered within the same community. To see which strategy is superior, we compare the performances of our model variants: implicit and explicit community detection, in personalized weight aggregation. Specifically, for the implicit setup, we use the formulation defined in Equation~\ref{eq:agg} without any modification. Meanwhile, for the explicit setup, we exclusively perform weight aggregation between clients, having a functional similarity score above 0.5, which we regard as forming the community. Note that, for those two variants, we use the same normalization trick in Equation~\ref{eq:agg} after identifying communities, and also the same random graph inputs for obtaining functional embeddings.

As shown in Table~\ref{tab:explicit_community}, we observe that the model, which implicitly captures the community structures during weight aggregation, consistently outperforms the other explicit one, except for only one case: Cora with the overlapping node scenario. We believe such an exceptional case on the Cora dataset with the overlapping node scenario might be because, the information in the other communities is especially not useful for this particular setup (i.e., the model might capture necessary knowledge only with subgraphs within the same community); therefore, completely ignoring the information from other communities contributes to the improved performance. Except for this case, the results in Table~\ref{tab:explicit_community} confirm that implicit modeling of community structures is generally better for personalized weight aggregation in subgraph FL.

\begin{wraptable}{t}{0.52\textwidth}
    \centering
    \vspace{-0.17in}
    \caption{\small Results on Non-Overlapping and Overlapping node scenarios with varying the number of clients on Cora. The Oracle model is not comparable, which trains with the global graph including missing edges.}
    \vspace{-0.11in}
    \resizebox{0.51\textwidth}{!}{
    \renewcommand{\arraystretch}{0.98}
    \begin{tabular}{lcccc}
        \toprule
        & \multicolumn{2}{c}{Non-overlapping}  & \multicolumn{2}{c}{Overlapping} \\
        \cmidrule(l{2pt}r{2pt}){2-3} \cmidrule(l{2pt}r{2pt}){4-5}
        Model & 5 Clients & 20 Clients & 10 Clients & 50 Clients \\
        \midrule
        Oracle & 85.07 & 85.47 & 85.08 & 85.28 \\
        \midrule
        Local & 81.30 & 80.30 & 73.98 & 76.63 \\
        FedAvg & 74.45 & 69.50 & 76.48 & 53.99 \\
        FedGNN & 81.51 & 70.10 & 70.63 & 56.91 \\
        FedSage+ & 72.97 & 57.97 & 77.52 & 55.48 \\
        \midrule
        FED-PUB (Ours) & 83.70 & 81.75 & 79.60 & 77.84 \\
        \bottomrule
    \end{tabular}
    }
    \label{tab:oracle}
    \vspace{-0.15in}
\end{wraptable}

\subsection{Impacts of Missing Edges to Performance Degeneration}
\label{appendix:results:missingedge}

In this subsection, we empirically demonstrate how much the performance is degenerated due to the missing edge problem, and how much the performance gain we obtain when we train with missing edges. In this work, we mainly conjecture that, due to the crucial issue of missing edges in subgraph FL, all FL methods that observe edges only within each subgraph show inferior performances to the Oracle method that trains on the entire graph including missing edges. To validate this claim, we first train the Oracle model on the connected global graph, and then evaluate it on disjoint subgraphs over all clients, on the Cora dataset of both Non-overlapping and Overlapping node scenarios with varying the number of clients. Note that this Oracle method is unrealistic in subgraph FL since it observes missing edges between subgraphs of distributed clients; therefore, this method is designed only for revealing the degenerated performance from missing edges and is not comparable to other FL methods. 

As shown in Table~\ref{tab:oracle}, the Oracle model outperforms all the other methods, while our FED-PUB achieves the closest performance to the Oracle. The above results bring us to conclude that, due to the problem of missing edges, all FL methods, which train with edges only within each subgraph, are inferior to the Oracle method. Also, this concluding result further suggests that the missing edge problem negatively affects the incompatible knowledge issue. Specifically, since all client models are trained on partial subgraphs, which are parts of the larger global graph, the trained parameters in the client and the aggregated parameters in the server might not capture globally meaningful knowledge that is helpful to all the other clients, while the Oracle model can capture. Therefore, the Oracle model can obtain superior performance to FL methods.

\section{Discussion on Limitations and Potential Societal Impacts}
\label{appendix:limitation}

In this section, we discuss the limitations and potential societal impacts of our work.

\paragraph{Limitations}
While our personalized subgraph FL framework, namely FED-PUB, is generally applicable regardless of subgraph types (e.g., unipartite graphs or bipartite graphs), our experiments are mainly done with unipartite graphs, since they are the most popular setups. However, the efficacy of our FED-PUB on the other types of graphs, such as bipartite graphs, would be interesting to investigate, which have not been explored much so far, and we leave this as future work.

\paragraph{Potential Societal Impacts}
The FL mechanism is important for preserving user's privacy, and, while this mechanism is actively studied in image and language domains, it gets little attention in graphs. However, we believe that our work comprehensively investigates and sufficiently tackles unique challenges in subgraph FL, such as missing nodes, edges, and their community structures, by proposing the novel approaches of using functional embeddings and local sparse masks.

The potential positive impact of our work on society is that, our FED-PUB can contribute to various domains that utilize graph-structured data, such as social, recommendation, and patient networks. Here we would like to emphasize the importance of our subgraph FL method, especially in social and recommendation networks. In current real-world applications, all user's interactions with other users in social networks and with other products in recommendation networks may be stored on the server. However, this may not preserve the user's privacy, but also has potential risks of user data leakage from the server, such that storing user's data in the server is not recommended by the existing data protection regularizations such as GDPR~\footnote{https://gdpr-info.eu/}. Yet, by applying our subgraph FL framework to this domain, we expect such problems could be alleviated by not storing user's interaction data in the server, but sharing only the locally trained machine learning models from client subgraphs.

However, the transmitted model parameters from the client to the server may hold privacy-sensitive information. While addressing it is not the main focus of our work (i.e., we assume that model parameters are transmittable without compromising privacy as in many FL works~\citep{McMahan2017CommunicationEfficientLO, arivazhagan2019federated, li2020federated}), the research community may need to put further effort on whether the model parameters are safe, and how to make them safer if they are not. Note that if they are not safe to share, we may further use differential privacy techniques~\cite{privacy/1, privacy/2}.

%% file: figures/fig_hyper_params.tex
\begin{figure}[t]
    \vspace{0.1in}
    \small
    \centering
    \begin{minipage}{0.49\textwidth}
        \resizebox{\textwidth}{!}{
            \renewcommand{\tabcolsep}{5.5mm}
            \renewcommand{\arraystretch}{1.2}
            \begin{tabular}{c c c c }
                \toprule 
        	    \midrule
        	    \centering
        	    \textbf{$\lambda_1$} & $\lambda_2$ & Accuracy [$\%$] & Sparsity [$\%$] \\
        	    \midrule
        	        3e-1 & 1e-3 & \textbf{79.62} $\pm$ 0.23 & 28.93	$\pm$ 0.52\\
                    5e-1 & 1e-3 & 79.42 $\pm$ 0.37 & 42.38	$\pm$ 0.35\\
                    7e-1 & 1e-3 & 78.68 $\pm$ 0.59 & 56.94	$\pm$ 0.29\\
                    9e-1 & 1e-3 & 77.36 $\pm$ 0.99 & \textbf{74.87}	$\pm$ 0.34\\
        	    \midrule
        	    \bottomrule 
            \end{tabular}
        }
    \end{minipage}
    \hfill
    \begin{minipage}{0.49\textwidth}
        \resizebox{\textwidth}{!}{
            \renewcommand{\tabcolsep}{5.5mm}
            \renewcommand{\arraystretch}{1.2}
            \begin{tabular}{c c c c }
                \toprule 
        	    \midrule
        	    \centering
        	    $\lambda_1$ & \textbf{$\lambda_2$} &  Accuracy [$\%$] & Sparsity [$\%$]  \\
        	    \midrule
        	        7e-1 & 1e-3 & 78.68 $\pm$ 0.59 & 56.94	$\pm$ 0.29 \\
                    7e-1 & 1e-2 & 78.56 $\pm$ 0.05 & 56.61	$\pm$ 0.32 \\
                    7e-1 & 1e-1 & \textbf{79.46} $\pm$ 0.41 & \textbf{57.41}	$\pm$ 1.33 \\
                    7e-1 & 1e-0 & 79.31 $\pm$ 0.45 & 57.28	$\pm$ 0.16 \\
        	    \midrule
        	    \bottomrule 
            \end{tabular}
        }
    \end{minipage}
    \vspace{-0.07in}
    \captionof{table}{\textbf{Sensitivity analyses on hyperparameters $\lambda_1$ and $\lambda_2$} by varying their values. We report the model performance and sparsity.}
    \label{fig:appendix_lambda}
\end{figure}